\documentclass[lettersize,journal]{IEEEtran}
\usepackage{amsmath,amsfonts}
\usepackage{algorithmicx}
\usepackage{algorithm}
\usepackage[noend]{algpseudocode}
\usepackage{array}
\usepackage[caption=false,font=scriptsize,labelfont=sf,textfont=sf]{subfig}
\usepackage{textcomp}
\usepackage{stfloats}
\usepackage{url}
\usepackage{bm}
\usepackage{multirow}
\usepackage{caption}
\usepackage{lipsum}
\usepackage{booktabs}
\usepackage{graphicx} 
\usepackage{verbatim}
\usepackage{threeparttable}
\usepackage{cite}
\usepackage{color}
\usepackage{mathrsfs}
\usepackage{makecell}
\definecolor{linkblue}{rgb}{0, 0.19, 0.32}
\usepackage[colorlinks, linkcolor=black]{hyperref}
\hypersetup{
    colorlinks=true, %设置链接的颜色
    urlcolor=linkblue
    }
\newcommand \transpose {\mathsf{T}} % matrix\vector transpose symbol

\begin{document}
% \title{Learning-based Detection and Tracking of Deformable Linear Objects from Occluded Point Clouds}
% \title{Real-time Detecting and Tracking Deformable Linear Objects from RGB-D Images with Occlusion}

\title{UniStateDLO: Unified Generative State Estimation and Tracking of Deformable Linear Objects\\Under Occlusion for Constrained Manipulation}

\author{
Kangchen Lv*, Mingrui Yu*, Shihefeng Wang, Xiangyang Ji, and Xiang Li
        % <-this % stops a space
\thanks{All authors are with the Department of Automation, Tsinghua University, Beijing, China (email: lkc24@mails.tsinghua.edu.cn; ymr20@mails.tsinghua.edu.cn; wshf21@mails.tsinghua.edu.cn; xyji@tsinghua.edu.cn; xiangli@tsinghua.edu.cn). Corresponding author: Xiang Li. }
% \thanks{Manuscript received April 19, 2021; revised August 16, 2021.}
}

\twocolumn[{
\renewcommand\twocolumn[1][]{#1}
\maketitle
\vspace{-1cm}
\begin{center}
    \captionsetup{type=figure, justification=justified, singlelinecheck=false}
    \includegraphics[width=\linewidth]{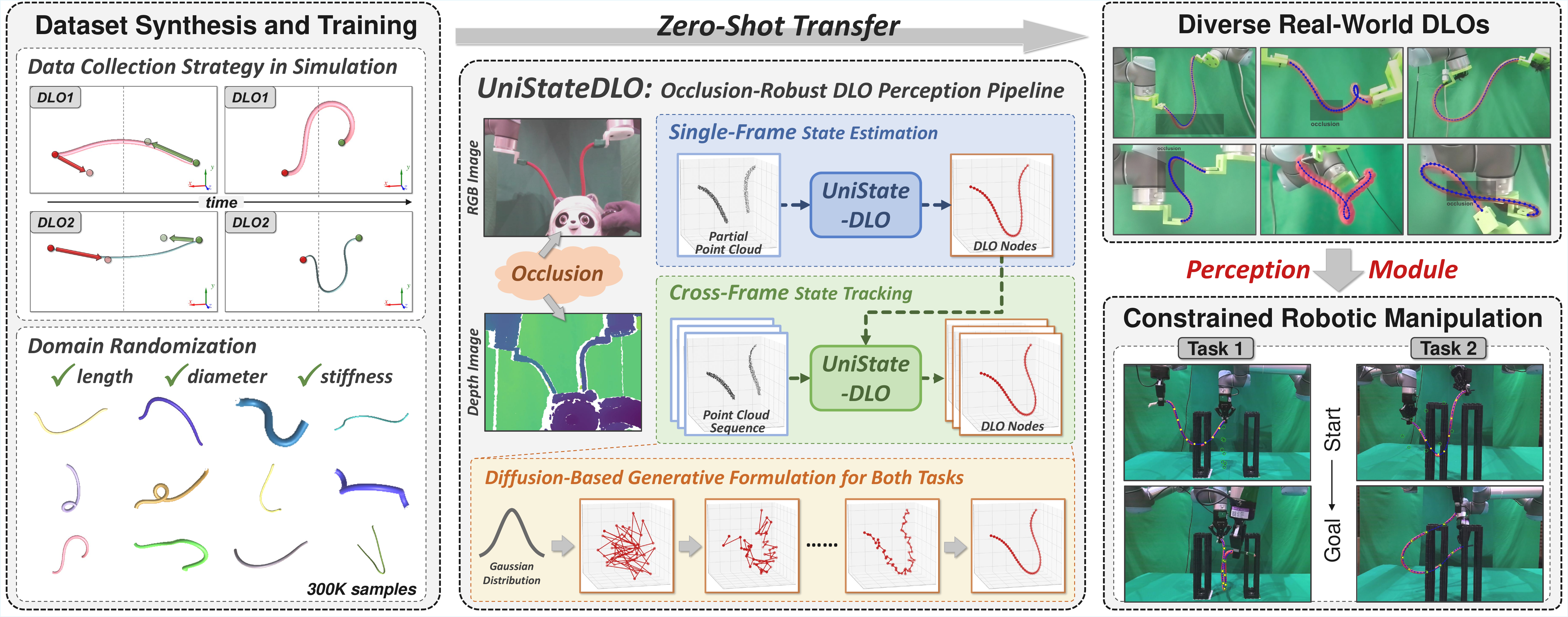}
    \captionof{figure}{We propose \textbf{UniStateDLO}, a novel unified perception framework for deformable linear objects (DLOs) that supports both \textit{single-frame} state estimation and \textit{cross-frame} tracking of DLOs under severe occlusions. Leveraging diffusion-based generative modeling, UniStateDLO reconstructs complete DLO configurations from even highly partial point clouds with strong accuracy, robustness and real-time performance. Trained entirely on synthetic data, it generalizes in a zero-shot manner to diverse real-world DLOs and provides a reliable perception front-end for constrained manipulation tasks.
    }
\end{center}
}]

\begin{abstract}
% Accurately and robustly estimating the state of deformable linear objects (DLOs), such as ropes and wires, is crucial for DLO manipulation and other applications. 
Perception of deformable linear objects (DLOs), such as cables, ropes, and wires, focuses on accurately and robustly estimating their 3-D states, which is the cornerstone for successful downstream manipulation.
Although vision-based methods have been extensively explored, they remain highly vulnerable to occlusions that commonly arise in constrained manipulation environments due to surrounding obstacles, large and varying deformations, and limited viewpoints.
Moreover, the high dimensionality of the state space, the lack of distinctive visual features, and the presence of sensor noises further compound the challenges of reliable DLO perception.
To address these open issues, this paper presents UniStateDLO, the first complete DLO perception pipeline with deep-learning methods that achieves robust performance under severe occlusion, covering both {\em single-frame} state estimation and {\em cross-frame} state tracking from partial point clouds.
Both tasks are formulated as conditional generative problems, leveraging the strong capability of diffusion models to capture the complex mapping between highly partial observations and high-dimensional DLO states.
UniStateDLO effectively handles a wide range of occlusion patterns, including initial occlusion, self-occlusion, and occlusion caused by multiple objects. In addition, it exhibits strong data efficiency as the entire network is trained solely on a large-scale synthetic dataset, enabling zero-shot sim-to-real generalization without any real-world training data.
Comprehensive simulation and real-world experiments demonstrate that UniStateDLO outperforms all state-of-the-art baselines in both estimation and tracking, producing globally smooth yet locally precise DLO state predictions in real time, even under substantial occlusions. Its integration as the front-end module in a closed-loop DLO manipulation system further demonstrates its ability to support stable feedback control in complex, constrained 3-D environments.
The project page is available at \href{https://unistatedlo.github.io/}{https://unistatedlo.github.io/}.
\end{abstract}


\section{Introduction}
\IEEEPARstart{D}{eformable} linear objects (DLOs), including ropes, wires, and cables, are one-dimensional deformable structures that frequently appear in manufacturing, service, and surgical applications \cite{zhu2021challenges, yin2021modeling, gao2022hierarchical}.
Unlike rigid objects, the shape of DLOs will vary along their length due to bending and deformation. 
Enabling robotic systems to automatically manipulate DLOs in tasks, such as shape control \cite{yu2022global, lv2022dynamic}, cable routing \cite{jin2022robotic, luo2024multistage}, and knot tying \cite{peng2024tiebot,jiang2024automated}, fundamentally relies on accurate and real-time DLO state estimation, which serves as the cornerstone for closed-loop control.

Although many approaches have been developed in recent years to improve DLO perception, the infinite-dimensional state space and frequent occlusions in constrained environments make it still a challenging research issue. First, the state space of DLO possesses nearly infinite degrees of freedom under deformation. In practice, its state is often simplistically represented by a discretized chain of uniformly distributed nodes \cite{huo2022keypoint, lv2022dynamic}, yet this representation still results in hundreds or even thousands of dimensions. Moreover, occlusions caused by obstacles or self-interactions, particularly during manipulation in constrained environments, demand greater robustness in perception systems to reliably reconstruct the full DLO state from partial observations.

A complete 3-D DLO state estimation pipeline can be roughly divided into three stages: segmentation (i.e., segmenting pixel-level DLO masks from scenes \cite{caporali2022fastdlo,caporali2023rtdlo, zanella2021auto}), detection (i.e., estimating the DLO state in a single frame \cite{yan2020self, keipour2022deformable}), and tracking (i.e., capturing deformations across sequential frames \cite{yang2022particle, wang2021tracking, xiang2023trackdlo}). 
For single-frame estimation, existing approaches often begin by extracting DLO skeleton lines from 2-D images \cite{keipour2022deformable, kicki2023dloftbs} or 3-D point clouds \cite{wnuk2020kinematic, zhaole2023robust}, followed by merging disconnected segments through manually designed strategies. Some researchers also incorporate data-driven methods \cite{lyu2023learning} to enhance the performance under occlusion or complex topology. 
For cross-frame tracking, a classical line of work formulates the problem as non-rigid point set registration with multiple geometric constraints \cite{schulman2013tracking, tang2017state, tang2018track, chi2019occlusion, wang2021tracking, xiang2023trackdlo}, where DLO nodes are modeled as centroids of a Gaussian Mixture Model (GMM), and the observed point cloud is treated as samples drawn from it. Some recent works also explore to perform DLO tracking with particle filtering \cite{yang2022particle} and 3-D Gaussian splatting techniques \cite{dinkel2025dlo}.

However, previous single-frame estimation approaches only exploit individual frames and neglect temporal continuity, whereas tracking approaches critically depend on accurate and robust initialization. In constrained manipulation scenarios, the common occurrence of long-term and large-scale occlusions further degrades their performance. To address these challenges, we introduce UniStateDLO, a unified framework that formulates both DLO state estimation and tracking under occlusions as a conditional generation task. Motivated by the remarkable capability of diffusion models\cite{ho2020denoising} in learning complex probabilistic distributions, we hypothesize that they can effectively resolve the uncertainty of DLO nodes given partial observations. 
Intuitively, one could condition the diffusion model on a global embedding extracted from the DLO point cloud, which then samples 3-D node locations through iterative denoising. However, due to the weak visual distinctiveness of DLO point clouds, such global features often lack the fine-grained, node-wise geometric cues, which motivates conditioning the model on richer local features to fully leverage the potential of generative modeling.

For DLO state estimation, we propose a novel two-branch network architecture with a diffusion-based fusion module to generate the final 3-D node predictions. Both branches share a PointNet++\cite{qi2017pointnet++} encoder but focus on complementary information: one leverages global features to achieve robustness under occlusions, while the other exploits local features to ensure precise node-wise estimation. By using the 3-D node predictions from both branches as per-node local conditions for the diffusion model, our approach can reconstruct occluded portions of the DLO while maintaining high local accuracy.
Once the state estimation is done in the initial frame, inter-frame node motions are predicted iteratively based on the previous frame's results to enable sequential tracking. K-nearest-neighbor-based feature aggregation module is employed to extract per-node features around last-frame nodes, which provide local conditions for the subsequent diffusion model in a manner consistent with the single-frame estimation framework. 
By leveraging generative modeling through diffusion models in both single-frame estimation and cross-frame tracking, UniStateDLO can \textit{imagine} complete DLO configurations from even heavily occluded point clouds, delivering a robust perception module for precise and reliable DLO manipulation in constrained scenarios.

Our model is trained on a large-scale synthetic dataset of 300K samples and generalizes to real-world DLOs with substantially different physical properties in a zero-shot manner. Across both simulation and real-world evaluations with diverse occlusion patterns, our approach consistently outperforms existing baselines. Its deployment as the real-time perception front-end in challenging shape control tasks within constrained environments, where multiple obstacles introduce large-scale, long-term occlusions and require continuous collision avoidance, further demonstrates its effectiveness. 
% To the best of our knowledge, this is the first method capable of achieving accurate and robust DLO state estimation and tracking under severe occlusions, enabling seamless integration into closed-loop DLO manipulation systems.

% In a recent work \cite{lyu2023learning}, the authors introduce a two-branch architecture for occlusion-robust 3-D DLO state estimation from point clouds in individual frames. 
% Here, we extend single-frame state estimation to sequential  tracking by exploiting temporal information to reduce inconsistencies across consecutive frames, a capability crucial for closed-loop manipulation. We further propose the diffusion-based generative modeling formulation to enhance robustness under occlusions, replacing the non-rigid point set registration–based fusion module used in our prior single-frame estimation method. In addition, more detailed evaluations are conducted to comprehensively demonstrate the effectiveness of our approach against state-of-the-art DLO state estimation and tracking algorithms.
In summary, our primary contributions are as follows: 
\begin{enumerate}
\item We present UniStateDLO, a unified DLO perception pipeline that supports both single-frame state estimation and cross-frame tracking from partial point clouds, achieving strong robustness to severe occlusions while preserving temporal consistency and accuracy.
\item We formulate both state estimation and tracking as conditional generative tasks, leveraging diffusion models to resolve node-level uncertainty under occlusions and reconstruct the full DLO configurations accurately.
\item We conduct extensive simulation and real-world experiments, demonstrating the outperformance of our method over existing works and its applicability as a reliable perception front-end in constrained manipulation tasks.
\end{enumerate}

\section{Related Works}
\subsection{DLO Manipulation and Perception}
Manipulating DLOs, such as cables and wires, is crucial for a wide range of manufacturing and assembly applications. Extensive researches have explored autonomous robotic manipulation of DLOs in diverse tasks, including general shape control \cite{yu2022global,lv2022dynamic}, cable routing through clips \cite{jin2022robotic, luo2024multistage}, cable sorting \cite{gao2022hierarchical, gao2023development}, knot tying \cite{peng2024tiebot, jiang2024automated}, and untangling of knots or multiple wires \cite{grannen2020learning, huang2024untangling}.
In most manipulation frameworks, visual perception serves as the foundation for downstream planning and control by providing the 3-D DLO configuration in real time. However, achieving accurate and robust perception remains highly challenging due to the high-dimensional state space and complex deformations of DLOs during manipulation. In particular, constrained environments \cite{yu2024generalizable, tang2024learning} often introduce severe occlusions caused by both environmental obstacles and self-intersections, further complicating reliable perception with strong robustness.
Some prior works \cite{yu2022global, sintov2020motion} on DLO manipulation, though not primarily focused on perception, simplify the sensing of DLOs by detecting visual markers uniformly attached along the DLO. More generally, a complete DLO perception pipeline typically consists of three stages: segmentation, detection, and tracking: segmenting the DLO region from raw observations, estimating its 3-D state in single frames, and temporally tracking across frames, respectively. As DLO segmentation has been extensively studied \cite{caporali2022fastdlo, caporali2023rtdlo, zanella2021auto} and can also be achieved by general-purpose segmentation systems such as SAM \cite{kirillov2023segment, zhang2023faster}, we primarily focus on the latter two stages, single-frame estimation and cross-frame tracking, in this article. The limitations of existing approaches and the advantages of our proposed UniStateDLO are summarized in Table~\ref{tab:comparison_related_works}.

{\renewcommand{\arraystretch}{1.2}
\begin{table*}
\centering
\caption{Comparison of existing single-frame DLO state estimation and cross-frame tracking methods.}
\setlength\tabcolsep{3pt}
\begin{tabular}{p{2.85cm}<{\centering}|p{2.7cm}<{\centering}|p{5cm}<{\centering}|p{6.6cm}<{\centering}}
\toprule
\textbf{Task} & \textbf{Category} &  \textbf{Methods} &\textbf{Limitations / \textit{Advantages}}\\
\midrule
\multirow{9}{*}{\centering Single-Frame Estimation} & \multirow{5}{*}{\centering 2-D Image-Based} &  CNN-based Keypoint Detection: Yan et al. \cite{yan2020self}, Huo et al. \cite{huo2022keypoint} & \multirow{2}{*}{\centering Only suitable for planar shapes, cannot handle occlusion} \\
 & & Triangulation: Caporalli et al. \cite{caporali2023deformable, caporali2025robotic} & Need multi-view cameras with known viewpoints\\
 &  & Fit 2-D Skeletons with Curves: Keipour et al.\cite{keipour2022deformable}, DLOFTBs\cite{kicki2023dloftbs} & \multirow{2}{*}{\centering Sensitive to noises, not robust under severe occlusion} \\
  \cmidrule(lr){2-4}
 & \multirow{4}{*}{\centering 3-D Point Cloud-Based} &  Fit 3-D Skeletons with Curves: Wnuk et al. \cite{wnuk2020kinematic}, Sun et al. \cite{zhaole2023robust}, Cao et al. \cite{cao2025deformable} & \multirow{2}{*}{\centering Cannot generalize well on diverse DLOs in 3-D space}\\
 &  & Learning From Point Clouds: Lv et al. \cite{lyu2023learning} &  Rely on manually-tuned registration params for fusion\\
 &  & \textbf{\textit{Ours: UniStateDLO}} & \textit{Strong occlusion-robustness, accuracy and generalization} \\
\midrule
\multirow{9}{*}{\centering Cross-Frame Tracking} & \multirow{5}{*}{\centering Registration-Based} & Regis. + Physics Simulation: Schulman et al.\cite{schulman2013tracking}, Tang et al.\cite{tang2018framework}, SPR \cite{tang2018track} & \multirow{2}{*}{Need simulation engines, highly time-consuming}\\
 &  & Regis. + FEM: Wang et al. \cite{wang2022real} & Restrict to certain materials, hard to generalize\\
 &  & Regis. + Topological Constraints: CDCPD \cite{chi2019occlusion}, CDCPD2 \cite{wang2021tracking}, TrackDLO \cite{xiang2023trackdlo} & Require known initial state, low accuracy and temporal smoothness under heavy occlusion\\
 \cmidrule(lr){2-4}
 & Particle Filter-Based & Yang et al. \cite{yang2022particle} & Assume known occlusion mask, hard to generalize\\
 \cmidrule(lr){2-4}
 & 3-D GS-Based &  DLO-Splatting \cite{dinkel2025dlo} & Need multi-view cameras and simulator, low speed\\
 \cmidrule(lr){2-4}
 & Learning-Based &  \textbf{\textit{Ours: UniStateDLO}} & \textit{Strong occlusion-robustness, accuracy and generalization}\\
\bottomrule
\end{tabular}
\label{tab:comparison_related_works}
\end{table*}}

\subsection{Single-Frame DLO State Estimation}
Accurately estimating the DLO state from a single frame is fundamental for DLO perception, either as a standalone prediction from individual frames or as the initialization for subsequent tracking.
Yan et al. \cite{yan2020self} encode RGB images into sequential segments, and Huo et al. \cite{huo2022keypoint} detect 2-D keypoints in images with CNN and then refine them geometrically, but both assume full visibility.
To handle occlusions, many works extract 2-D skeletons from binary masks, which often fragment under occlusions, and reconnect them smoothly, optionally lifting the result to 3-D using depth data. Following this idea, Keipour et al. \cite{keipour2022deformable} design several geometric cost functions to merge skeleton segments, while Kicki et al. \cite{kicki2023dloftbs} perform B-spline fitting across them. Caporali et al. \cite{caporali2023deformable, caporali2025robotic} exploit a multi-view stereo-based approach to reconstruct the 3-D DLO shape from multiple 2-D images. 
Point cloud–based methods instead extract centerlines in 3-D space directly: Wnuk et al. \cite{wnuk2020kinematic} operate directly on raw points, while Sun et al. \cite{zhaole2023robust} and Cao et al. \cite{cao2025deformable} further refine the DLO shape with a discrete elastic rod model \cite{bergou2008discrete}. 
Despite these efforts, existing geometric pipelines remain brittle under severe occlusions and generalize poorly across DLOs with diverse physical properties. Lv et al. \cite{lyu2023learning} introduce the first data-driven approach, using a dual-branch network followed by non-rigid registration-based fusion, but the fusion module relies heavily on manually tuned parameters and is still sensitive to large missing regions.
In this paper, we unify single-frame estimation and cross-frame tracking under a conditional generative formulation that learns the distribution of DLO state fully from large-scale data, achieving improved accuracy and robustness under diverse occlusion patterns and physical variations.

\subsection{Cross-Frame DLO State Tracking}
Tracking across frames differs from single-frame estimation in that it aims to accurately infer the current DLO state given historical information while enforcing temporal continuity and topological consistency.
Most existing DLO tracking methods are built upon non-rigid point-set registration algorithms such as Coherent Point Drift (CPD) \cite{myronenko2010point} and Global-Local Topology Preservation (GLTP) \cite{ge2014non}, which treat DLO nodes as Gaussian Mixture Model (GMM) centroids and use the EM algorithm to maximize the likelihood of observing the current point cloud.
To impose physical constraints of DLOs, several works utilize physics simulation to augment registration: Tang et al. \cite{tang2017state} integrate CPD with a physics engine for iterative updates, and SPR \cite{tang2018framework} further incorporates locally linear topology regularization. 
Because physics simulation is computationally expensive and often impractical for real-world scenarios, recent efforts move toward simulation-free tracking. Wang et al. \cite{wang2022real} uses finite element method (FEM) to avoid simulation; CDCPD \cite{chi2019occlusion} and CDCPD2 \cite{wang2021tracking} introduce stretching and convex geometric constraints; and TrackDLO \cite{xiang2023trackdlo} leverages motion coherence to infer occluded-node spatial velocities from visible ones.
Meanwhile, data-driven alternatives have emerged, including particle filtering in a low-dimensional latent space \cite{yang2022particle} and 3-D Gaussian Splatting for complex topological deformations \cite{dinkel2025dlo}.
In contrast to these approaches, we adopt an end-to-end generative modeling framework that directly predicts node-wise motion through a conditional diffusion process, achieving more consistent performance under severe occlusions and large-scale motions.

\subsection{Diffusion Models for State Estimation}
Diffusion models \cite{ho2020denoising} are a class of probabilistic generative models that generate samples from the prior distribution via an iterative denoising process. Owing to their strong capability in modeling high-dimensional, complex distributions, diffusion models have been widely adopted in domains such as image generation \cite{ho2022cascaded, rombach2022high}, motion planning \cite{janner2022diffuser, carvalho2023motion}, and policy learning \cite{chi2024diffusion}.
Researchers have also adapted diffusion models for human pose \cite{choi2023diffupose,shan2023diffusion} and hand pose estimation \cite{ivashechkin2023denoising} based on RGB images, where the 2D-to-3D lifting process is modeled probabilistically. For example, D3DP \cite{shan2023diffusion} learns an iterative denoiser conditioned on 2-D keypoints to recover 3-D poses, whereas Ivashechkin et al. \cite{ivashechkin2023denoising} condition the diffusion process on CNN features. To avoid performance bottlenecks imposed by 2-D regression models, HandDiff \cite{cheng2024handdiff} instead conditions directly on 3-D joint-wise local features.
For deformable object perception, UniClothDiff \cite{tian2025diffusion} similarly employs a diffusion model conditioned on a global embedding produced by Transformer to reconstruct full cloth states with self-occlusions.
In this paper, we explore diffusion-based generative formulation for DLO perception from partial point clouds. However, the thin and elongated geometry of DLOs provides limited cues for distinguishing individual nodes, making global feature insufficient for precise predictions. To overcome this limitation, 
we design a node-wise local conditioning scheme that enables diffusion models to better resolve DLO state uncertainty under occlusion.

\section{Overview}
\subsection{Problem Statement}
\begin{figure} [tb]
  \centering 
    \includegraphics[width=0.985\linewidth]{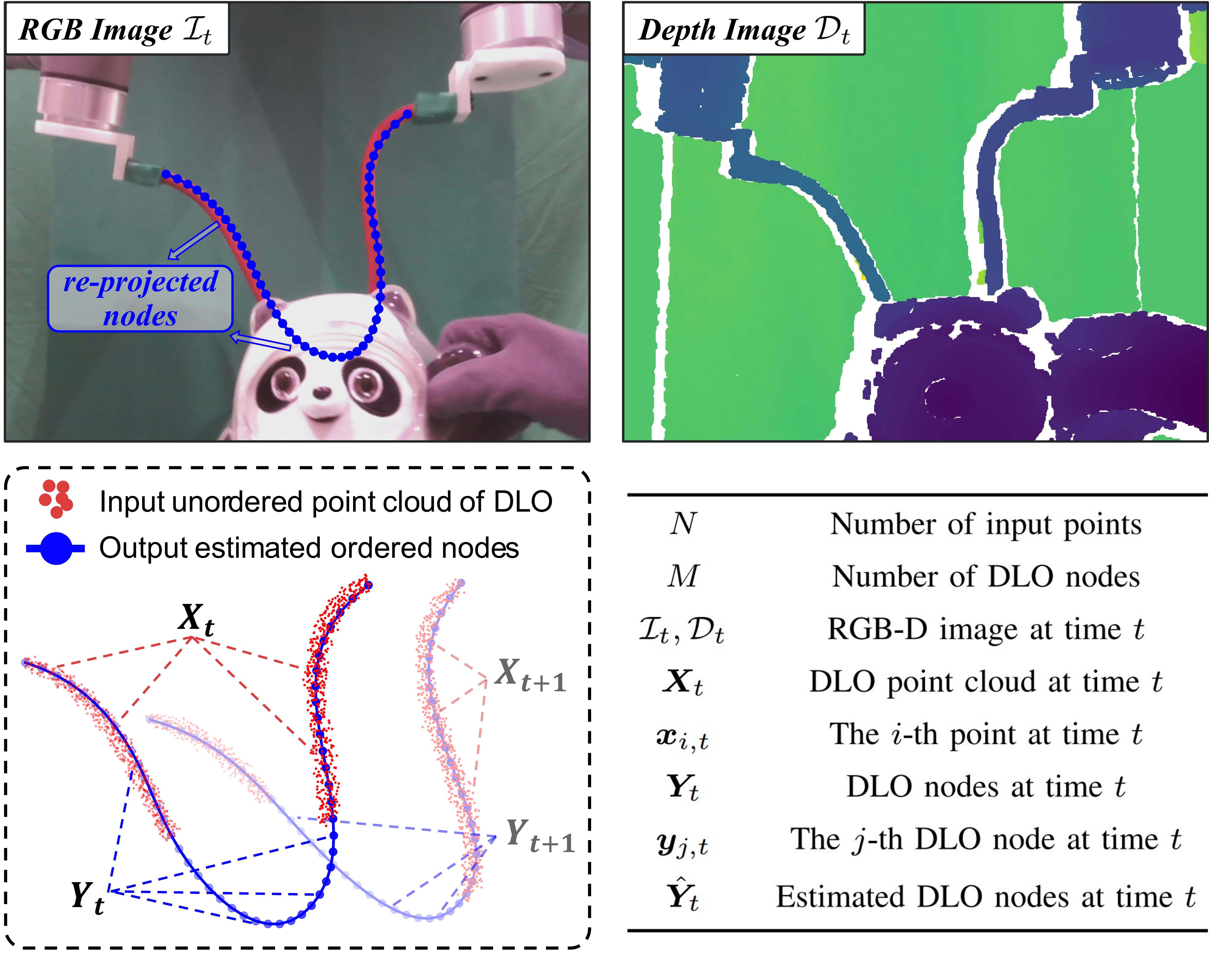} 
  \vspace{-0.5mm}
  \caption{Illustration of the DLO perception task and the notation of key variables. Given partial DLO point clouds (red points) extracted from RGB-D images, single-frame state estimation and cross-frame tracking aim to reconstruct a sequential chain of nodes (blue connected dots), either independently from each frame or across a temporal sequence.}
  \label{fig:notation}
    % \vspace{-2mm}
\end{figure}

As illustrated in Fig.~\ref{fig:notation}, the DLO point cloud $\bm X_t \in \mathbb{R}^{N \times 3}$ is first obtained from the scene using the RGB image $\mathcal{I}_t$ and depth image $\mathcal{D}_t$ captured at timestep $t$. Following prior works\cite{lyu2023learning, yu2022global, xiang2023trackdlo}, the DLO state is represented as a discretized chain of uniformly distributed \textit{nodes}, $\bm Y_t=\left[\bm y_{1,t}, \bm y_{2,t}, \cdots , \bm y_{M,t} \right]^\transpose{} \in \mathbb{R}^{M\times 3}$, where the predefined node number $M$ is chosen to sufficiently capture the DLO configuration. Note that the input point cloud $\bm X_t$ is unordered, whereas the DLO nodes in $\bm{Y}_t$ are ordered from one endpoint to the other, with indices $1,2,\cdots,M$.
The DLO perception problem is therefore formulated as estimating node coordinates $\bm{\hat{Y}}_{t}$ either from the current point cloud $\bm{X}_t$ (single-frame estimation) or from a temporal sequence (cross-frame tracking). Given partial and noisy point clouds caused by occlusions, imperfect segmentation, and depth sensing errors, our objective is to minimize $\lVert \bm{\hat{Y}}_t - \bm Y_t \rVert$ without relying on any explicit priors about DLO properties or occlusion regions.

% \begin{table}
% \centering
% \caption{Notations.}
% % \setlength\tabcolsep{3pt}
% \begin{tabular}{cc} 
% \toprule
% \addlinespace[3pt]
% $N$ & Number of input points \\
% \addlinespace[4pt]
% $M$ & Number of DLO nodes \\
% \addlinespace[4pt]
% $\mathcal{I}_t, \mathcal{D}_t$ & RGB-D image at time $t$ \\
% \addlinespace[4pt]
% $\bm X_t$ & DLO point cloud at time $t$ \\
% \addlinespace[4pt]
% $\bm x_{i,t}$ & The $i$-th point at time $t$ \\
% \addlinespace[4pt]
% $\bm Y_t$ & DLO nodes at time $t$ \\
% \addlinespace[4pt]
% $\bm y_{j,t}$ & The $j$-th DLO node at time $t$ \\
% \addlinespace[4pt]
% $\hat{\bm{Y}}_t$ & Estimated DLO nodes at time $t$ \\
% \addlinespace[1pt]
% \bottomrule
% \end{tabular}
% \end{table}

\subsection{Overall Pipeline}
\begin{figure*} [tb]
  \centering 
    \includegraphics[width=\textwidth]{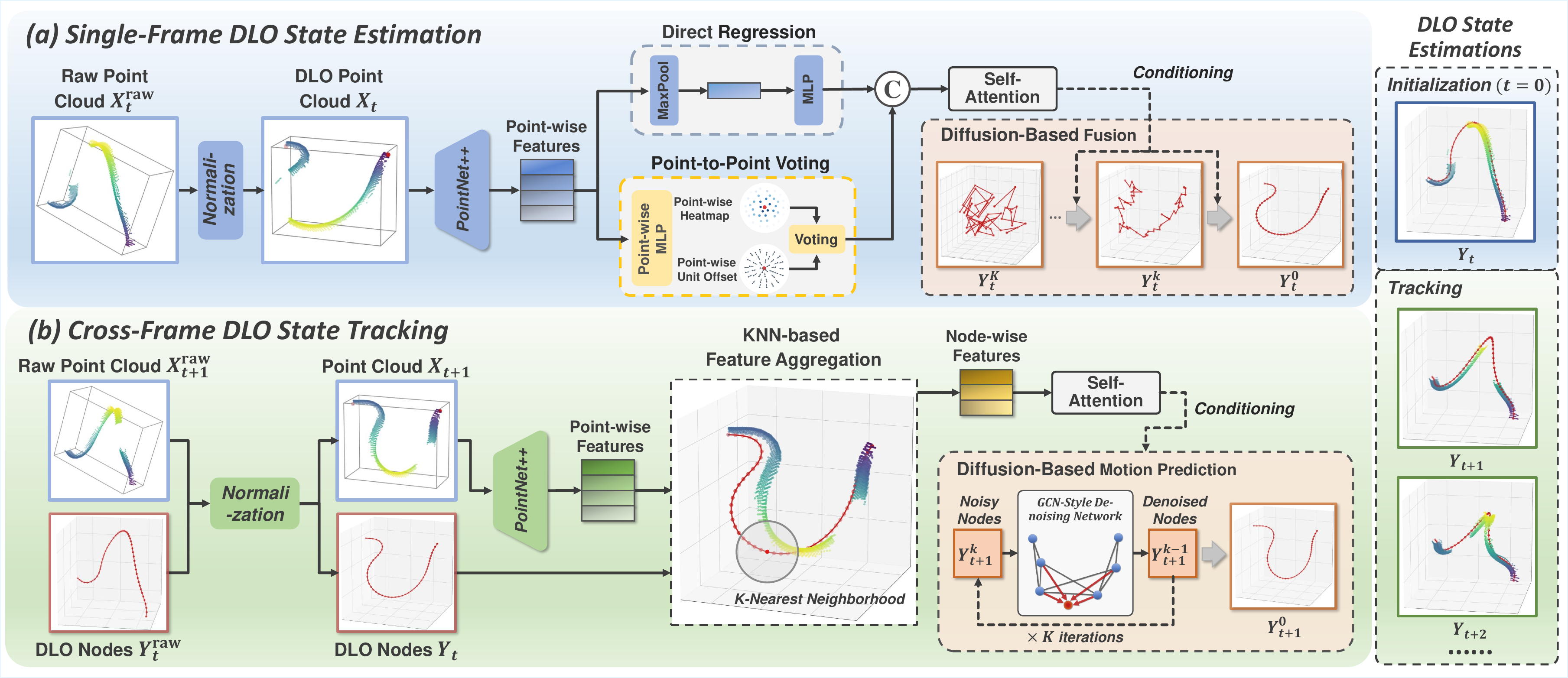} 
  \vspace{-0.55cm}
  \caption{Overview of the proposed UniStateDLO pipeline, comprising \textbf{\textit{Single-Frame State Estimation}} for initialization and \textbf{\textit{Cross-Frame State Tracking}} for sequential motion tracking. Given a partial DLO point cloud, state estimation module first produces coarse predictions through two complementary branches based on PointNet++ features, and then refines them via a diffusion model. For cross-frame tracking, a KNN-based feature aggregation module extracts node-wise local features around the previous frame's predictions, followed by another diffusion model to infer per-node cross-frame motion.}
  \label{fig:overview}
\end{figure*}

The overall UniStateDLO framework, consisting of both single-frame estimation and cross-frame tracking with occlusion robustness, is illustrated in Fig.~\ref{fig:overview}.
Although the single-frame estimation module, which aims to infer the DLO configuration solely from the current partial point cloud, can be applied to each frame independently, the lack of temporal information prevents it from ensuring topological consistency and temporal smoothness. Therefore, it is primarily used to produce an accurate and robust initial state at $t=0$. 
Once the initial state is obtained, the cross-frame tracking module then takes the current point cloud $\bm X_{t+1}$ with the previously estimated nodes $\bm Y_{t}$ as input, and predicts per-node motion across consecutive frames. Even under severe occlusions, this sequential tracking is able to recover accurate DLO shapes while preserving structural properties.
If tracking failure is detected, such as during long-term and extreme occlusions, the single-frame estimation module can be invoked again to reinitialize and resume reliable tracking.

For single-frame state estimation, the raw point cloud is first transformed into a canonical coordinate system using the two endpoints, ensuring consistent global orientation and improving robustness to large viewpoint variations. Unless otherwise specified, the normalized point cloud is denoted as $\bm X_t$ for simplicity.
Point-wise features are then extracted using a PointNet++ encoder. Although the node positions can be directly regressed from the global feature via an MLP, the thin and feature-sparse nature of DLO point clouds makes global embeddings insufficient for capturing fine-grained local geometry, hindering accurate node discrimination.
To address this, we introduce two complementary branches: a regression branch that leverages global information and a voting branch that exploits local point-to-point cues. Their coarse predictions are subsequently fused by a conditional generative fusion module, which uses a diffusion model to learn the complex mapping from coarse to final states, achieving estimates that are both globally robust to occlusion and locally precise.

For cross-frame state tracking, the current point cloud $\bm X_{t+1}$ and the previous-frame nodes $\bm Y_t$ are both transformed with the same normalization procedure as in single-frame estimation. Since node motions between adjacent frames are typically small, the last-frame nodes serve as coarse predictions for the current frame. After passing $\bm X_{t+1}$ through another PointNet++ encoder, we extract node-wise local features using a k-nearest-neighbor (KNN)–based aggregation module, where each previous-frame node acts as a centroid and gathers features within its neighborhood.
Conditioned on node-wise features, a diffusion model predicts the motion across frames, with a graph convolutional layer incorporated into the denoising process to better capture the spatial connectivity of the DLO structure. Note that the denoising network architecture in single-frame estimation and cross-frame tracking is identical. In the following several sections, we will sequentially describe the single-frame state estimation module (Sec.~\ref{section:detection}), the cross-frame tracking module (Sec.~\ref{section:tracking}), and the whole pipeline including pre- and post-processing methods (Sec.~\ref{section:pipeline}).

\section{Single-Frame DLO State Estimation}\label{section:detection}

In this section, we introduce the two-branch architecture, including $\textit{Direct Regression}$ and $\textit{Point-to-Point Voting}$, together with the diffusion-based fusion module for single-frame DLO state estimation. Since no temporal information is involved in this section, we omit timestep notation $t$ for clarity and denote the DLO point cloud as $\bm X$ and DLO nodes as $\bm Y$.

% We exploit a hierarchical PointNet++ encoder \cite{qi2017pointnet++} to extract 3D local geometric features, which are then processed by two complementary branches: an direct regression branch for global estimation and a point-to-point voting branch for local estimation. The coarse predictions from these two branches are further refined by a diffusion-based fusion module, yielding both occlusion-robust and accurate node estimations.

% As shown in Fig.~\ref{fig:overview}, this single-frame DLO state detection method contains an \textit{Direct Regression} branch and a \textit{Point-to-Point Voting} branch which focuses on the global and local geometry information, respectively.  
% Then, a deformable registration module is designed to leverage the advantages of both branches and fuse their two predictions to output the final estimated node sequence. 

\subsection{Direct Regression Branch}

The most intuitive approach is to train a regression network that maps the input point cloud $\bm X \in \mathbb{R}^{N\times 3}$ to the output node coordinates $\bm Y \in \mathbb{R}^{M\times 3}$, referred to as \textit{Direct Regression}. 
We employ a PointNet++ encoder \cite{qi2017pointnet++}, denoted as $\mathcal{F}(\cdot)$, to extract point-wise local features $\bm F_{\rm local} \in \mathbb{R}^{N\times d}$ from $\bm X$. A max pooling layer is then applied to aggregate the global feature $\bm F_{\rm global} \in \mathbb{R}^{d}$, which is finally fed into a multi-layer perceptron ${\rm MLP}_{\rm reg}$ to produce the predicted node coordinates $\bm{\hat{Y}}_{\rm reg}$:
\begin{equation}
    \bm{\hat{Y}}_{\rm reg} = {\rm{MLP}}_{\rm reg}({\rm MaxPool} (\mathcal{F}(\bm X))).
\end{equation}
To improve robustness to outliers and avoid vanishing gradients near zero, we adopt an L1 loss rather than an MSE loss. Given the ground-truth node coordinates $\bm Y^*$, the training objective becomes:
\begin{equation}
    \mathcal{L}_{\rm reg}= \Vert \bm{\hat{Y}}_{\rm reg} - \bm Y^* \Vert.
\end{equation}

In practice, this simple network produces smooth DLO configurations even under substantial occlusions, showing it sufficiently captures the overall characteristics of DLO shapes. However, relying solely on global features, which discards fine-grained point-wise local information, makes it difficult to distinguish individual nodes. As a result, the predictions often exhibit a slight 3-D bias compared to the ground-truth states (see Fig.~\ref{fig:detection_occlusion}), limiting its suitability for real-world applications.

\subsection{Point-to-Point Regression Branch}

To overcome the limitations of the direct regression branch, we design a point-to-point voting framework that leverages local geometric information more effectively, inspired by prior works \cite{wan2018dense, ge2018point}.
Instead of aggregating features with a max pooling layer, this branch produces point-wise estimations $\hat{\bm{Y}}^{1}_{\rm vot},\hat{\bm{Y}}^{2}_{\rm vot},\cdots,\hat{\bm{Y}}^{N}_{\rm vot}$ from each input point $\bm{x}_1, \bm{x}_2, \cdots, \bm{x}_N$.
Concretely, for each input point $\bm{x}_i$ and each node $\bm{y}_j$, the network regresses an offset vector $\bm{O}_{i,j}$ pointing from $\bm{x}_i$ to $\bm{y}_j$. 
During inference, the point-to-point estimation is computed as $\hat{\bm{y}}^i_j=\bm{x}_i + \hat{\bm{O}}_{i,j}$ and the set of point-wise predictions $\hat{\bm{Y}}^{i}_{\rm vot}$ are then aggregated through a voting scheme to produce the final DLO node estimation $\hat{\bm{Y}}_{\rm vot}$.

\begin{figure} [tb]
  \centering 
    \includegraphics[width=8.25cm]{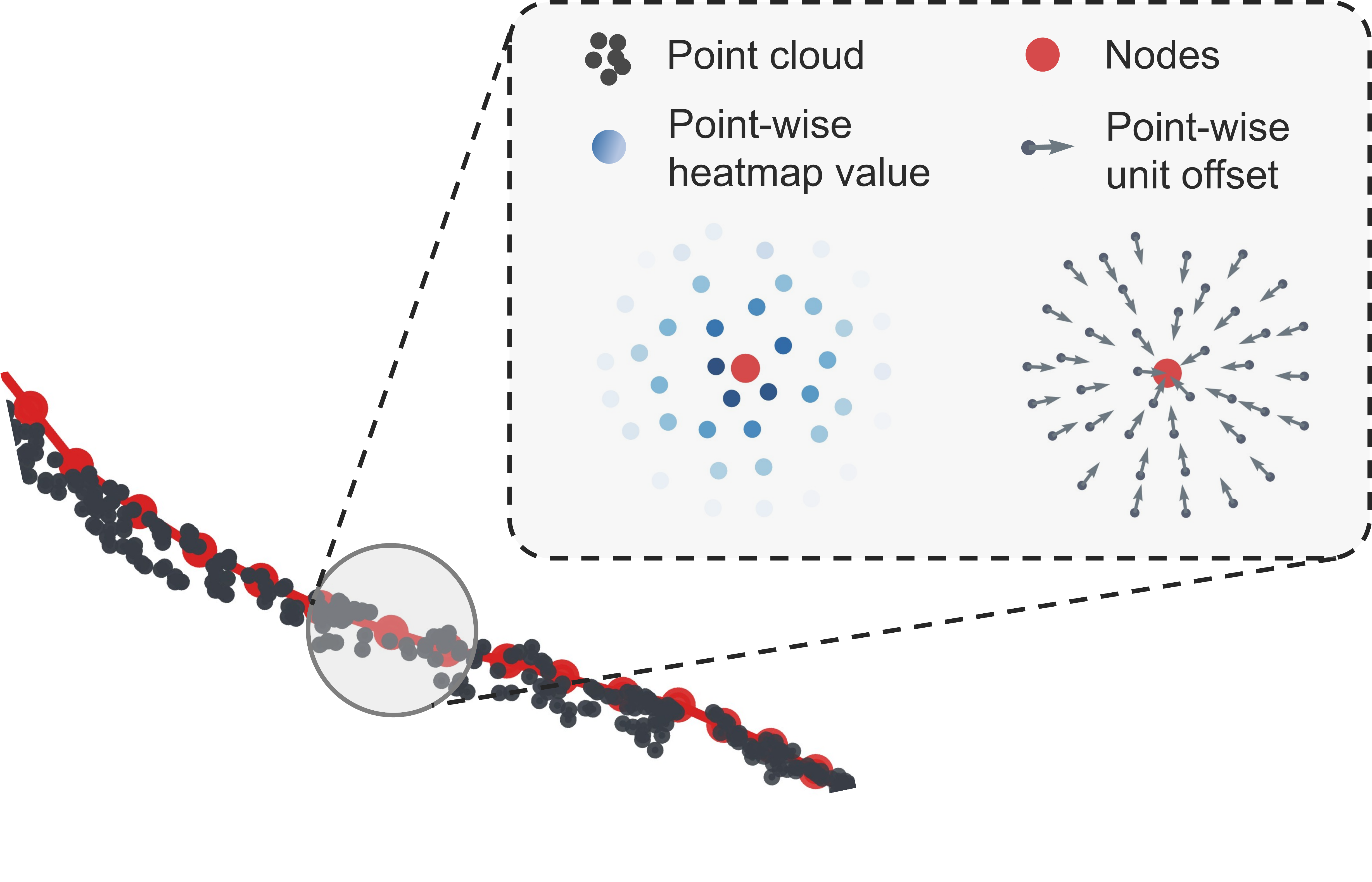} 
  \vspace{-2.5mm}
  \caption{Demonstration of predicted point-wise heatmap value and unit offset. Considering the neighborhood of one node, the points closer to it will have a higher heatmap value (visualized as deeper color), and the unit offset represents the normalized direction from the input point to the desired node.}
  \label{fig:voting}
    \vspace{-2mm}
\end{figure}

We further decompose each point-wise offset vector $\bm O_{i,j}$ into two components to facilitate easier and more stable network training: a heatmap value $H_{i,j}$ encoding the distance from $\bm x_i$ to $\bm y_j$, and a unit offset vector $\bm U_{i,j}$ indicating the direction, as illustrated in Fig.~\ref{fig:voting}. 
To exclude the impact of noisy and inaccurate estimations from distant points, we further constrain ground-truth supervision of heatmap value to only those point-wise estimations whose corresponding input points lie within the neighborhood of the target node. Accordingly, given a neighborhood radius $r$, the ground-truth heatmap value $H_{i,j}^*$ is defined as:
\begin{equation}\label{eq:Hij}
    H^*_{i,j}=\left\{
\begin{array}{ccl}
1-\|\bm x_i-\bm y_j\|/r      &  , &    \|\bm x_i-\bm y_j\| <r,\\
0     & , & \|\bm x_i-\bm y_j\| \geq r,
\end{array} \right.
\end{equation}
and the ground-truth unit offset vector $\bm U_{i,j}^*$ is defined as:
\begin{equation}\label{eq:Uij}
    \bm U^*_{i,j}=(\bm y_j-\bm x_i)\,/\,\|\bm x_i-\bm y_j\|\,.
\end{equation}

From the local features $\bm F_{\rm local}\in \mathbb{R}^{N\times d}$, the heatmap values $\hat{\bm H}\in \mathbb{R}^{N \times M}$ and offset vectors $\hat{\bm U}\in \mathbb{R}^{N\times M \times 3}$ are predicted with two separate point-wise MLP layers. Thus, the estimations are expressed as:
\begin{equation}
\hat{\bm H} = {\rm Sigmoid} \left( {\rm MLP}_{\rm heatmap}(\bm F_{\rm local}) \right),
\end{equation}
\begin{equation}
\hat{\bm U} = {\rm Normalize} \left( {\rm MLP}_{\rm offset}(\bm F_{\rm local}) \right),
\end{equation}
where the Sigmoid function constrains the predicted heatmap values to the range $[0,1]$, and $\text{Normalize}(\cdot)$ enforces unit length for each predicted offset vector.

During inference, the point-wise estimation for node $\bm y_j$ from input point $\bm x_i$ can be obtained based on the definitions of the heatmap value and unit offset vector in Eq.~\ref{eq:Hij} and Eq.~\ref{eq:Uij}, formulated as:
\begin{equation}
\hat{\bm y}^{i}_j = r \cdot(1-\hat{H}_{i,j})\cdot\hat{\bm U}_{i,j} + \bm x_i.
\end{equation}

Because input points closer to the target node generally provide richer local geometric information, their predictions are expected to be more reliable. Therefore, in the point-to-point voting scheme, the predicted heatmap value $\hat{H}_{i,j}$ is used as a confidence score for $\hat{\bm y}^{i}_j$ and only the $K$ points with the highest confidence scores are retained for the $j$-th node. 
The final node estimation $\hat{\bm y}_{j}$ is then obtained via a confidence-weighted aggregation:
\begin{equation}
\hat{\bm y}_{j} = \left( \sum_{i \in \mathcal{K}} \hat{H}_{i,j}\,\hat{\bm y}^i_{j} \right)/\sum_{i \in \mathcal{K}} \hat{H}_{i,j},
\end{equation}
where $\mathcal{K}$ denotes the set of indices corresponding to the selected $K$ highest-confidence points.

This point-to-point voting branch is supervised using the ground-truth heatmap and offset vectors defined in Eq.~\ref{eq:Hij} and Eq.~\ref{eq:Uij}, with the training objective:
\begin{equation}
\begin{aligned}
    \mathcal{L}_{\rm vot}= 
    \frac{1}{N} \sum_{j=1}^M \sum_{i=1}^N & \left[ (\hat{H}_{i,j} - H_{i,j}^*)^2 + \Vert\hat{\bm U}_{i,j} - \bm U_{i,j}^*\Vert^2 \right]. 
\end{aligned}
\end{equation}

The direct regression and point-to-point voting branches share the same PointNet++ encoder and are jointly optimized with the overall loss:
\begin{equation}\label{eq:detection_loss}
\begin{aligned}
    \mathcal{L}_{\rm tot} =  \lambda_{\rm reg}\mathcal{L}_{\rm reg} +  \lambda_{\rm vot}\mathcal{L}_{\rm vot}.
\end{aligned}
\end{equation}

This point-to-point voting scheme yields highly precise estimations when the local neighborhood of the target node contains sufficient input points, demonstrating its effectiveness in capturing fine-grained geometric information. However, its performance inherently degrades under occlusions (see Fig.~\ref{fig:detection_occlusion}): when too few input points are available near an occluded node, the lack of informative local geometry leads to significantly inaccurate predictions for the invisible portions.

% This point-to-point voting scheme achieves precise state estimations when the local region of the target node contains sufficient input points, indicating its effectiveness in learning local geometric information, which is essential for fine-grained estimation. However, its architecture inherently limits performance under heavy occlusions: when too few input points are available in the local neighborhood of a node, the prediction for the occluded part becomes significantly inaccurate due to the lack of informative local geometry.

% focuses on local regions and learns the local information for precise estimation well. 
% However, 
% However, when heavy occlusion occurs, there are even no valid input points in the neighborhood around some feature points, resulting that the heatmap values of all input points are close to zero. 
% In these cases, the estimated feature points in the occluded part of input point cloud are unreliable and might be far away from the true positions.

\subsection{Diffusion-Based Fusion}
As discussed above, the regression branch is globally robust under occlusions but locally inaccurate, whereas the voting branch is locally precise but highly sensitive to partial observations. To combine these complementary strengths, Lv et al.~\cite{lyu2023learning} first identify visible regions based on node-wise confidence scores and estimate a non-rigid transformation from the unoccluded regression nodes to their corresponding voting nodes. 
Although applying this transformation to the regression nodes can recover a plausible global configuration, this fusion process relies heavily on manually tuned registration parameters and remains fragile under large-scale occlusions or when generalizing to DLOs with diverse physical properties.
In this paper, we adopt a learning-based generative formulation that captures the complex high-dimensional distribution of ground-truth states, allowing the model to \textit{infer} complete configurations from the two-branch predictions. Specifically, the final state estimation is generated through a denoising diffusion process conditioned on both branches' results, enabling end-to-end learning from large-scale data and yielding more accurate, robust, and generalizable fusion without the need for handcrafted registration parameter tuning.

% In this paper, we advance the fusion module by adopting a generative formulation to capture the complex high-dimensional mapping, where final state estimation is modeled as a denoising diffusion process conditioned on the predictions of both branches.

Specifically, the two-branch fusion process is formulated as fitting a conditional probability distribution $p(\bm{Y}\,|\,\bm{\hat{Y}}_{\rm reg},\bm{\hat{Y}}_{\rm vot})$, where the regression prediction $\bm{\hat{Y}}_{\rm reg}$ and the voting prediction $\bm{\hat{Y}}_{\rm vot}$ serve as conditions to guide the denoising process. Following the standard denoising diffusion probabilistic model (DDPM) \cite{ho2020denoising}, Gaussian noise is progressively added to the ground-truth sample $\bm Y^0$ drawn from real distribution $q(\bm Y)$ during the forward process, while a conditional denoising model $\epsilon_\theta$ is trained to iteratively reconstruct the noise-free DLO nodes ${\bm Y}^0$ in the reverse process. The forward diffusion process is defined as:
\begin{equation}
    q(\bm Y^k|\bm Y^{k-1}) = \mathcal{N}(\bm Y^k;\sqrt{1-\beta^k}\,\bm Y^{k-1},\,\beta^k\bm I).
    \label{eq:1}
\end{equation}

% \begin{figure} [tb]
%   \centering 
%     \includegraphics[width=8.85cm]{figs/fusion.pdf} 
% %   \vspace{-3mm}
%   \caption{Illustration of the diffusion-based fusion module. 
%   The nodes estimated by regression (blue points) are always globally smooth but imprecise, whereas voting results are locally precise but unreliable inside the occluded region (green points). Conditioned on the coarse estimations from both branches, a diffusion-based generative model fuses their outputs to obtain the final node sequence (purple points).}
%   \label{fig:fusion}
%     \vspace{-2mm}
% \end{figure}

\begin{figure*} [tb]
  \centering 
    \includegraphics[width=\linewidth]{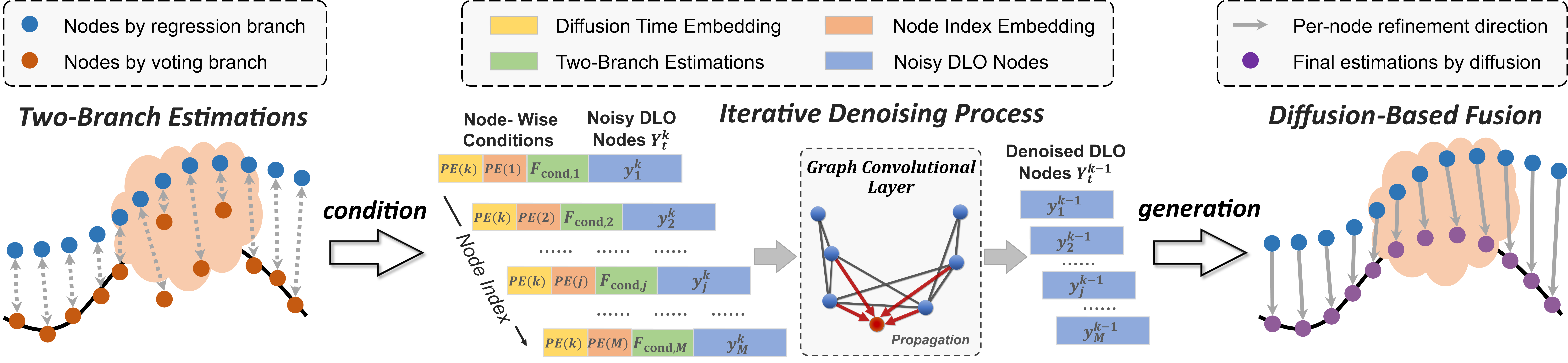} 
  \vspace{-2mm}
  \caption{Illustration of the diffusion-based fusion module. 
  The nodes estimated by regression (blue points) are always globally smooth but imprecise, whereas voting results (orange points) are locally precise but unreliable inside the occluded region. Conditioned on the coarse estimations from both branches, a diffusion-based generative model incorporated with graph convoluntional layer fuses their outputs to obtain the final node sequence (purple points).}
  \label{fig:fusion}
    \vspace{-2mm}
\end{figure*}

Given the variance $\beta^k\in[0,1]$ predefined by a noise scheduler, the sampling process of noisy sample $\bm Y^k$ can be simply rewritten as:
\begin{equation}
    q(\bm Y^k|\bm Y^0) = \mathcal{N}(\bm Y^k;\sqrt{\Bar{\alpha}^k}\bm Y^{0},(1-\Bar{\alpha}^k)\bm I),
    \label{eq:2}
\end{equation}
where ${\alpha}^k=1-\beta^k$ and $\Bar{\alpha}^k=\prod _{s=1}^k{\alpha}^s$.

% = p(\bm Y^{K})\prod_{k=1}^K p_\theta(\bm Y^{k-1}|\bm Y^k)

To perform the reverse process, the posterior $q(\bm Y^{k-1}|\bm Y^k)$ is approximated by a neural network to learn $p_\theta(\bm Y^{k-1}|\bm Y^k)$. Consequently, the joint distribution of total samples $p_\theta(\bm Y^{0:K})$ can be expressed by a series of learned Gaussian distributions:
\begin{equation}
\begin{aligned}
   p_\theta(\bm Y^{k-1}|\bm Y^k) = \mathcal{N}(\bm Y^{k-1};\mu_{\theta}(&\bm Y^k,\bm{\hat{Y}}_{\rm reg},\bm{\hat{Y}}_{\rm vot}, k),\\
   &\Sigma_{\theta}(\bm Y^k, \bm{\hat{Y}}_{\rm reg},\bm{\hat{Y}}_{\rm vot}, k)),
   \end{aligned}
\end{equation}
where the mean $\mu_{\theta}$ and variance $\Sigma_{\theta}$ are predicted by the neural network parameterized by $\theta$. Using Bayes' theorem, the true posterior $q(\bm Y^{k-1}|\bm Y^k, \bm Y^0)$ admits a closed-form Gaussian distribution:
\begin{equation}\label{eq:sampling}
   q(\bm Y^{k-1}|\bm Y^k, \bm Y^0) = \mathcal{N}(\bm Y^{k-1};\Tilde{\mu}^k(\bm Y^k, \bm Y^0),\Tilde{\beta}^k \bm I),
\end{equation}
with $\Tilde{\mu}^k(\bm Y^k, \bm Y^0)=\frac{\sqrt{\alpha^k}(1 - \bar{\alpha}^{k-1})}{1 - \bar{\alpha}^k} \bm Y^k + \frac{\sqrt{\bar{\alpha}^{k-1}}\beta^k}{1 - \bar{\alpha}^k} \bm Y^0$, and $\Tilde{\beta}^k={\frac{1 - \bar{\alpha}^{k-1}}{1 - \bar{\alpha}^k} \beta^k}$. The diffusion model can therefore be trained by minimizing the KL divergence between the two distributions above.

To incorporate local information interaction between neighboring nodes into the denoising network, we adopt a graph convolutional network (GCN)–style architecture instead of the U-Net \cite{ronneberger2015u} commonly used in DDPM. 
Since the DLO can be naturally represented as a sequential chain, we construct a graph $\mathcal{G}=(\mathcal{V},\mathcal{E})$, where the vertices $\mathcal{V}$ correspond to the DLO nodes and the edges $\mathcal{E}$ connect the pair of nodes whose distance falls below a threshold. This formulation explicitly capturing the spatial connectivity of nodes and enables effective propagation and aggregation of features on the graph structure.
Given the regression prediction $\bm{\hat{Y}}_{\rm reg}\in \mathbb{R}^{M\times3}$ and the voting prediction $\bm{\hat{Y}}_{\rm vot}\in \mathbb{R}^{M\times3}$, we first enhance the node-wise features by applying a multi-head self-attention (MHSA) \cite{vaswani2017attention} layer across all nodes to gather global contextual information:
\begin{equation}
   \bm F_{\rm cond}=[{\rm MHSA}(\bm{\hat{Y}}_{\rm reg}), \,{\rm MHSA}(\bm{\hat{Y}}_{\rm vot})],
\end{equation}
where $[\cdot,\cdot]$ denotes feature concatenation, and sinusoidal positional embeddings of the 3-D node coordinates are included.

% \begin{figure} [tb]
%   \centering 
%     \includegraphics[width=8.85cm]{figs/gcn.pdf} 
% %   \vspace{-3mm}
%   \caption{Denoising process.}
%   \label{fig:gcn}
%     \vspace{-2mm}
% \end{figure}

Subsequently, the noisy sample $\bm Y^{k}$ from the previous step, the denoising step embedding, and the node index embedding are concatenated with the node-wise features $\bm F_{\rm cond}$. For the $j$-th node, the corresponding $j$-th row of $\bm F_{\rm cond}$ is updated via a node-wised shared MLP:
\begin{equation}\label{eq:pe}
   \bm F^{'}_{{\rm cond},j}= {\rm MLP}_{\rm denoise}([\bm F_{{\rm cond},j},\, \bm y^{k}_j,PE(k), PE(j)]),
\end{equation}
where $PE(k)$ and $PE(j)$ denote the sinusoidal positional embedding of denoising step $k$ and the node index $j$, respectively.

The updated features $\bm F^{'}_{\rm cond}\in\mathbb{R}^{M\times d}$ are then processed by a graph convolutional layer to aggregate information from neighboring nodes. With the affinity matrix $\bm A\in\mathbb{R}^{M\times M}$, which encodes spatial connectivity, and a learnable weight matrix $\bm W\in\mathbb{R}^{d\times d}$, the node-wise features are updated as:
\begin{equation}\label{eq:gcn}
   \hat{\bm F}_{\rm cond}=\sigma(\bm A \bm F^{'}_{\rm cond}\bm W),
\end{equation}
where $\sigma$ denotes a non-linear activation function such as ReLU. In practice, we stack three such layers to sufficiently capture the complex denoising mapping. A node-wise MLP finally predicts the denoised sample $\hat{\bm Y}^{0}$ at the current step of the denoising process. The entire denoising network can therefore be written as: 
\begin{equation}
  \hat{\bm Y}^{0} = \mu_{\theta}(\bm Y^k,\bm{\hat{Y}}_{\rm reg},\bm{\hat{Y}}_{\rm vot}, k).
\end{equation}

During inference, the iterative sampling process follows Eq.~\ref{eq:sampling} and is given by
\begin{equation}\label{eq:diffusion_sampling}
   \bm Y^{k-1} = \frac{\sqrt{\Bar{\alpha}^{k-1}}\beta^k}{1-\Bar{\alpha}^k} \hat{\bm Y}^{0}+\frac{\sqrt{\alpha^{k}}(1-\Bar{\alpha}^{k-1})}{1-\Bar{\alpha}^k}\bm Y^k + \sqrt{\Tilde{\beta}^k}\,z,
\end{equation}
where $z\sim \mathcal{N}(\bm 0,\bm I)$. The diffusion model is trained with the regression and voting branches kept frozen, using the following supervision objective:
\begin{equation}\label{eq:diffusion_training}
   \mathcal{L}_{\rm diff} = \mathbb{E}_{\bm{Y}^*,k,\epsilon} \left[ \Vert\hat{\bm Y}^{0} - \bm{Y}^*\Vert^2\right].
\end{equation}

\section{Cross-Frame DLO State Tracking}\label{section:tracking}

The single-frame estimation method described above can be applied independently in each frame to provide current DLO state, but the temporal continuity and topological consistency can not be guaranteed. For instance, the overall DLO length should remain approximately constant, and the motions of spatially adjacent nodes across frames should be coherent. Under heavy occlusions, although single-frame estimation may still yield a smooth and plausible shape within the current frame, the inferred states in occluded regions can vary significantly between adjacent frames, severely limiting its application for closed-loop DLO manipulation.
Therefore, cross-frame state tracking is essential: by leveraging previous estimations as priors, the tracker can enforce temporal smoothness and preserve motion coherence, while using single-frame estimation only for initialization in the first frame.

\subsection{KNN-Based Feature Aggregation}
Given the previous-frame state $\bm Y_{t-1}$, the goal of the tracking algorithm is to accurately estimate the current state $\hat{\bm Y}_{t}$ from the new point cloud $\bm X_t$, while remaining robust even under heavy occlusions. We likewise formulate this tracking task as a conditional generation problem, similar to the single-frame setting, with both $\bm X_t$ and $\bm Y_{t-1}$ fed into the diffusion model. As discussed earlier, global features directly extracted from the DLO point cloud lack sufficient geometric details to distinguish individual nodes, making node-wise local features essential for precise predictions. 
In the single-frame case, no prior information on current node coordinates is available, so a two-branch network is employed to produce coarse estimates that then serve as conditions for denoising. In contrast, for cross-frame tracking, the previous state $\bm Y_{t-1}$ already provides a strong and typically close prior to the current configuration, enabling us to extract node-wise local conditions directly without requiring an additional coarse prediction stage.

\begin{figure} [tb]
  \centering 
    \includegraphics[width=0.98\linewidth]{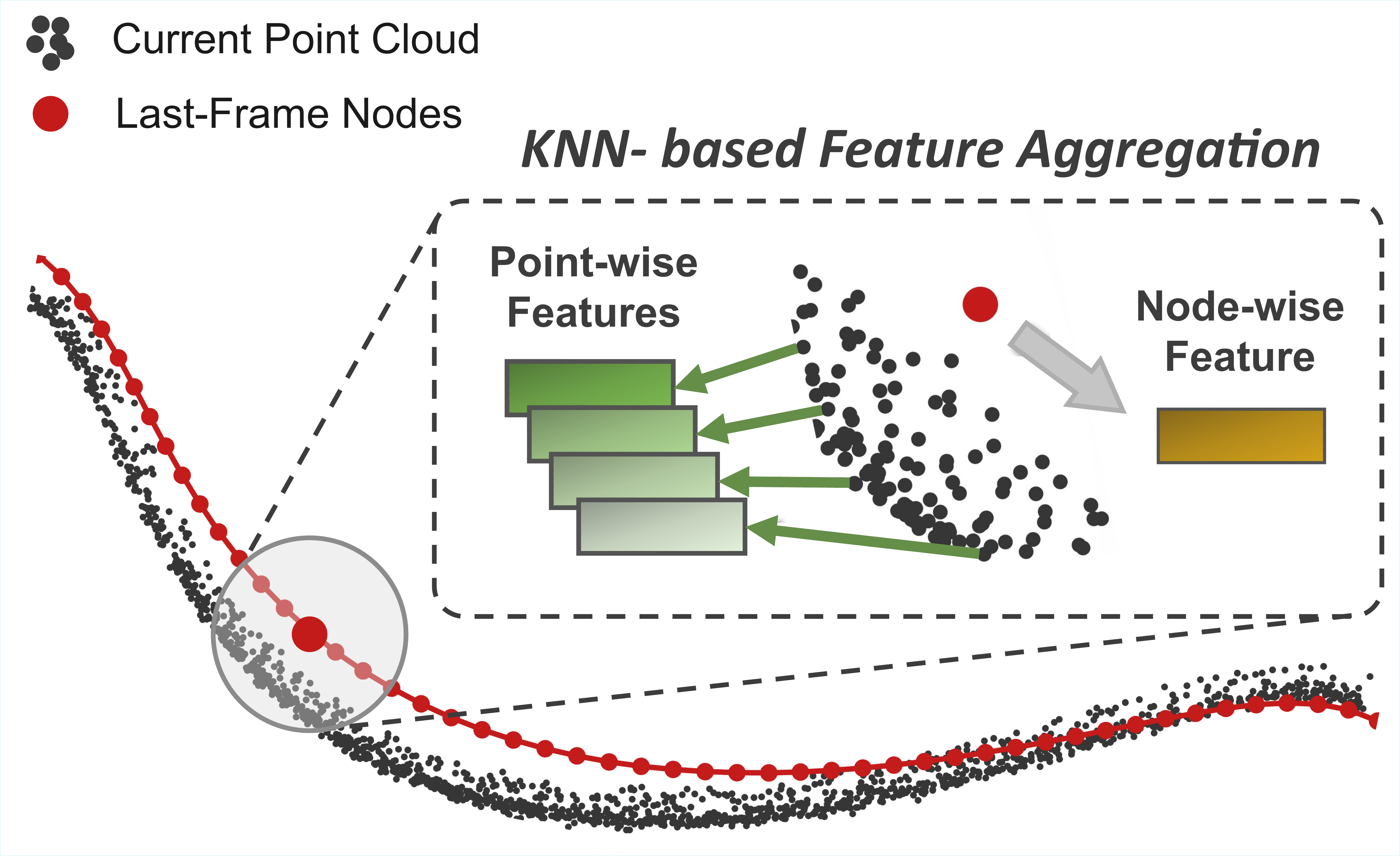} 
  \vspace{1mm}
  \caption{Demonstration of KNN-based feature aggregation module. Given point-wise features extracted by PointNet++, this module uses each previous-frame node as the sampling centroid and aggregates features from its local neighborhood to construct node-wise representations.}
  \label{fig:knn}
    \vspace{-1mm}
\end{figure}

With the PointNet++ backbone identical to that used in single-frame estimation, point-wise features $\bm F^{\rm point}_t \in \mathbb{R}^{N\times d}$ are first extracted from the current DLO point cloud $\bm X_t$. To incorporate temporal priors, we then construct node-wise local features by aggregating information around each node in the previous frame $\bm Y_{t-1}$ as shown in Fig.~\ref{fig:knn}. Specifically, for each node, we gather features from its K-nearest neighbors in $\bm X_t$, forming that node's representation.
This feature aggregation module follows the structure of the PointNet++ set abstraction layer, but with  the sampling step omitted and the previous-frame nodes directly treated as centroids for the subsequent grouping operation. This design effectively encodes fine-grained local geometry while leveraging $\bm Y_{t-1}$ as a strong and reliable prior, making the extracted features well-suited for precise motion prediction under occlusions. The resulting per-node features $\bm F^{\rm node}_t \in \mathbb{R}^{M\times d}$ are obtained as:
\begin{equation}
   \bm F^{\rm node}_t = {\rm KNNEncoder}(\bm F^{\rm point}_t, \bm Y_{t-1}),
\end{equation}
where ${\rm KNNEncoder}(\cdot)$ denotes this KNN-based feature aggregation module operating around the nodes $\bm Y_{t-1}$ from the previous frame.

\subsection{Diffusion-Based Motion Prediction}

Since the KNN-based aggregation module above captures only local geometric information, we further enhance the node-wise features using a Multi-Head Self-Attention (MHSA) layer to incorporate global contextual relationships among all nodes. This global receptive field enables the model to reason about long-range dependencies and overall DLO structure. The node-wise conditions for generative state tracking are given by:
\begin{equation}
   \bm F^{\rm cond}_t = {\rm MHSA}(\bm F^{\rm node}_t).
\end{equation}

The denoising network architecture for state tracking follows the same overall design as in single-frame estimation. Specifically, for the $j$-th node at denoising step $k$, the corresponding row of $\bm F^{\rm cond}_t$ is concatenated with the noisy sample $\bm y^{k-1}_{j,t}$, the denoising step embedding $PE(k)$, and the node index embedding $PE(j)$ (as in Eq.~\ref{eq:pe}). These concatenated features are then passed through several graph convolutional layers (Eq.~\ref{eq:gcn}), which explicitly encode the spatial connectivity and local interactions within the DLO structure. Finally, a node-wise MLP produces the denoised prediction $\hat{\bm Y}^0_{t}$ at the current step. Formally, the denoising network for state tracking is expressed as:
\begin{equation}
  \hat{\bm Y}^{0}_t = \mu_\theta(\bm Y^k_t,\bm F^{\rm cond}_t, \bm Y_{t-1}, k).
\end{equation}

The sampling and training procedures also follow the same formulation as in single-frame estimation (Eq.~\ref{eq:diffusion_sampling} and Eq.~\ref{eq:diffusion_training}). Under this generative formulation, the DLO state is iteratively refined across diffusion steps, conditioned on both the current-frame point cloud features and the previous-frame estimation. This enables the model to maintain temporal smoothness, enforce motion coherence, and recover accurate node coordinates even under substantial occlusions, ultimately supporting reliable long-horizon tracking in challenging scenarios.

\section{Pre- and Post-Processing}\label{section:pipeline}

Building on the proposed single-frame estimation and cross-frame tracking modules, a complete occlusion-robust DLO perception pipeline can then be established: single-frame estimation provides an accurate and reliable initialization, while the tracking module predicts node motions across frames. In this section, we introduce several pre-processing and post-processing techniques, including point cloud normalization and the utilization of known endpoint poses, to further enhance robustness and overall performance.

\subsection{Point Cloud Normalization}

\begin{figure} [tb]
  \centering 
  \includegraphics[width=\linewidth]{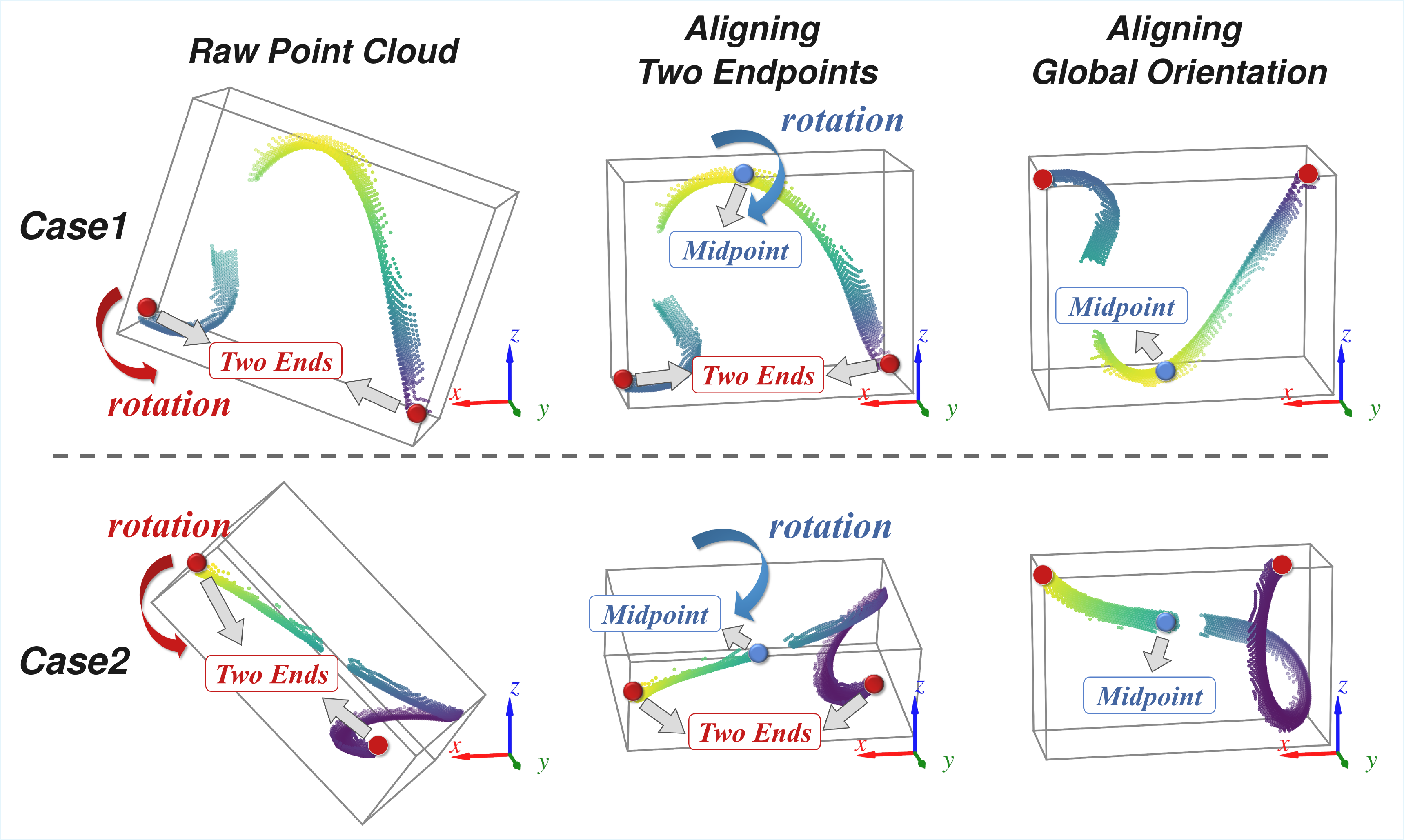} 
\vspace{-0.5cm}
\caption{Visualization of the DLO point cloud normalization process. First, the two endpoints are aligned by translating and rotating the raw point cloud so that one lies at the origin and the other on the $x$-axis. Then, the global orientation is further determined by rotating the DLO midpoint to lie on the $yOz$ plane, followed by scale and mean normalization.}
  \label{fig:normalization}
\end{figure}

To address the large variations of global DLO orientations, which significantly complicate the mapping from point cloud to state, we introduce a point cloud normalization strategy as a pre-processing step. This module aligns the DLO into a consistent global canonical orientation, reducing redundancy in orientation variance and enabling both the state estimation and tracking networks to be more orientation-robust.
For hand pose estimation, prior works \cite{ge2018hand, ge2018point} normalize hand point clouds using an oriented bounding box, whose axes are determined via principal component analysis (PCA) over the input coordinates.
In contrast, because DLO primarily deforms along a single dimension, its global orientation can be normalized more effectively and reliably by controlling the positions of its two endpoints and transforming the point cloud into a canonical coordinate system for our task.

The normalization process is illustrated in Fig.~\ref{fig:normalization}. Given the two endpoints of the DLO at timestep $t$, denoted as $\bm e^1_t$ and $\bm e^2_t$, we first translate and rotate the raw point cloud $X_t^{\rm raw}$ so that one endpoint is mapped to the origin and the other lies on the positive $x$-axis. Without loss of generality, we place $\bm e^1_t$ at the origin and align $\bm e^2_t$ with the $x$-axis.
While this step fixes the DLO along a canonical axis, the point cloud can still freely rotate around the $x$-axis. To eliminate this ambiguity and enforce a consistent global orientation across different initial poses, we further constrain the midpoint of the DLO to lie on the $yOz$ plane, which uniquely determines the remaining degree of rotational freedom. In practice, we approximate this midpoint by selecting the point whose $x$-coordinate is closest to the midpoint between the two endpoints, and then apply an additional rotation that places it on the $yOz$ plane. Let $\bm R^{\rm can}_t$ denote the overall normalization rotation matrix, comprising both the alignment of $\bm e^2_t$ to the $x$-axis and the subsequent midpoint alignment. The normalized point cloud in the canonical coordinate system is then given by:
\begin{equation}\label{eq:trans}
    \bm X_t^{\rm can}=(\bm X_t^{\rm raw} -\bm e^1_t)\cdot \bm R_t^{\rm can}.
\end{equation}

After the canonical transformation, we further normalize the point cloud by centering it to zero mean and scaling it by a size factor $L^{\rm can}_t$, defined as the maximum side length of the axis-aligned bounding box of $\bm X_t^{\rm can}$ along the $x$, $y$, and $z$ dimensions. The final normalized point cloud is computed as:
\begin{equation}
    \bm X_t=(\bm X_t^{\rm can}-\Bar{\bm X}_t^{\rm can}) /L^{\rm can}_t.
\end{equation}

Obviously, the accurate 3-D positions of the two DLO endpoints are not always available when performing the normalization in Eq.~\ref{eq:trans}. During training, we directly use the ground-truth terminal nodes, but obtaining reliable endpoint positions becomes a practical challenge at inference time. In certain manipulation scenarios, for example, when both ends of the DLO are grasped by robotic arms, the endpoint positions can be accurately retrieved from the manipulators' poses. Otherwise, for single-frame estimation, we can first infer a coarse global DLO state from the unnormalized point cloud to estimate the endpoints, and then re-run inference after transforming the point cloud into the canonical coordinate system. During tracking, the terminal nodes from the previous frame naturally serve as coarse endpoints for the current frame, since the inter-frame motion is typically not large.

% \begin{figure} [tb]
% \vspace{-0.2cm}
%   \centering 
%   % \hspace{-0.15cm}
%   \subfloat[raw point cloud]{ 
%     \includegraphics[width=0.14\textwidth]{figs/normalization/before_norm_seq_640_frame_132.png}
%   } 
%   \subfloat[normalize stage1]{ 
%     \includegraphics[width=0.14\textwidth]{figs/normalization/mid_norm_seq_640_frame_132.png}
%   } 
%   \subfloat[normalize stage2]{ 
%     \includegraphics[width=0.14\textwidth]{figs/normalization/after_norm_seq_640_frame_132.png}
%   } \\
%   \caption{Normalization visualization.}
%   \label{fig:normalization}
% \end{figure}

\subsection{State Post-Processing}

When accurate positions of the two DLO endpoints are available, we further exploit this information by redistributing the nodes produced by the network. Following common post-processing practices in prior works \cite{zhaole2023robust, kicki2023dloftbs}, we apply B-spline fitting to refine the node sequence. Concretely, several potentially unreliable nodes near each end (e.g., the closest three) are discarded, and the two known endpoints are appended to the remaining estimated nodes. A 3-D B-spline curve with a very small smoothness parameter is then fitted so that it closely interpolates both the endpoints and the retained nodes. Finally, $M$ uniformly spaced points are sampled along the curve, producing a sequence of DLO nodes that is smooth and precisely constrained to the known endpoints.

To mitigate the impact of large-scale occlusions and accumulated errors during cross-frame tracking, we further incorporate a tracking-failure detection mechanism and reinitialize the process when necessary. Under extreme occlusions, for example, when the DLO is almost entirely hidden by obstacles, the motion becomes invisible to the camera, and once visibility is restored, the tracked state may diverge significantly from the true configuration. Similarly, during long-term occlusions, accumulated prediction errors can destabilize iterative tracking and degrade accuracy.
To handle these cases, we monitor whether the per-frame node displacement exceeds a predefined threshold; if so, tracking failure is triggered. The single-frame estimation module is then invoked again to recover a reliable DLO state in the current frame, which serves as a new initialization for resuming cross-frame tracking. The overall UniStateDLO pipeline is summarized in Alg.~\ref{alg:pipeline}.

\begin{algorithm}[t]

\caption{The whole pipeline of UniStateDLO} 
\vspace{0.05cm}
\hspace*{0.02in} {\bf Input: } DLO Point Cloud $\bm X_t$\\
\hspace*{0.02in} {\bf Output: } DLO Nodes $\bm \hat{Y}_t$
{\setlength{\baselineskip}{1.2\baselineskip}
\begin{algorithmic}[1]
\If{$t=0$ \textbf{or} \textit{Re-init}} \hfill $\triangleright$ \textbf{Single-Frame Estimation}
\State $\bm X_t\leftarrow\text{Normalize}(\,\bm X_t\,)$
\State $\bm{\hat{Y}}^{\rm reg}_t,\,\bm{\hat{Y}}^{\rm vot}_t \leftarrow \text{Regression}(\bm X_t),\,\text{Voting}(\bm X_t)$
\State $\bm{\hat{Y}}_t \leftarrow \text{DiffFusion}(\bm{\hat{Y}}^{\rm reg}_t, \bm{\hat{Y}}^{\rm vot}_t)$
\State $\bm{\hat{Y}}_t \leftarrow\text{Denormalize}(\,\bm{\hat{Y}}_t\,)$
\Else{}  \hfill $\triangleright$ \textbf{Cross-Frame Tracking}
\State $\bm X_{t}, \,\bm{\hat{Y}}_{t-1}\leftarrow\text{Normalize}(\,\bm X_{t}, \,\bm{\hat{Y}}_{t-1}\,)$
\State $\bm F_{t}^{\rm node}\leftarrow\text{KNNEncoder}(\bm X_{t}, \,\bm{\hat{Y}}_{t-1})$
\State $\bm{\hat{Y}}_t \leftarrow \text{DiffTrack}(\bm F_{t}^{\rm node})$
\If{$\bm{\hat{Y}}_t-\bm{\hat{Y}}_{t-1}>T$} \hfill $\triangleright$  \textbf{Tracking Failure}
\State \text{ \textit{Re-init} by re-running single-frame estimation}
\Else \State $\bm{\hat{Y}}_t \leftarrow\text{Denormalize}(\,\bm{\hat{Y}}_t\,)$
\EndIf
\EndIf
\State $\bm{\hat{Y}}_t \leftarrow\text{PostProcess}(\,\bm{\hat{Y}}_t\,)$
\end{algorithmic}
\label{alg:pipeline}
}
\end{algorithm}

\section{Implementation Details}
\subsection{Large-Scale Training Data Synthesis}
We construct a large-scale synthetic DLO point cloud dataset entirely in simulation and train our network to exclusively on this data. Avoiding the costly and time-consuming data collection and annotation process in real-world scenarios, our model can be transferred directly from simulation to diverse real-world DLOs, demonstrating strong generalization capability. The dataset is generated using the Unity3D engine \cite{unity} in combination with the Obi Rope package \cite{obi}, which provides a particle-based physics model that represents DLOs as chains of oriented particles subject to stretching, bending, and twisting constraints.

\begin{figure} [tb]
  \centering 
  \includegraphics[width=8.75cm]{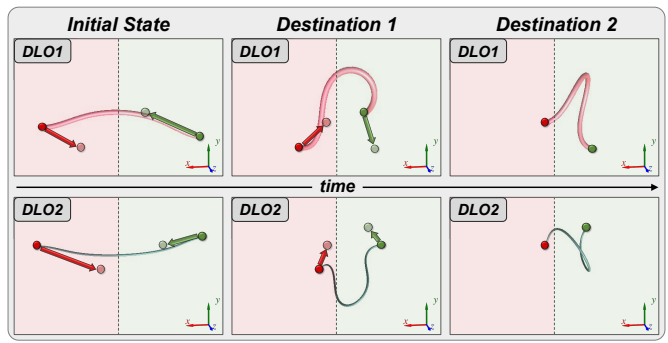} 
% \vspace{-0.2cm}
\caption{Illustration of the data collection process in simulation. The red and green regions denote the workspaces of the left and right endpoints. Starting from an almost straight configuration, a random target is sampled for each endpoint within its workspace, and both endpoints move toward their targets at constant velocity. Once reached, new targets are sampled and the process repeats until the sequence ends.}
  \label{fig:collection}
\end{figure}

To comprehensively capture a wide variety of configurations in the dataset, the simulated DLO is continuously manipulated by two grippers rigidly grasping its endpoints with a randomized motion strategy. As shown in Fig.~\ref{fig:collection}, the workspace of the two grippers is divided by a central vertical plane. At the beginning of each motion interval, a random target pose is uniformly sampled within each gripper's workspace, with orientations restricted to a feasible range. Both grippers then move smoothly toward their targets at constant velocities, generating diverse deformations while avoiding tangling or overstretching. Once the targets are reached, new destinations are sampled and executed repeatedly until the sequence ends.

During simulation, we record the RGB-D observations $\mathcal{I}_t, \mathcal{D}_t$ from a front-view camera, together with the corresponding ground-truth 3-D particle positions $Y_t$ at each simulation step. To further reduce the sim-to-real gap and improve generalization, we apply domain randomization by varying the camera poses and DLOs' physical parameters, including length (0.2m$\,\sim\,$1.0m), radius (2.5mm$\,\sim\,$10mm), and stiffness. In total, 1000 randomized DLO sequences are generated, each containing 300 consecutive frames, yielding a large-scale synthetic dataset of 300K frames. The sequences are randomly split into 80\% for training and 20\% for validation.

\subsection{Model Architecture and Training Settings}
The input point cloud is first downsampled to $N = 1024$ points using the farthest point sampling (FPS) algorithm, and then fed into our model, which is trained to predict $M = 50$ DLO nodes. The PointNet++ encoder \cite{qi2017pointnet++} contains 4 point set abstraction layers followed by 4 feature propagation layers, producing point-wise features of dimension $d = 256$.
For the point-to-point voting scheme, the neighborhood radius is set to $r = 0.02$, and each node aggregates estimations from the top $K = 64$ points. 
In the diffusion-based module, used for both single-frame estimation and tracking, the diffusion step and node index embeddings are each set to a dimension of 128. The denoising process is executed over 100 timesteps with a cosine noise scheduler, and a 10-step DDIM sampler \cite{song2021denoising} is employed during inference to accelerate sampling.

To simulate realistic occlusions and obtain partial point clouds, we randomly remove regions from the DLO segmentation masks and then project the remaining pixels from the RGB-D images $\mathcal{I}_t$ and $\mathcal{D}_t$ into 3-D space. To further emulate sensor imperfections, Gaussian noise is added as random jitter.
The model is trained with a batch size of 128 on a single NVIDIA RTX 4090 GPU. For single-frame state estimation, we first train the two-branch network for 200 epochs using the Adam optimizer with a learning rate of 0.01. The two branches are then frozen, and the diffusion-based fusion module is trained for an additional 300 epochs using AdamW with a learning rate of $1\times10^{-4}$.
For cross-frame tracking, the entire network is trained end-to-end for 300 epochs using AdamW with the same learning rate. The cosine learning rate scheduler is applied across all training stages.

\section{Simulation Results}
% In this section, we present extensive experiments to evaluate the proposed DLO state estimation and tracking algorithm in simulated environments. We begin by describing the data collection process for dataset construction, followed by quantitative comparisons of our method with state-of-the-art (SOTA) baselines under varying levels of occlusion. Finally, the zero-shot real-world performance and integration of our method into a closed-loop DLO manipulation task in constrained environments will be demonstrated in the next section.

\subsection{Evaluation Metrics}

Three evaluation metrics employed for a comprehensive comparison in simulation are introduced as follows:

\subsubsection{Mean Per-Node Error (MPNE)}
To evaluate overall accuracy, we compute the mean Euclidean distance between the estimated and ground-truth 3-D node positions, defined as
\begin{equation}
    {MPNE}=\frac{1}{M}\sum_{j=1}^M \Vert \hat{\bm y}_{j}-{\bm y}^*_{j}\Vert.
\end{equation}

\subsubsection{Percentage of Correct Node (PCN)}Inspired by the Percentage of Correct Keypoints (PCK) metric in pose estimation, we define a node as correct if its estimation error is within $T_{\rm dlo}=10$ mm. This metric is given by
\begin{equation}
    PCN=\frac{1}{M}\sum_{j=1}^M \,\delta(\hspace{0.1mm}\Vert \hat{\bm y}_{j}-{\bm y}^*_{j}\Vert<T_{\rm dlo}\hspace{0.1mm}).
\end{equation}

\subsubsection{Node Sequence Smoothness (NSS)}Following \cite{zhaole2023robust}, we also measure the physical smoothness of the estimated node sequence with the mean of squared angles between adjacent nodes, defined as
\begin{equation}
    {NSS}=\frac{1}{M-2}\sum_{j=2}^{M-1} \arccos \left(\frac{(\hat{\bm y}_{j}-\hat{\bm y}_{j-1})\cdot(\hat{\bm y}_{j+1}-\hat{\bm y}_{j})}{\Vert\hat{\bm y}_{j}-\hat{\bm y}_{j-1}\Vert\,\Vert\hat{\bm y}_{j+1}-\hat{\bm y}_{j}\Vert}\right)^2.
\end{equation}

\begin{table*}
\centering
\caption{Quantitative comparison of UniStateDLO and baselines for single-frame estimation under different occlusion levels.}
\renewcommand{\arraystretch}{1.1}
\setlength\tabcolsep{3pt}
\begin{tabular}{p{3.8cm}|p{0.98cm}<{\centering}p{0.95cm}<{\centering}p{0.95cm}<{\centering}|p{0.98cm}<{\centering}p{0.95cm}<{\centering}p{0.95cm}<{\centering}|p{0.98cm}<{\centering}p{0.95cm}<{\centering}p{0.95cm}<{\centering}|p{0.98cm}<{\centering}p{0.95cm}<{\centering}p{0.95cm}<{\centering}} 
\toprule
\multirow{2}{*}{Method} & \multicolumn{3}{c|}{No occlusion} & \multicolumn{3}{c|}{10\% occluded} & \multicolumn{3}{c|}{30\% occluded} & \multicolumn{3}{c}{50\% occluded} \\
 & MPNE$\,\downarrow$ & PCN$\,\uparrow$ &NSS$\,\downarrow$ & MPNE$\,\downarrow$ & PCN$\,\uparrow$ &NSS$\,\downarrow$ & MPNE$\,\downarrow$ & PCN$\,\uparrow$ &NSS$\,\downarrow$ & MPNE$\,\downarrow$ & PCN$\,\uparrow$ &NSS$\,\downarrow$\\ 
\midrule
DLOFTBs\cite{kicki2023dloftbs} & 8.62 & 86.07 & \textbf{0.0281} & 10.99 & 81.32 & 0.0323 & 17.16 & 72.48 & 0.0361 & 24.42 & 61.68 & 0.0405 \\
Sun et al.\cite{zhaole2023robust} & 10.13 & 85.50 & 0.0293 & 11.09 & 82.30 & 0.0321 & 13.65 & 75.98 & 0.0347 & 16.44 & 63.46 & 0.0363 \\
\midrule
Ours (\textit{Direct Regression}) & 12.54 & 50.15 & 0.0565 & 13.12 & 46.38 & 0.0561 & 15.27 & 38.29 & 0.0579 & 27.78 & 9.95 & 0.0687 \\
Ours (\textit{Point-to-Point Voting}) & 3.78 & 93.37 & 0.0449 & 4.32 & 91.68 & 0.1168 & 19.48 & 73.84 & 0.4508 & 39.18 & 52.02 & 0.8385 \\
\textbf{Ours (\textit{Diffusion-Based Fusion})} & \textbf{3.51} & \textbf{94.85} & 0.0314 & \textbf{3.58} & \textbf{93.80} & \textbf{0.0319} & \textbf{4.54} & \textbf{89.35} & \textbf{0.0327} & \textbf{9.29} & \textbf{72.46} & \textbf{0.0337} \\
\bottomrule
\end{tabular}
\label{tab:detection_results}
\end{table*}

\subsection{Single-Frame State Estimation}
\subsubsection{Comparison with State-of-The-Art Methods}
We first compare the single-frame estimation performance of UniStateDLO against two representative state-of-the-art (SOTA) baselines on the validation split of the synthesized dataset: a) \textbf{DLOFTBs} \cite{kicki2023dloftbs}, which estimates DLO shapes by fitting a B-spline from its 2-D skeletons, enabling the reconnection of disjoint segments under occlusions; b) \textbf{Sun et al.} \cite{zhaole2023robust} which reconstructs the DLO shape via Bézier curves defined by two control points, followed by Discrete Elastic Rod (DER) refinement to enhance smoothness. 
For a fair comparison with these training-free methods based on handcrafted strategies, we apply B-spline fitting as post-processing with the two ground-truth endpoints appended to ensure that all methods produce the same number of nodes.

\begin{figure} [tb]
\vspace{-0.3cm}
  \centering 
  \hspace{-0.8cm}
  \subfloat{ 
    \includegraphics[width=0.175\textwidth]{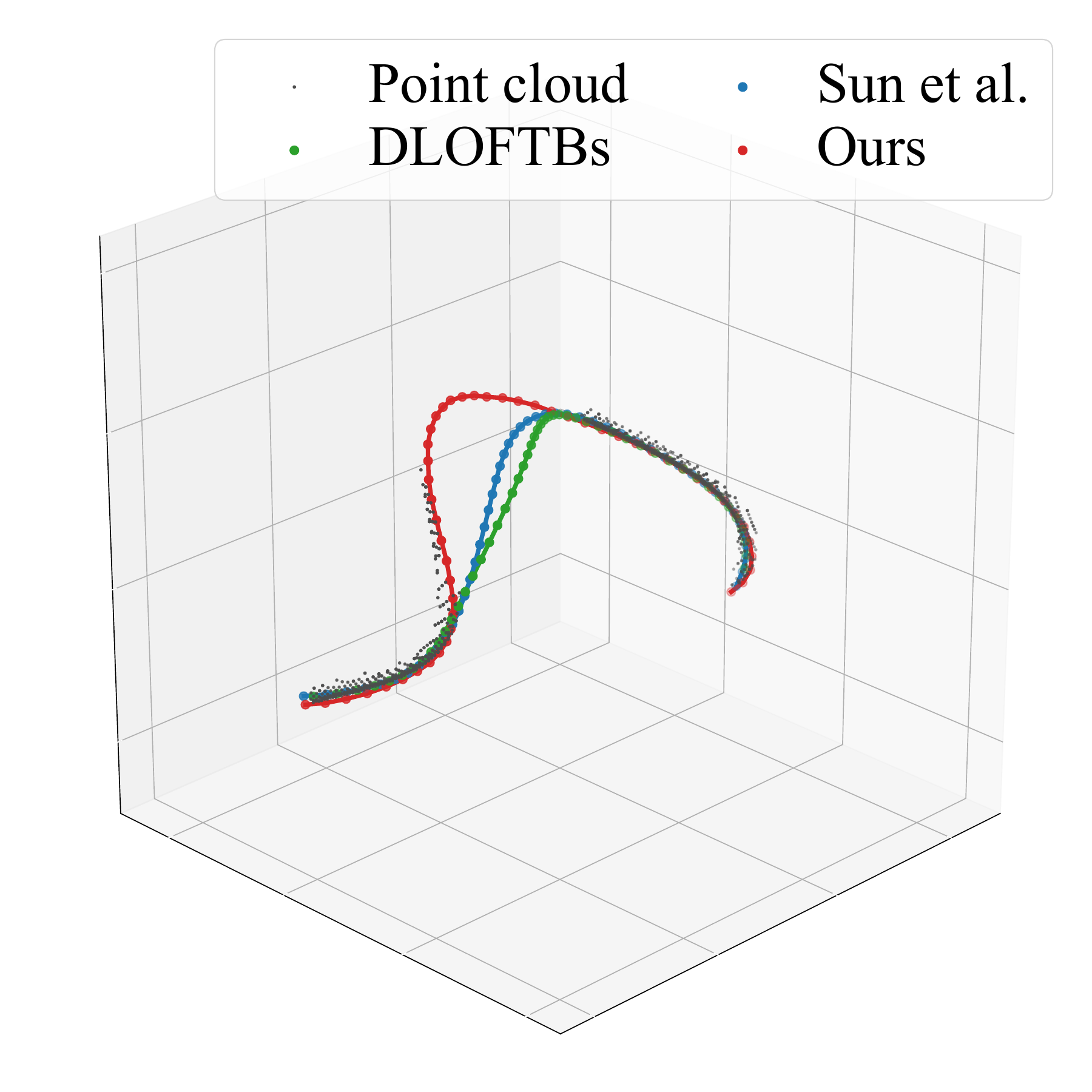}
    \label{fig:detection_baseline_1}
  } 
  \hspace{-0.485cm}
  \subfloat{ 
    \includegraphics[width=0.175\textwidth]{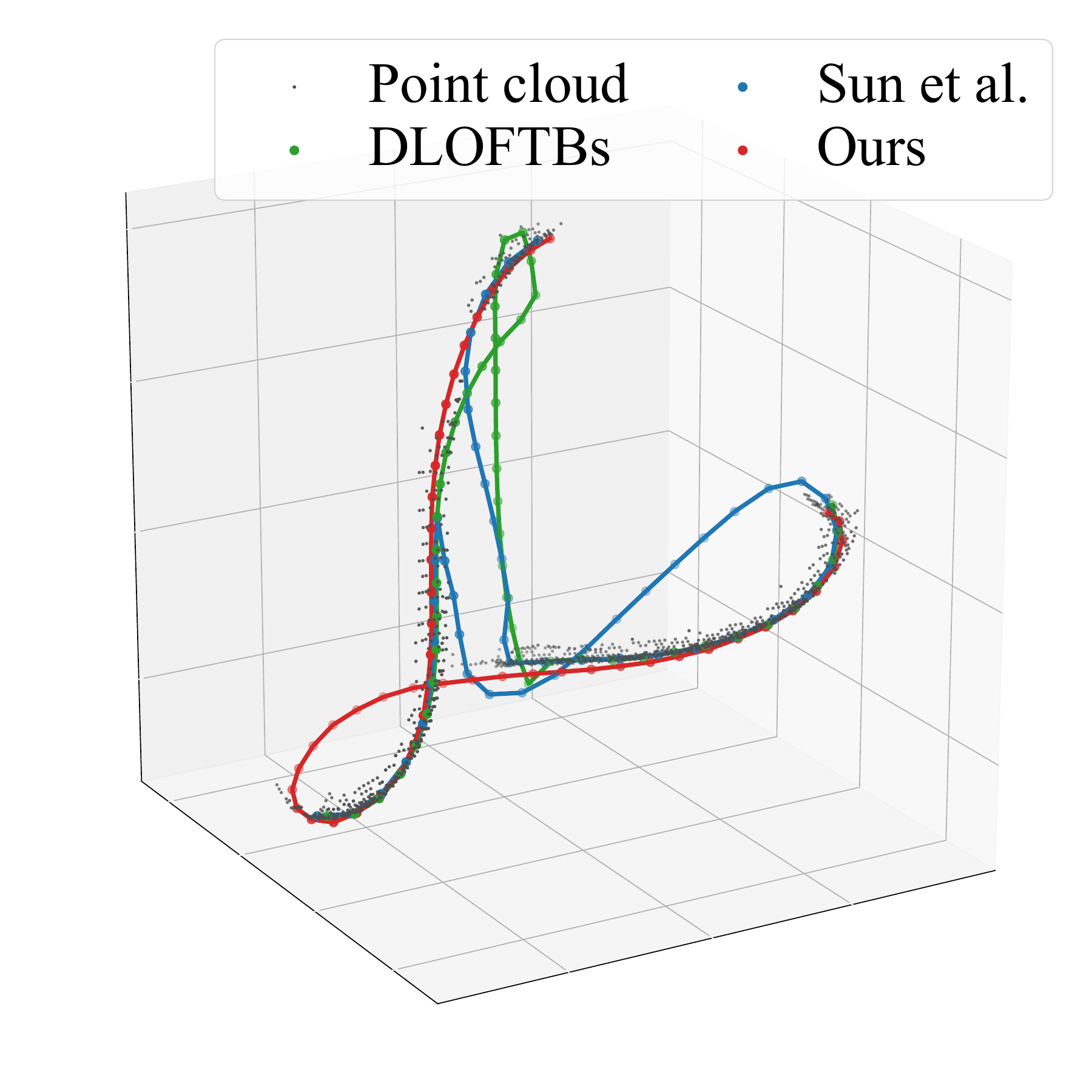}
    \label{fig:detection_baseline_2}
  } 
  \hspace{-0.485cm}
  \subfloat{ 
    \includegraphics[width=0.175\textwidth]{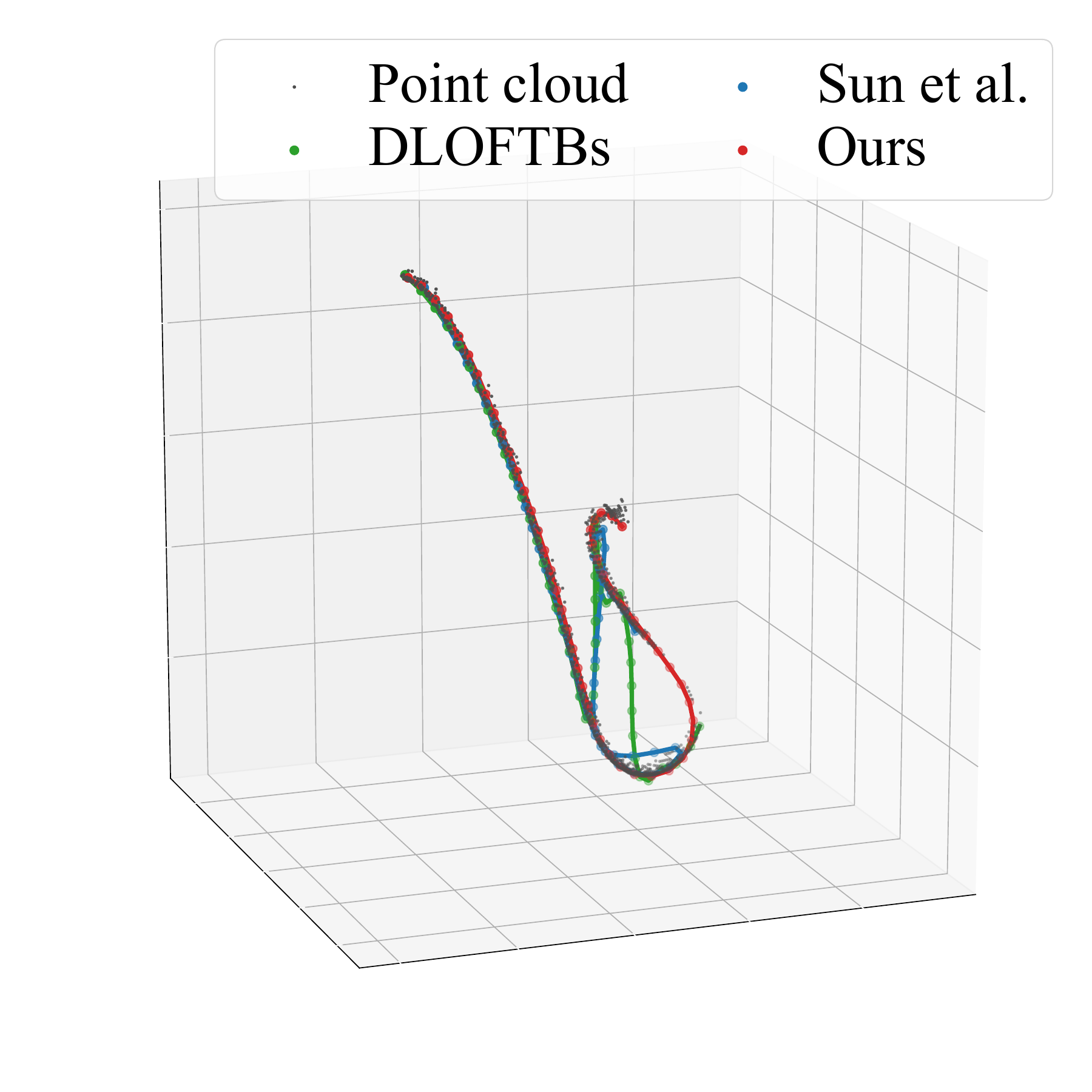}
    \label{fig:detection_baseline_3}
  } 
  \hspace{-0.6cm}
  \vspace{-0.05cm}
  \caption{Visualization of estimated DLO states produced by UniStateDLO, DLOFTBs and Sun et al. across several cases.}
  \label{fig:detection_baseline}
\end{figure}

Quantitative results under different occlusion levels are reported in Table~\ref{tab:detection_results}, where our full model is denoted as \textit{Diffusion-Based Fusion}. Occlusions are generated following the same procedure as in data collection by randomly masking regions in the RGB-D images. As shown in the table, UniStateDLO consistently achieves the best state estimation performance across nearly all occlusion levels and metrics, demonstrating strong capability to handle occlusions.
Notably, in the \textit{No occlusion} setting, frequent self-occlusions in complex DLO configurations still yield partial point clouds. Despite this, our model successfully predicts near 95\% of the nodes, whereas both baselines exhibit substantially higher errors. Since all outputs undergo a final B-spline refinement, the smoothness scores across methods remain comparable.
As occlusion severity increases, the baseline errors grow rapidly and their correct proportions drop sharply. In contrast, under heavy occlusions up to 50\%, UniStateDLO maintains a mean per-node error of 9.29 mm and correctly predicts 72.46\% of the nodes, significantly outperforming the baselines.

Several visualized examples are shown in Fig.~\ref{fig:detection_baseline}, where the estimated nodes are sequentially connected to visualize their ordering. Under light occlusions and relatively simple DLO configurations (Fig.~\ref{fig:detection_baseline_1}), both DLOFTBs and Sun et al. can roughly infer the occluded portions of the DLO, but their predictions remain noticeably less accurate than ours. When occlusions become severe or the DLO adopts more complex shapes (Fig.~\ref{fig:detection_baseline_2} and Fig.~\ref{fig:detection_baseline_3}), the two baselines frequently fail to recover correct connectivity between disjoint DLO segments from the 2-D masks, resulting in highly unreliable reconstructions. In contrast, UniStateDLO consistently produces accurate and occlusion-robust state estimations across all scenarios.

\begin{figure} [tb]
\vspace{-0.3cm}
  \centering 
  \hspace{-0.8cm}
  \subfloat[without occlusion]{ 
    \includegraphics[width=0.1725\textwidth]{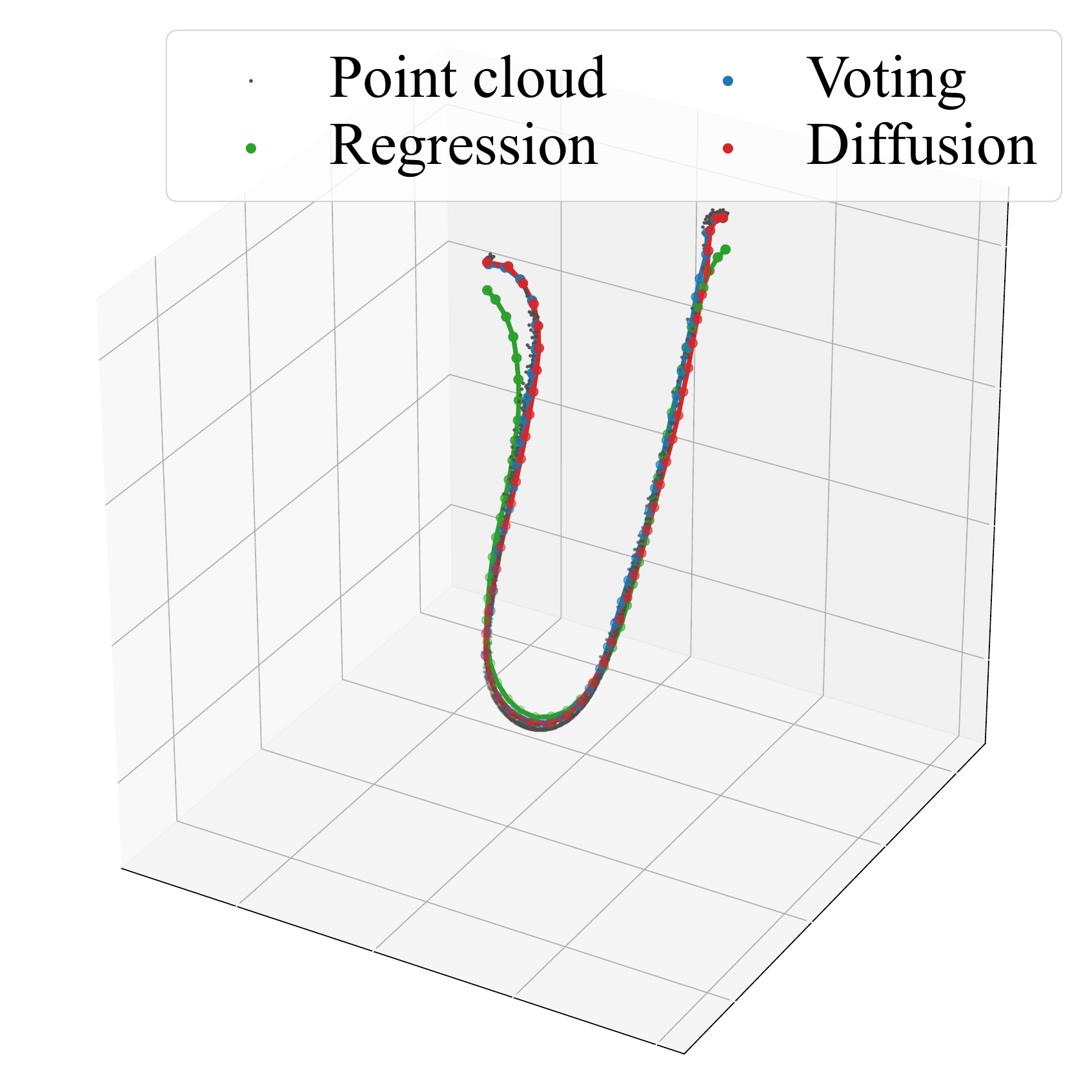}
  } 
  \hspace{-0.45cm}
  \subfloat[30\% occlusion]{ 
    \includegraphics[width=0.1725\textwidth]{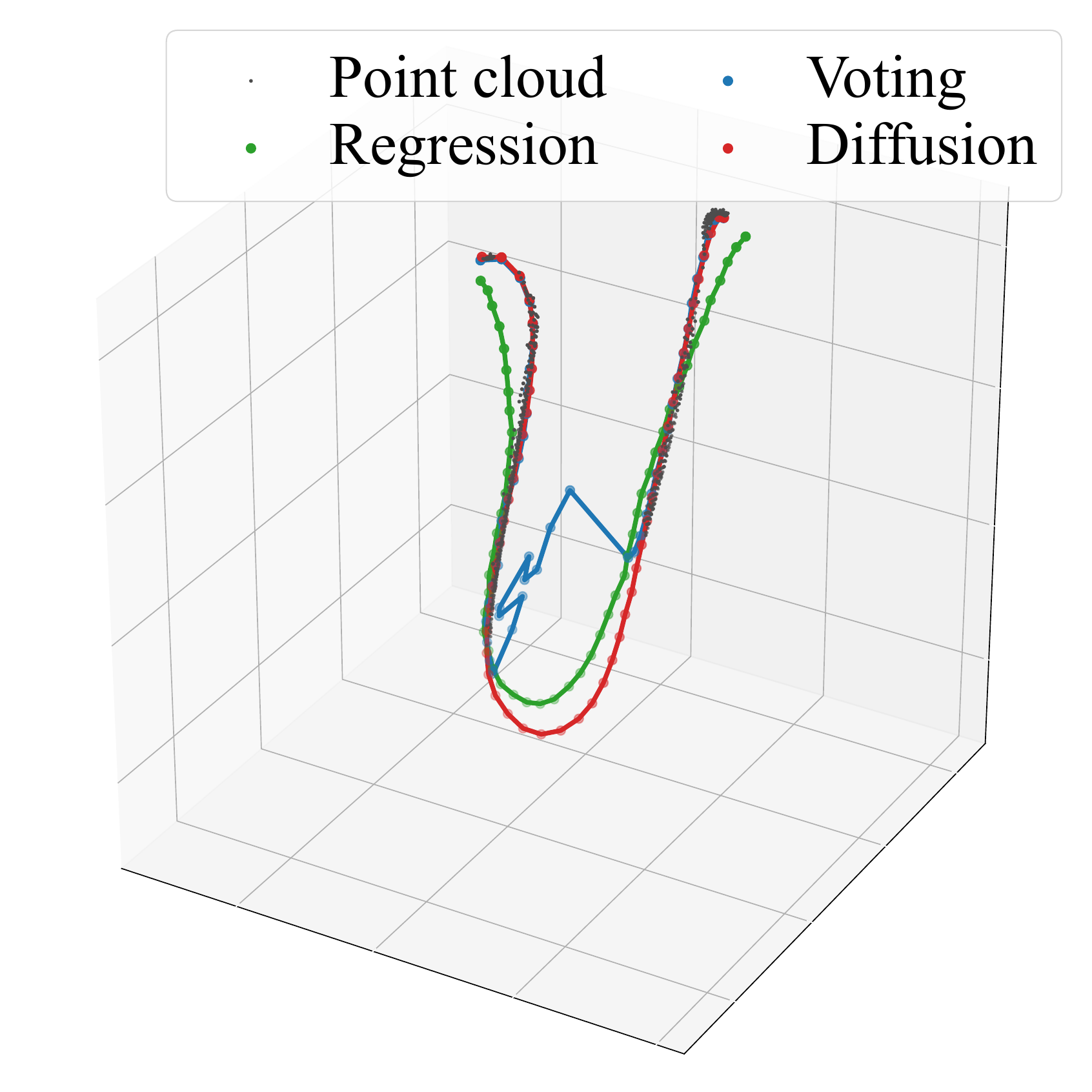}
  } 
  \hspace{-0.45cm}
  \subfloat[50\% occlusion]{ 
    \includegraphics[width=0.1725\textwidth]{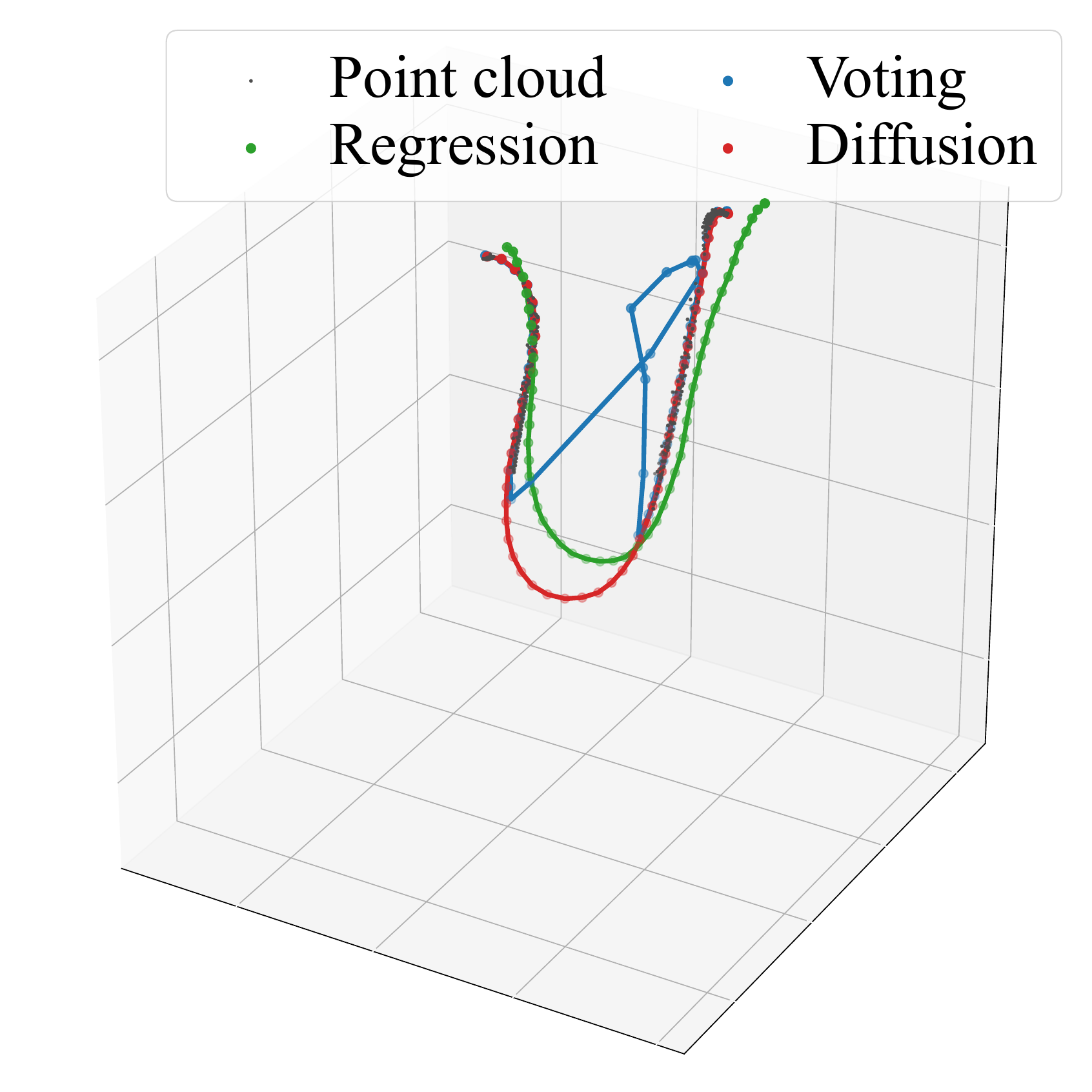}
    \label{fig:detection_occlusion_3}
  } 
  \hspace{-0.6cm}
  % \vspace{-0.1cm}
  \caption{Visualization of estimated DLO states produced by the regression branch, voting branch, and diffusion-based fusion module under different occlusion levels. }
  \label{fig:detection_occlusion}
\end{figure}

\subsubsection{Self-Comparisons}
We further conduct a quantitative comparison of the two intermediate branches, \textit{Direct Regression} and \textit{Point-to-Point Voting}, across different occlusion levels, with results summarized in Table~\ref{tab:detection_results}. The \textit{Direct Regression} branch exhibits a relatively large per-node error of 12.54 mm even in the absence of external occlusions, and its accuracy deteriorates steadily as the occlusion level increases. Nevertheless, as illustrated in Fig.~\ref{fig:detection_occlusion}, this branch remains highly robust: it consistently produces smooth and globally coherent shapes, even under severe occlusions, and effectively completes invisible segments of the DLO. 
In contrast, the \textit{Point-to-Point Voting} branch achieves excellent performance on complete point clouds (3.78 mm error) but experiences a drastic degradation under occlusions, reaching 39.18 mm error when 50\% of the DLO is occluded. While effectively leveraging local geometric features, this branch produces accurate estimations for visible regions but fails to infer reliable positions for occluded nodes due to missing local evidence.
Across all occlusion levels, our diffusion-based fusion method achieves the best overall accuracy and robustness by effectively combining the complementary strengths of the two branches. Even when most of the DLO is invisible from view (see Fig.~\ref{fig:detection_occlusion_3}), UniStateDLO still reconstructs a plausible and physically consistent global configuration from the highly incomplete point cloud observations.

To assess the contribution of the diffusion model to DLO state estimation, we first evaluate an \textit{End-to-End Diffusion} variant that removes the two-branch architecture and conditions the diffusion model solely on global features, with results shown in Fig.~\ref{fig:ablation}. Compared to the \textit{Direct Regression} branch, this variant replaces the MLP regression head with a diffusion-based generative module. The resulting improvements (MPNE of 9.26 mm vs.\ 12.54 mm without occlusion, and 11.75 mm vs.\ 15.27 mm under 30\% occlusion) clearly highlight the diffusion model's strong capability to capture the complex underlying distribution of DLO configurations.
However, despite these gains, the end-to-end diffusion variant still produces substantially higher per-node errors and lower overall accuracy than our full fusion-based model. This gap arises because relying solely on global features is insufficient for encoding the fine-grained local geometric cues required for accurate estimation from thin, textureless DLO point clouds, an issue similar to that observed in direct regression. Thus, the necessity of our proposed two-branch design is emphasized again: the coarse predictions provided by direct regression and point-to-point voting offer essential node-wise cues, which are then effectively fused by the diffusion model to overcome the limitations of each individual branch and achieve accurate and occlusion-robust DLO state estimations.

\subsubsection{Ablation Study}
\begin{figure} [tb]
\vspace{-0.25cm}
  \centering 
  \hspace{-0.43cm}
  \subfloat{ 
    \includegraphics[width=0.235\textwidth]{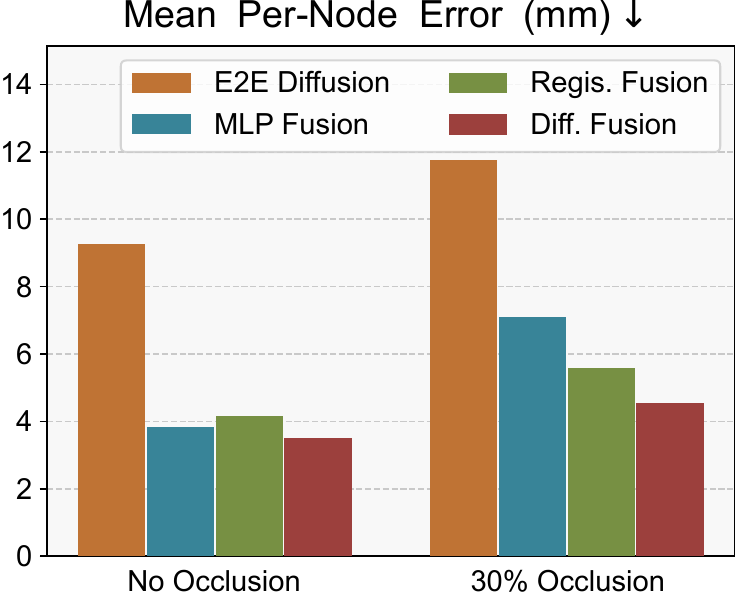}
  } 
  \hspace{-0.145cm}
  \subfloat{ 
    \includegraphics[width=0.235\textwidth]{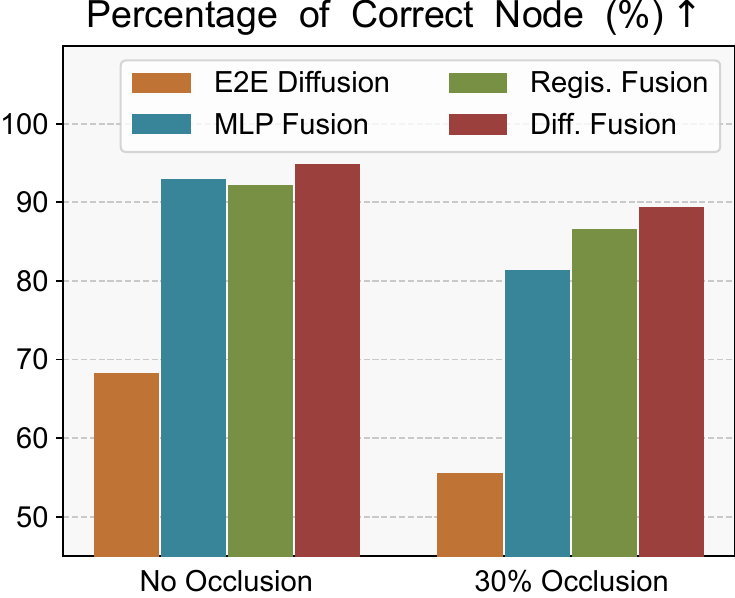}
  } 
  \hspace{-0.35cm}
  % \vspace{-0.2cm}
  \caption{Ablation on end-to-end diffusion and different two-branch fusion strategies. The left figure plots the MPNE metric under no occlusion and 30\% occlusion settings, while the right figure presents the PCN results. \textit{E2E Diffusion}: diffusion model conditioned on the global feature. \textit{MLP Fusion/Regis. Fusion}: fusion through an MLP or non-rigid point registration. \textit{Diff. Fusion}: diffusion-based fusion.}
  \label{fig:ablation}
\end{figure}

\begin{table}
\centering
\caption{Ablation study on different conditioning and denoising types for diffusion-based fusion module.}
\renewcommand{\arraystretch}{1.15}
\setlength\tabcolsep{3pt}
\begin{tabular}{c|p{1.9cm}<{\centering}p{1.1cm}<{\centering}|p{1.02cm}<{\centering}p{1.02cm}<{\centering}} 
\toprule
\makecell{Level of \\ occlusion} & \makecell{Cond. \\ Type} & \makecell{Denois. \\ Type} & MPNE$\,\downarrow$ & PCN$\,\uparrow$\\ 
\midrule
\multirow{4}{*}{No occlusion} & Global & Global & 3.98 & 92.94 \\
 & Global & Local & 4.05 & 92.73 \\
 &Local \textit{w/o GCN}& Local & 3.82 & 93.46 \\ 
 &Local \textit{w/ GCN}& Local & \textbf{3.51} & \textbf{94.85} \\
\midrule
\multirow{4}{*}{30\% occluded} & Global & Global & 5.26 & 86.95 \\
 & Global & Local & 5.45 & 87.44 \\
 &Local \textit{w/o GCN}& Local & 5.14 & 88.62 \\ 
 &Local \textit{w/ GCN}& Local & \textbf{4.54} & \textbf{89.35} \\
\bottomrule
\end{tabular}
\label{tab:ablation_conditioning}
\end{table}

Furthermore, we evaluate two alternative fusion strategies for the two-branch architecture, as illustrated in Fig.~\ref{fig:ablation}:
a) \textit{MLP Fusion}, which concatenates the regression and voting outputs and learns an MLP-based refinement mapping; and
b) \textit{Regis. Fusion} \cite{lyu2023learning}, which computes a non-rigid transformation aligning the visible voting estimations to the regression sequence via point-set registration, and then applies this transformation to refine the regression result.
In the absence of occlusion, the voting outputs are already highly accurate, so the MLP-based fusion primarily learns an identity mapping and performs slightly better than registration-based fusion. As occlusion increases, however, the MLP struggles to model the highly non-linear relationships for reliable refinement, whereas registration-based fusion demonstrates stronger robustness.
Across both occluded and unoccluded settings, our diffusion-based fusion consistently outperforms all ablated variants, underscoring the advantages of generative modeling to infer complete DLO states.

Different conditioning and denoising strategies within the diffusion model are also investigated, as summarized in Table~\ref{tab:ablation_conditioning}. Since the dimensionality of the DLO node sequence is relatively low, there are two ways to organize the denoising process: the node coordinates can be flattened into a 1-D vector, or preserved in their original 2-D structure, where the first dimension indexes individual nodes. We refer to these as the \textit{Global} and \textit{Local} denoising types, respectively.
Similarly, the two-branch estimations used as conditioning inputs can either be flattened or kept in their structured node-wise form, corresponding to \textit{Global} and \textit{Local} conditioning.
Experimental results show that the combination of local denoising and local conditioning, paired with a GCN module, delivers the best overall performance. This setting allows the denoising process to explicitly leverage the spatial structure of the DLO and incorporate rich contextual information among neighboring nodes, highlighting the importance of spatial reasoning in the diffusion-based fusion process.

\begin{table*}
\centering
\begin{threeparttable}
\caption{Quantitative comparison of UniStateDLO and baselines for cross-frame tracking under different occlusion levels.}
\renewcommand{\arraystretch}{1.1}
\setlength\tabcolsep{3pt}
\begin{tabular}{p{3.8cm}|p{0.98cm}<{\centering}p{0.95cm}<{\centering}p{0.95cm}<{\centering}|p{0.98cm}<{\centering}p{0.95cm}<{\centering}p{0.95cm}<{\centering}|p{0.98cm}<{\centering}p{0.95cm}<{\centering}p{0.95cm}<{\centering}|p{0.98cm}<{\centering}p{0.95cm}<{\centering}p{0.95cm}<{\centering}} 
\toprule
\multirow{2}{*}{Method} & \multicolumn{3}{c|}{No occlusion} & \multicolumn{3}{c|}{10\% occluded} & \multicolumn{3}{c|}{30\% occluded} & \multicolumn{3}{c}{50\% occluded} \\
 & MPNE$\,\downarrow$ & PCN$\,\uparrow$ &NSS$\,\downarrow$ & MPNE$\,\downarrow$ & PCN$\,\uparrow$ &NSS$\,\downarrow$ & MPNE$\,\downarrow$ & PCN$\,\uparrow$ &NSS$\,\downarrow$ & MPNE$\,\downarrow$ & PCN$\,\uparrow$ &NSS$\,\downarrow$\\ 
\midrule
CDCPD2~\cite{wang2021tracking} & 11.18 & 50.94 & 0.3526 & 12.58 & 49.09 & 0.4694 & 19.35 & 33.94 & 0.7574 & 28.31 & 26.27 & 1.0312 \\
TrackDLO~\cite{xiang2023trackdlo} & 5.76 & 86.91 & 0.0418 & 5.89 & 86.87 & 0.0414 & 6.94 & 85.07 & 0.0409 & 11.64 & 64.04 & 0.0412 \\
\midrule
\textbf{Ours (\textit{Cross-Frame Tracking})} & \textbf{2.92} & \textbf{95.66} & \textbf{0.0331} & \textbf{3.03} & \textbf{95.10} & \textbf{0.0348} & \textbf{3.89} & \textbf{92.24} & \textbf{0.0351} & \textbf{7.24} & \textbf{80.58} & \textbf{0.0469} \\
\bottomrule
\end{tabular}
\label{tab:tracking_results}

\begin{tablenotes}
\footnotesize
\item * The performance is reported after sequential tracking for 30 frames, with the initial state set to the ground-truth for fair comparison across methods.
\end{tablenotes}

\end{threeparttable}
\end{table*}

% \vspace{-0.2cm}

\begin{figure*} [tb]
\vspace{-0.1cm}
\centering 
\raisebox{0pt}[\height]{
  \subfloat[MPNE metric for Trajectory 1]{ 
    \includegraphics[height=0.2125\textwidth]{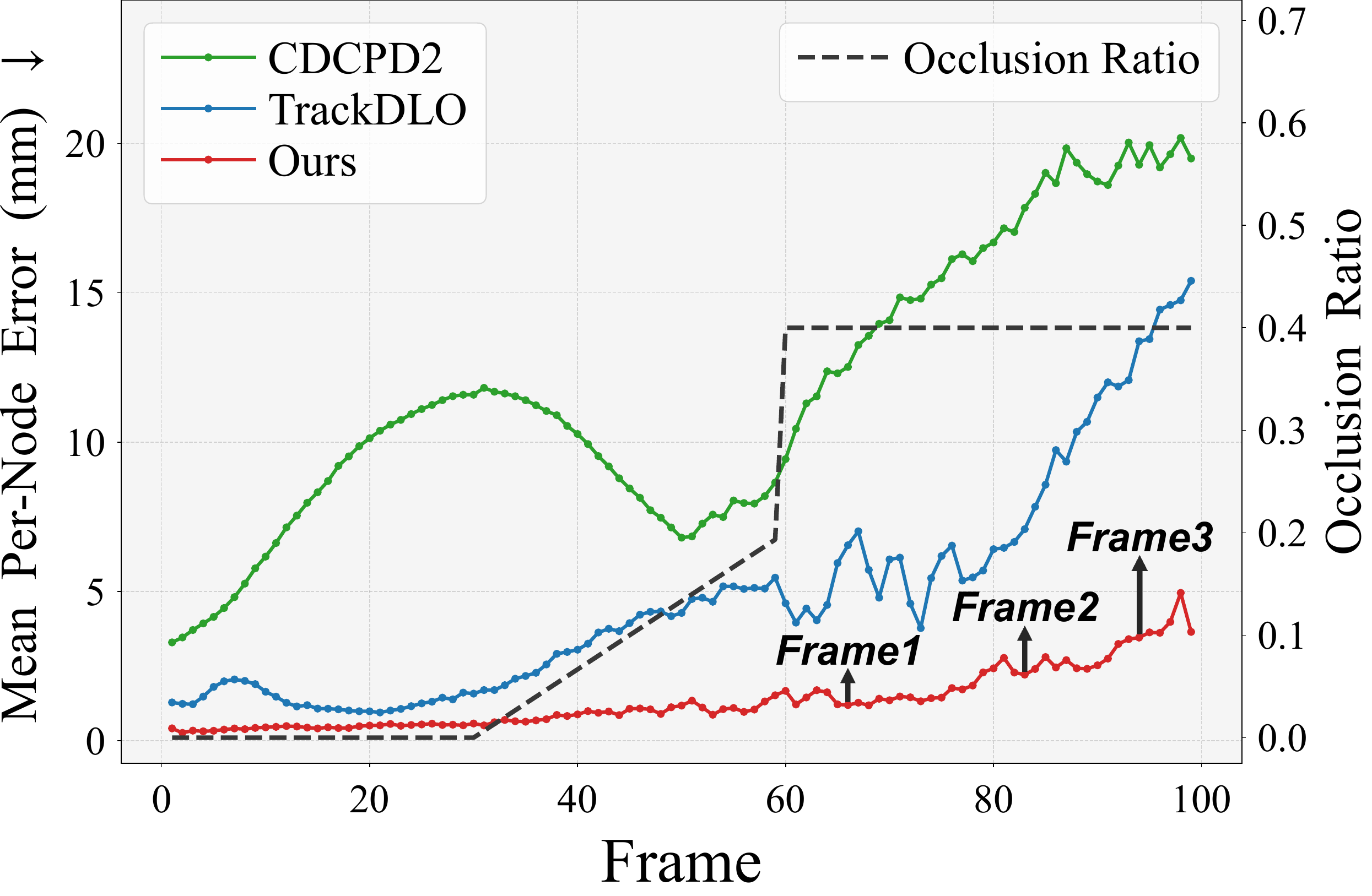}
  } }
  \hspace{-0.5cm}
  \raisebox{0pt}[\height]{
  \subfloat[Frame 1 in Trajectory 1]{ 
  \raisebox{4pt}[\height]{
    \includegraphics[height=0.202\textwidth]{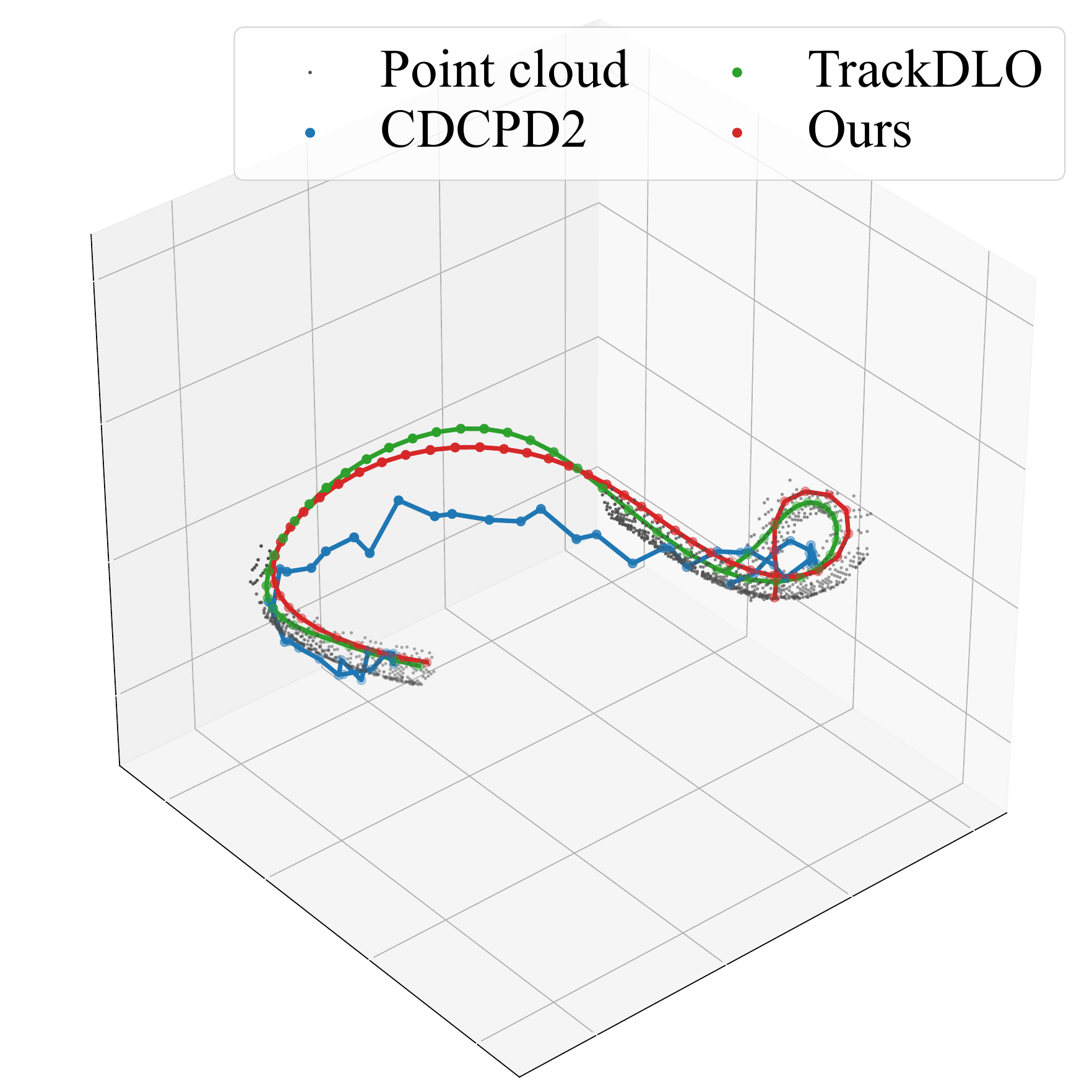}}
  } }
  \hspace{-0.6cm}
  \raisebox{0pt}[\height]{
  \subfloat[Frame 2 in Trajectory 1]{ 
  \raisebox{4pt}[\height]{
    \includegraphics[height=0.202\textwidth]{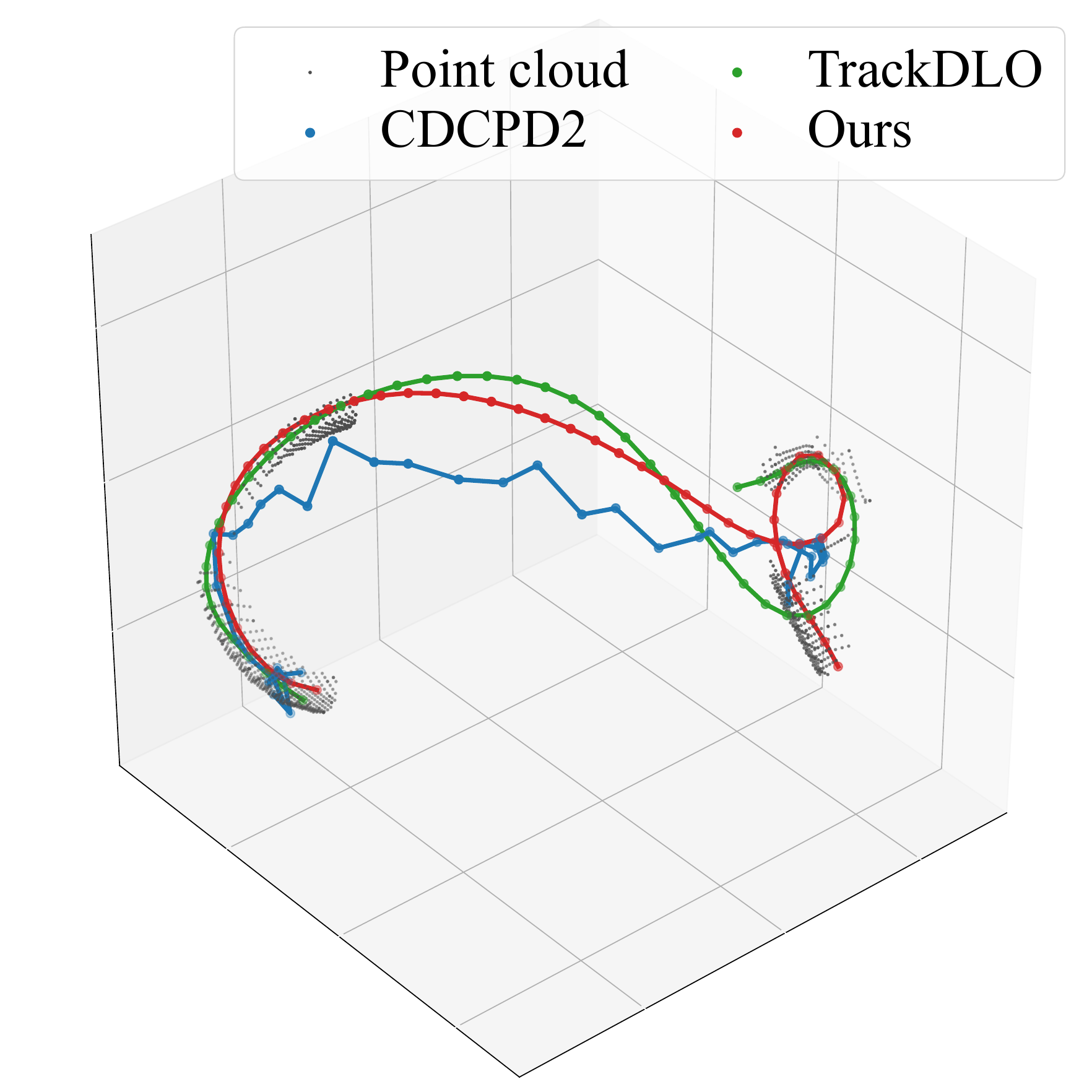}}
  } }
  \hspace{-0.6cm}
  \raisebox{0pt}[\height]{
  \subfloat[Frame 3 in Trajectory 1]{ 
  \raisebox{4pt}[\height]{
    \includegraphics[height=0.202\textwidth]{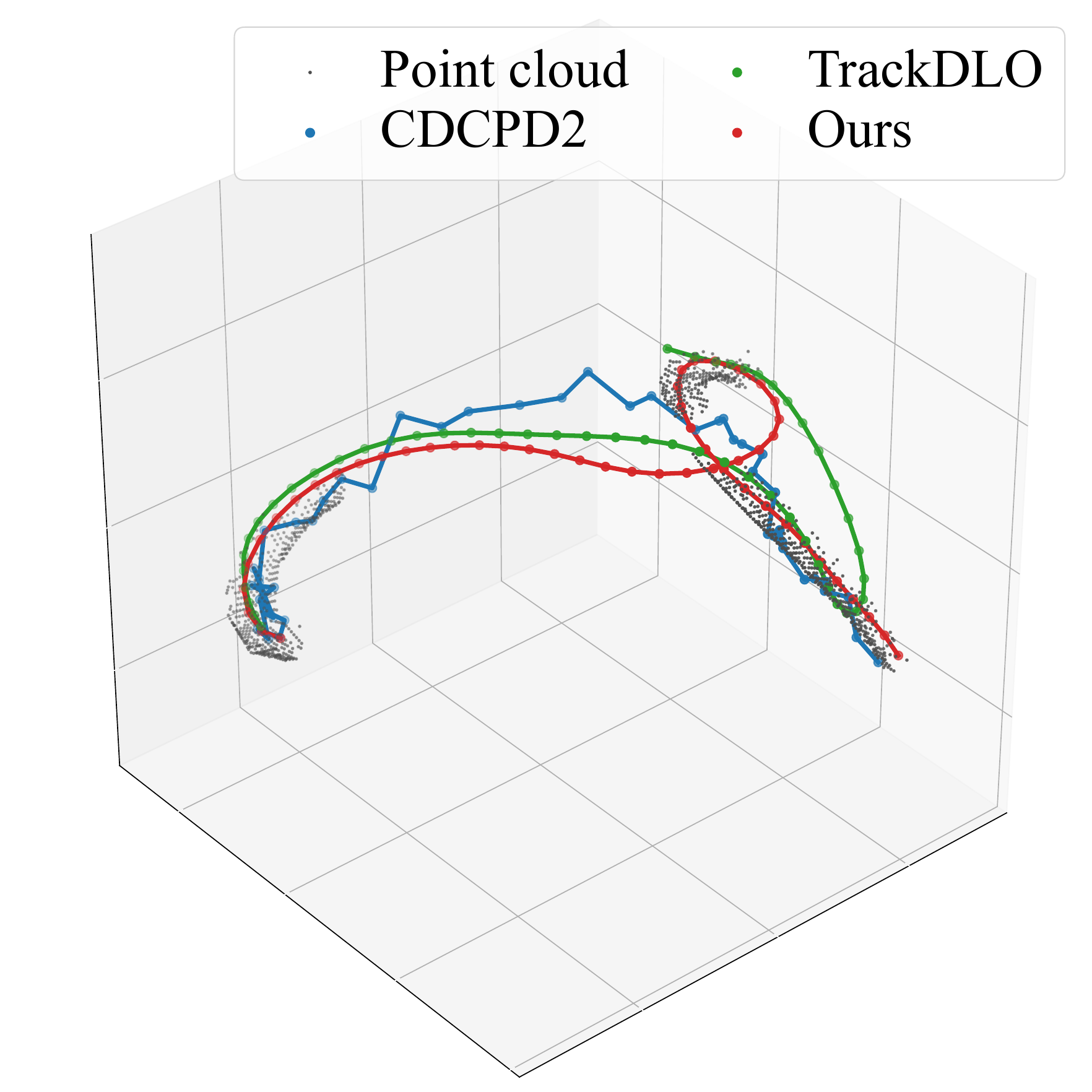}}
  }} \\
  
\raisebox{0pt}[\height]{
  \subfloat[MPNE metric for Trajectory 2]{ 
    \includegraphics[height=0.2125\textwidth]{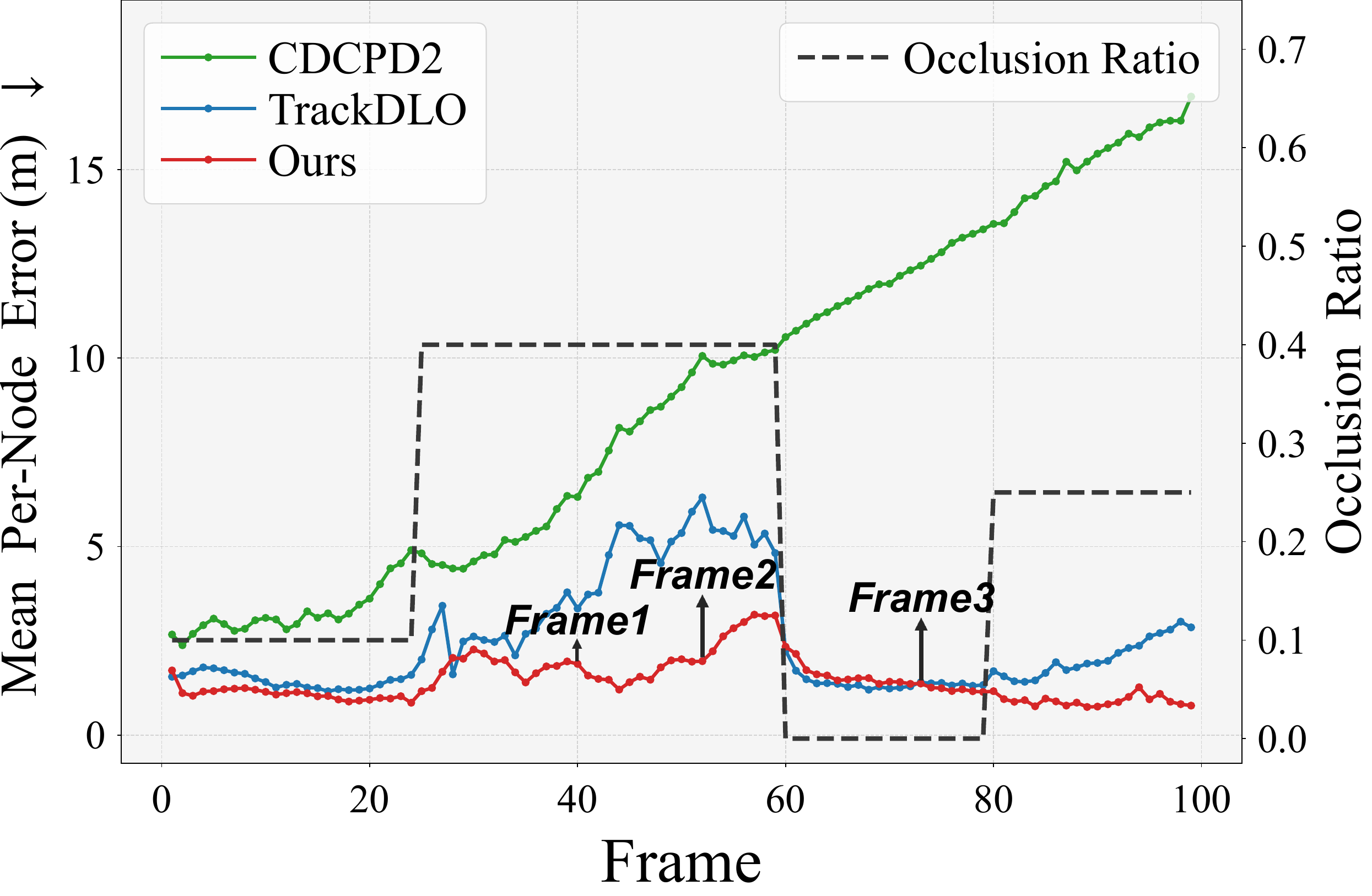}
  } }
  \hspace{-0.5cm}
  \raisebox{0pt}[\height]{
  \subfloat[Frame 1 in Trajectory 2]{ 
  \raisebox{4pt}[\height]{
    \includegraphics[height=0.202\textwidth]{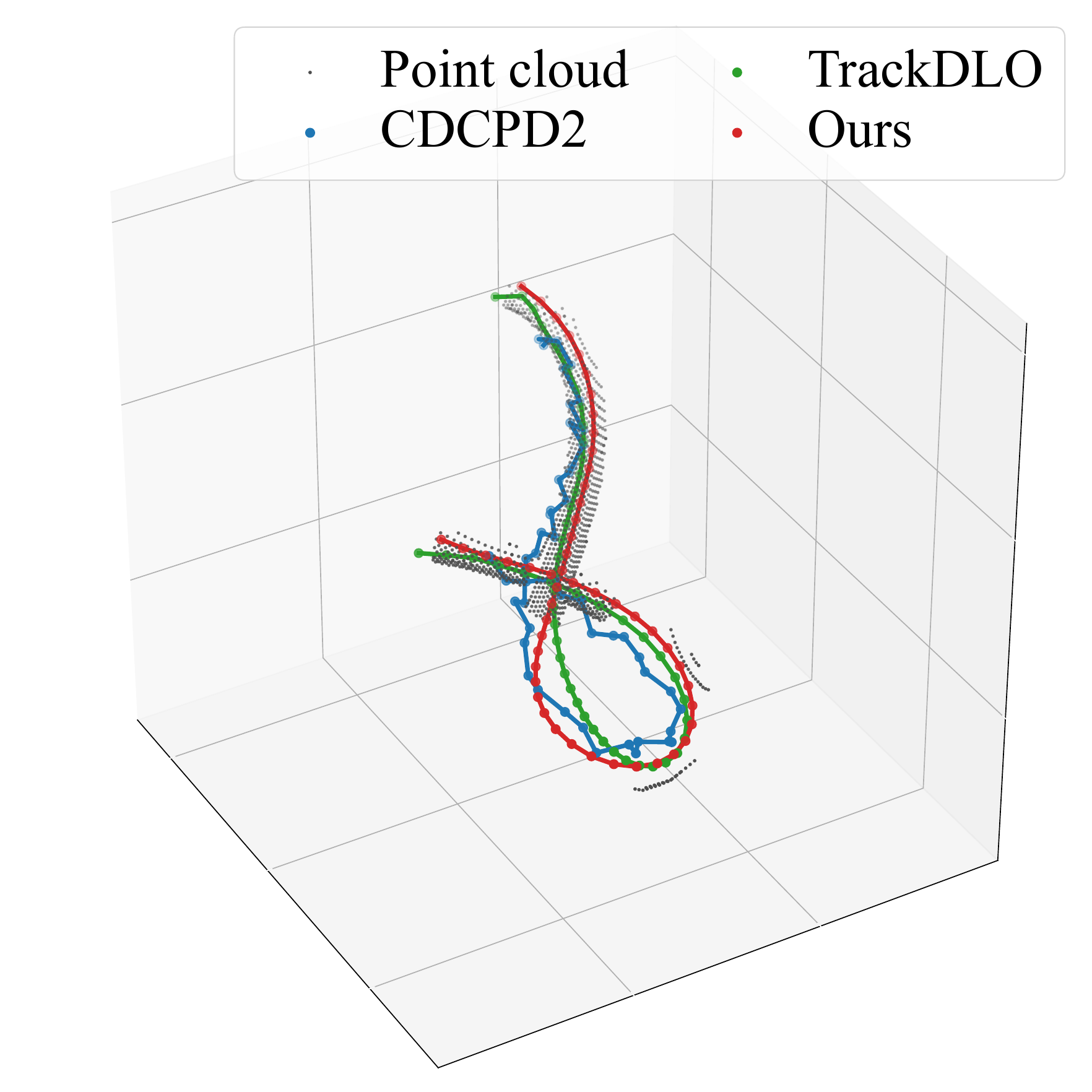}}
  } }
  \hspace{-0.6cm}
  \raisebox{0pt}[\height]{
  \subfloat[Frame 2 in Trajectory 2]{ 
  \raisebox{4pt}[\height]{
    \includegraphics[height=0.202\textwidth]{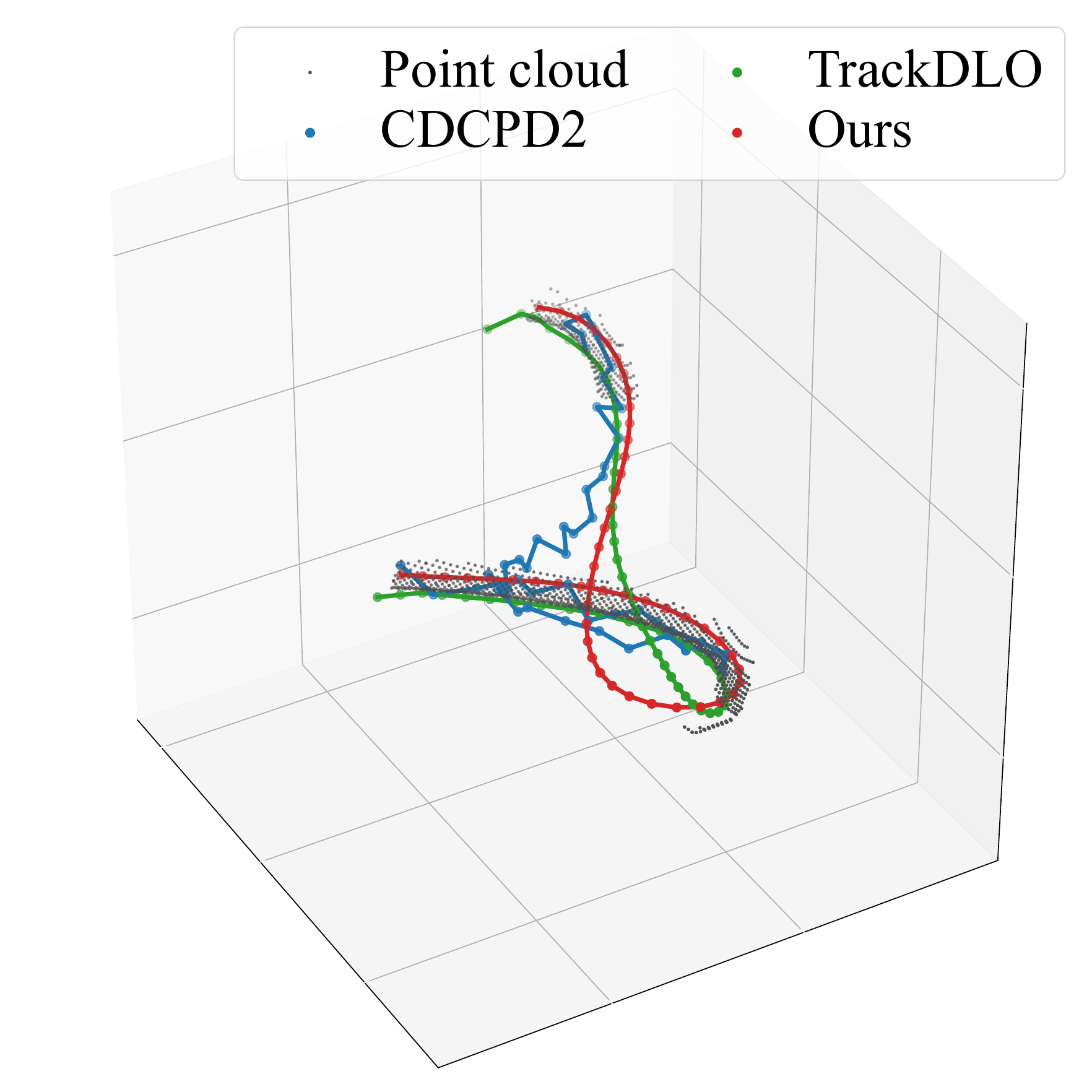}}
  } }
  \hspace{-0.6cm}
  \raisebox{0pt}[\height]{
  \subfloat[Frame 3 in Trajectory 2]{ 
  \raisebox{4pt}[\height]{
    \includegraphics[height=0.202\textwidth]{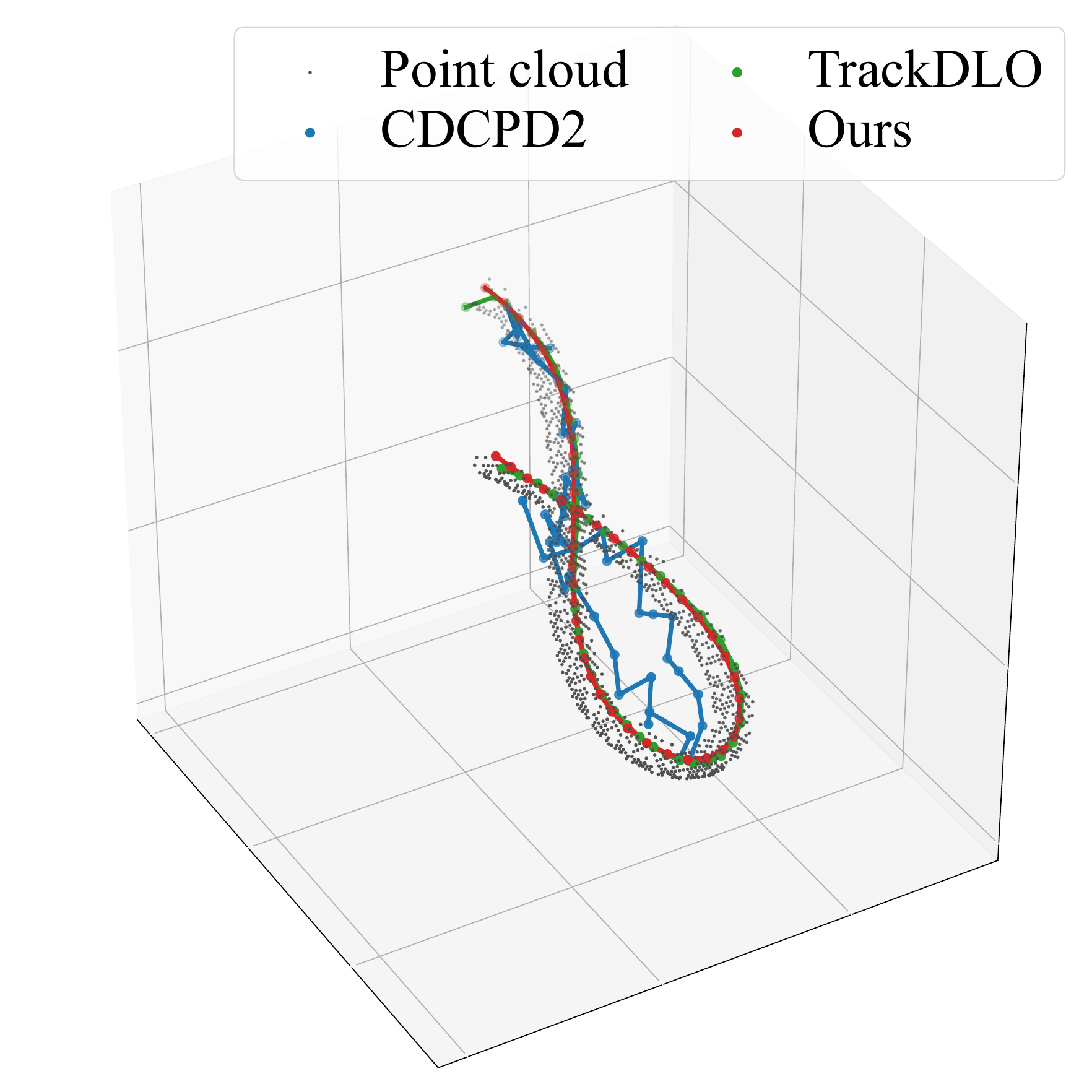}}
  }} 
  % \vspace{-2mm}
  \caption{Qualitative comparison of tracking performance among UniStateDLO (red), TrackDLO (blue), and CDCPD2 (green) on two DLO manipulation sequences in simulation. The first column presents the MPNE metric of all three methods over the full sequence, with the occluded portion of the DLO point cloud for each frame indicated by a black dotted line. The right three columns visualize the DLO point cloud and estimated nodes on three representative frames from each sequence.}
  \label{fig:tracking_baseline}
\end{figure*}

\subsection{Cross-Frame State Tracking}

\subsubsection{Comparison with State-of-the-art Methods}
For state tracking, we compare UniStateDLO against two strong state-of-the-art (SOTA) baselines based on non-rigid point set registration:
a) \textbf{CDCPD2} \cite{wang2021tracking}, which incorporates geometric constraints through regularization terms to enable robust deformable object tracking under occlusions;
and b) \textbf{TrackDLO} \cite{xiang2023trackdlo}, which further enforces segment-length preservation between nodes and applies Motion Coherence Theory to infer the positions of occluded nodes from visible ones.
Since both baselines are purely tracking-based and require external initialization, we use ground-truth DLO state in the first frame to provide the initial configuration for fair comparison. 

As shown in Table~\ref{tab:tracking_results}, UniStateDLO consistently outperforms the two baselines across all occlusion settings, achieving substantially lower tracking errors and higher overall accuracy. After continuous tracking for 30 frames, our method still maintains strong performance without significant error accumulation. Under this setting, the tracking error after 30 frames is still lower than the single-frame state estimation results reported in Table~\ref{tab:detection_results}. Moreover, cross-frame tracking provides improved temporal smoothness and better preserves topological consistency, as will be further demonstrated later.

Tracking performance on two challenging sequences is visualized in Fig.~\ref{fig:tracking_baseline}, where each sequence consists of 100 consecutive frames and the DLO is continuously deformed. The occlusion ratio varies over time to increase task difficulty, as indicated by the black dotted line. 
At the beginning of the first trajectory, although no external occlusion is present, CDCPD2 only roughly follows the DLO motion and quickly fails to preserve its geometric structure, whereas both TrackDLO and our method achieve low tracking errors. As the occlusion ratio gradually increases to 40\%, TrackDLO's accuracy deteriorates sharply, with its tracked nodes diverging from the true DLO configuration given heavily partial point clouds. In contrast, our method remains stable throughout, maintaining low errors and accurately recovering the DLO state even under severe occlusions and complex deformations.
For the second trajectory, the occlusion ratio peaks midway before decreasing to 0\% and 25\%. CDCPD2 exhibits a continuously increasing tracking error throughout the sequence, while TrackDLO performs better but still incurs substantially higher errors than ours, particularly under occlusions of up to 40\%. Once the occlusion disappears, our approach rapidly converges to the true DLO configuration, demonstrating strong self-correction capability after long-term occlusions.

\subsubsection{Ablation Study}

\begin{figure} [tb]
\vspace{-0.2cm}
  \centering 
  \hspace{-0.8cm}
  \subfloat[Frame 1]{ 
    \includegraphics[width=0.1715\textwidth]{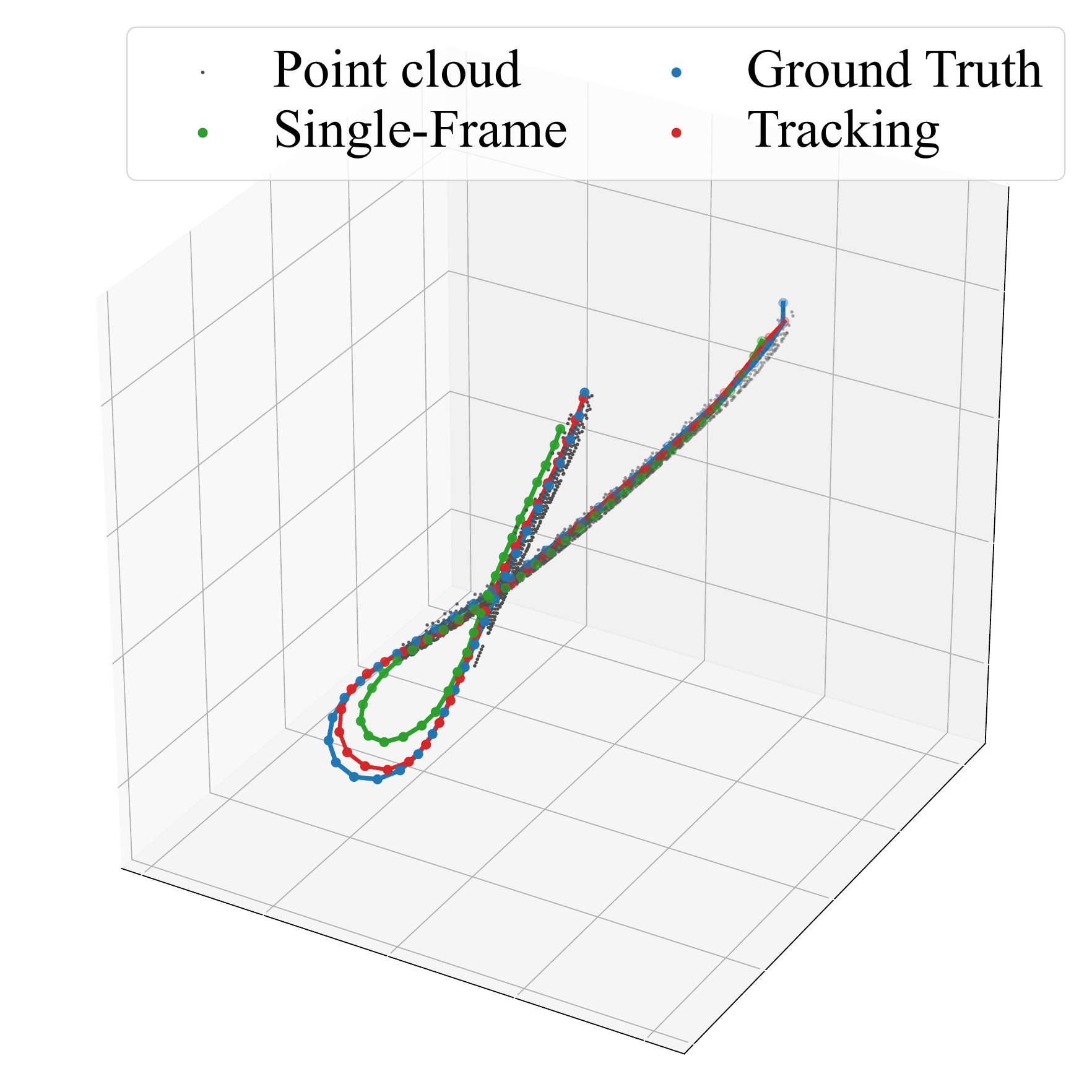}
  } 
  \hspace{-0.4cm}
  \subfloat[Frame 2]{ 
    \includegraphics[width=0.1715\textwidth]{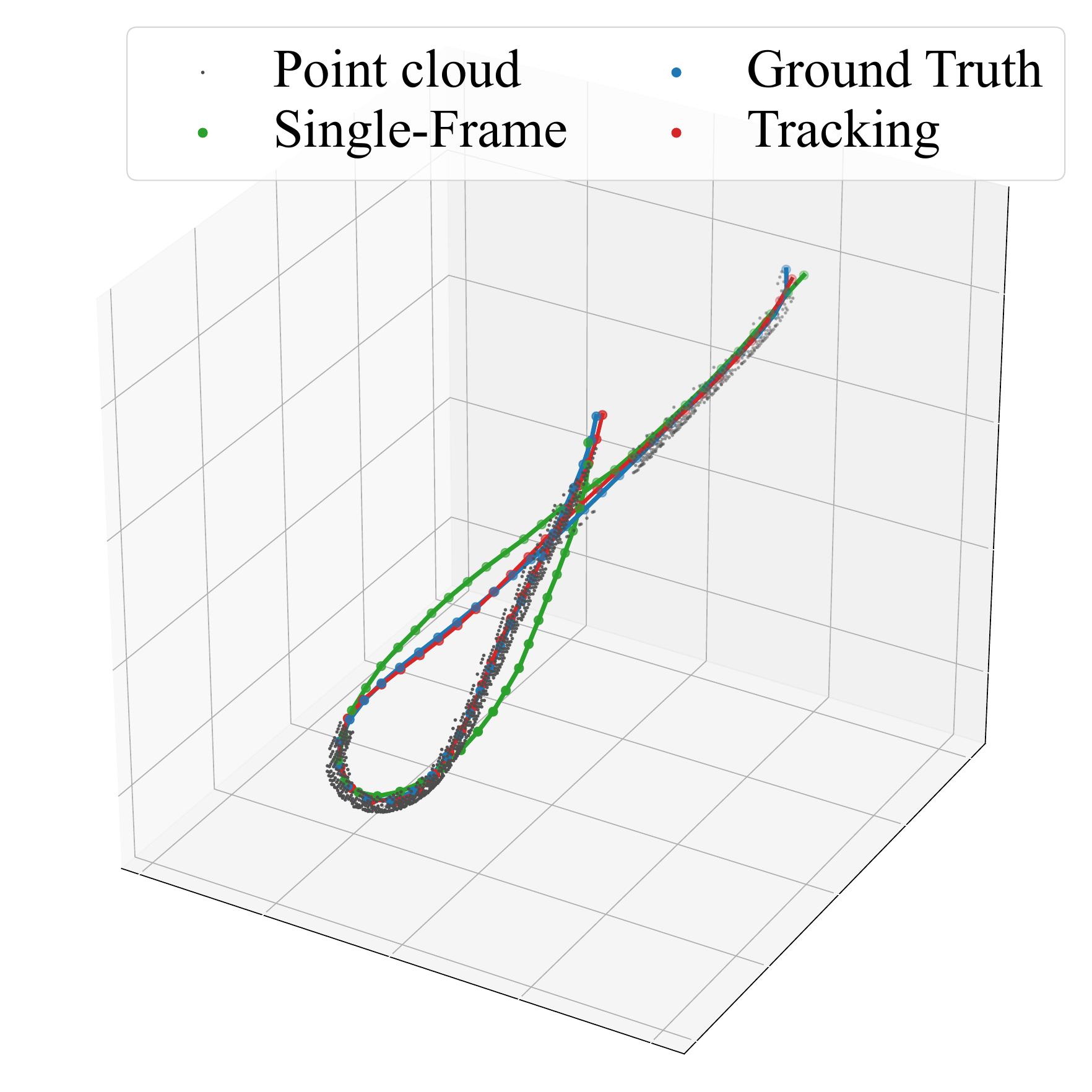}
  } 
  \hspace{-0.4cm}
  \subfloat[Frame 3]{ 
    \includegraphics[width=0.1715\textwidth]{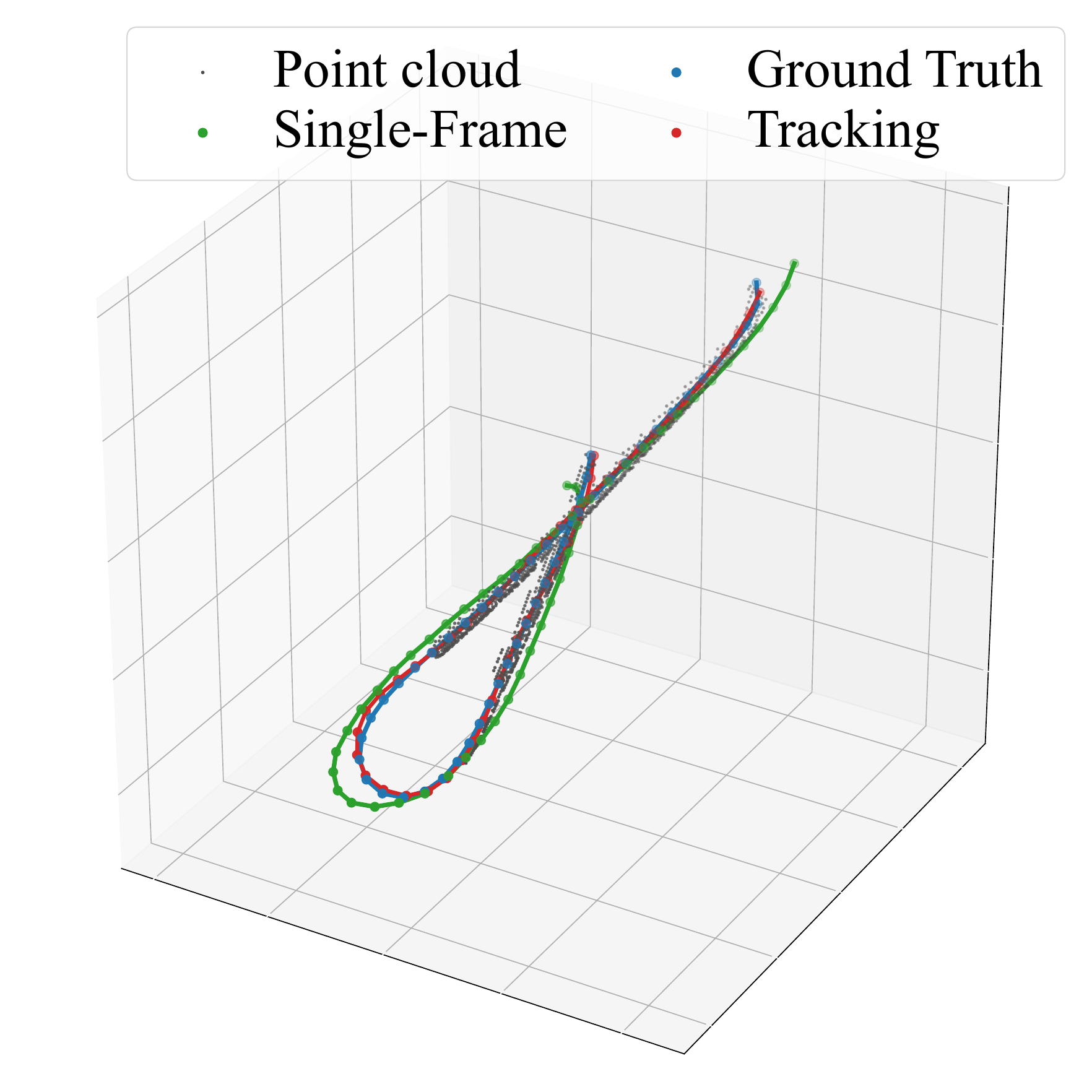}
  } 
  \hspace{-0.6cm}
  % \vspace{-2mm}
  \caption{Comparison of single-frame state estimation and cross-frame state tracking on several consecutive frames. Blue dots indicate ground-truth nodes, red denote tracking results, and green denote single-frame estimation results.}
  \label{fig:tracking_detection}
  % \vspace{-0.2cm}
\end{figure}

\begin{table}
\centering
\caption{Ablation study on the number of history frames used and the prediction types for cross-frame tracking.}
\renewcommand{\arraystretch}{1.1}
\setlength\tabcolsep{3pt}
\begin{tabular}{c|p{0.95cm}<{\centering}p{1cm}<{\centering}|p{1.25cm}<{\centering}p{1.25cm}<{\centering}p{1.25cm}<{\centering}} 
\toprule
\makecell{Level of \\ occlusion} & \makecell{History \\ Frames} & \makecell{Pred. \\ Type} & \makecell{2 Frames\\MPNE$\,\downarrow$} & \makecell{5 Frames\\MPNE$\,\downarrow$} & \makecell{10 Frames\\MPNE$\,\downarrow$}\\ 
\midrule
\multirow{5}{*}{No occlusion} & 2 & Del. & 0.453 & \textbf{0.978} & \textbf{1.603} \\
 & 2 & Abs. & 0.584 & 1.215 & 1.902 \\
 & 3 & Del. & 0.179 & 1.351 & 3.289 \\
 & 3 & Abs. & 0.283 & 1.673 & 3.478 \\
 & 4 & Del. & \textbf{0.139} & 1.376 & 3.703 \\ 
\midrule
\multirow{5}{*}{30\% occluded} & 2 & Del. & 0.559 & \textbf{1.163} & \textbf{2.689} \\
 & 2 & Abs. & 0.736 & 1.375 & 2.956 \\
 & 3 & Del. & 0.182 & 1.284 & 3.080 \\
 & 3 & Abs. & 0.288 & 1.736 & 3.532 \\
 & 4 & Del. & \textbf{0.142} & 1.373 & 3.477 \\
\bottomrule
\end{tabular}
\label{tab:ablation_tracking}
\end{table}

To further highlight the advantages of cross-frame tracking over single-frame estimation, we visualize the predictions of both approaches across consecutive frames in Fig.~\ref{fig:tracking_detection}. 
In these examples, the occluded regions vary dynamically, while the inter-frame node motion remains relatively small. When state estimation is performed independently for each frame, the estimated nodes (green points) reconstruct a plausible DLO configuration from the current partial point cloud but exhibit large frame-to-frame variations, making this single-frame estimation insufficient for closed-loop manipulation. In contrast, the tracked nodes (red points) maintain strong temporal continuity and topological consistency, closely following the ground-truth states even under severe and dynamically changing occlusions.

We also conduct an ablation study on the number of historical frames used and the prediction type to investigate whether incorporating longer temporal histories can further improve performance, as shown in Table~\ref{tab:ablation_tracking}. For the prediction types, \textit{Del.} denotes inter-frame motion, while \textit{Abs.} denotes absolute 3-D positions. Except for the case with two history frames, which leverages only the last frame's node predictions as priors to aggregate local features as adopted in our final method, history nodes from several past frames are encoded using an MLP, and the resulting embeddings are concatenated with local features to condition the diffusion model.
The per-node errors after continuous tracking for 2, 5, and 10 frames are reported here.
Experimental results show that while incorporating longer histories improves short-horizon DLO state tracking accuracy, it significantly degrades long-term tracking. We attribute this to overfitting to past states and the accumulation of errors over time. Furthermore, predicting cross-frame motions is more effective and stable than directly estimating absolute node positions in the next frame.

% \begin{table}
% \centering
% \caption{Ablation studies of different fusion methods.}
% \setlength\tabcolsep{3pt}
% \begin{tabular}{c|p{2.6cm}|p{0.98cm}<{\centering}p{0.95cm}<{\centering}p{0.95cm}<{\centering}} 
% \toprule
% \makecell{Level of \\ occlusion} & Method & MPNE$\,\downarrow$ & 
% PCN$\,\uparrow$ & 
% NSS$\,\downarrow$\\ 
% \midrule
% \multirow{4}{*}{No occlusion} & E2E Diffusion & 9.25 & 58.01 & 0.0323 \\
%  & MLP Fusion & 3.85 & 88.82 & 0.0327 \\
%  & Registration Fusion & 4.14 & 87.46 & 0.0756 \\
%  & Diffusion Fusion & \textbf{3.51} & \textbf{90.77} & \textbf{0.0314} \\ 
% \midrule
% \multirow{4}{*}{30\% occluded} & E2E Diffusion & 11.75 & 46.36 & 0.0332 \\
%  & MLP Fusion & 7.11 & 73.14 &  0.2074 \\
%  & Registration Fusion & 5.57 & 79.89 & 0.0792 \\
%  & Diffusion Fusion & \textbf{4.54} & \textbf{85.26} & \textbf{0.0327} \\ 
% \bottomrule
% \end{tabular}
% \end{table}

\begin{figure} [tb]
\centering
    {\includegraphics[width=0.995\linewidth]{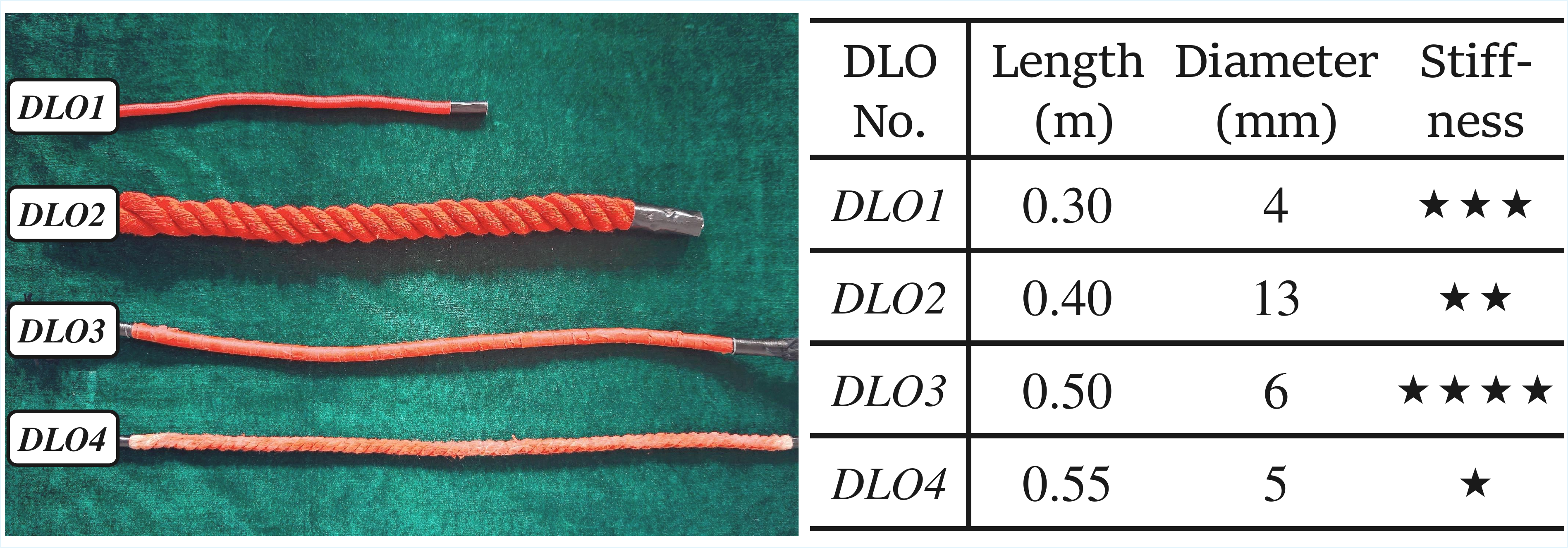}}
  \vspace{-0.45cm}
  \caption{DLOs used in real-world experiments and the physical parameters of each DLO.}
  \label{fig:dlos}
  \vspace{-0.1cm}
\end{figure}

\section{Real-world Experiments}\label{real-world section}
\subsection{Real-world Setup}

\begin{figure*} [tb]
% \vspace{-0.05cm}
\centering
% \setlength{\tabcolsep}{0pt} % 调整列间距
% \renewcommand{\arraystretch}{-0.2} % 调整行间距
% \begin{tabular}{cc}
% \specialrule{0em}{0pt}{0pt}
\subfloat{ 
    {\includegraphics[width=0.96\linewidth]{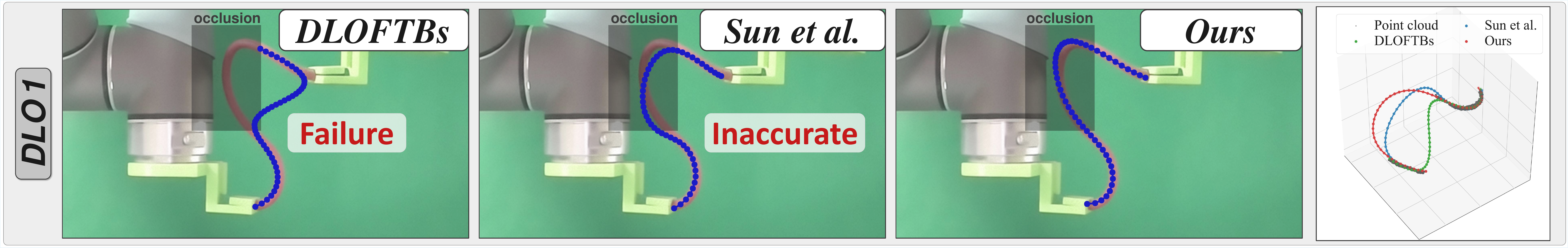}}  } 
  \\
  \vspace{-0.325cm}
  \subfloat{ 
    {\includegraphics[width=0.96\linewidth]{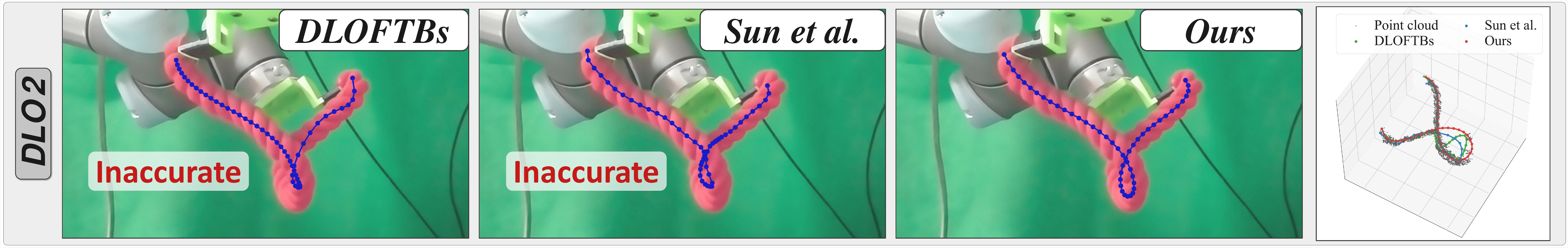}}  } 
  \\
  \vspace{-0.325cm}
  \subfloat{ 
    {\includegraphics[width=0.96\linewidth]{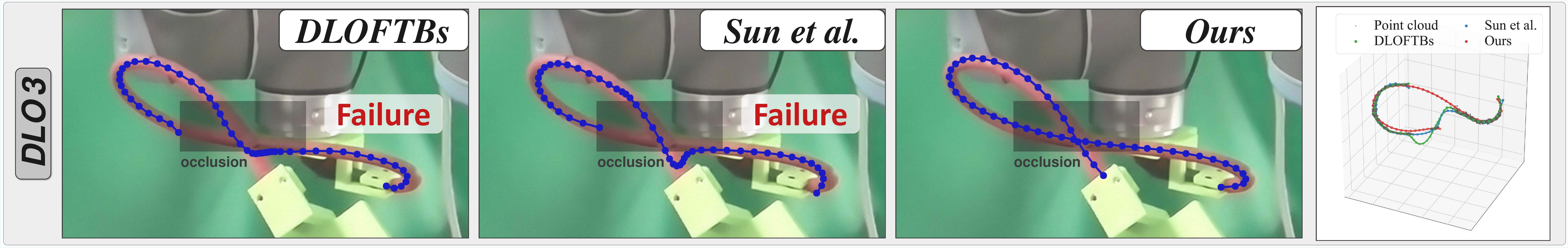}}  } 
  \\
  \vspace{-0.325cm}
  \subfloat{ 
    {\includegraphics[width=0.96\linewidth]{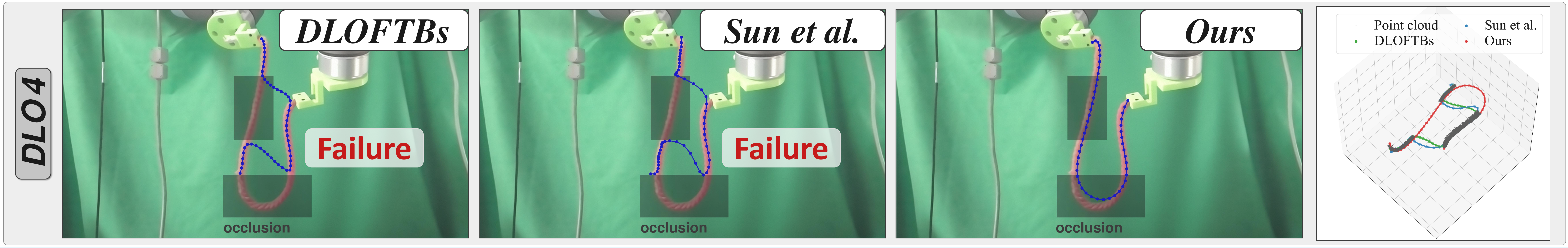}}  }
% \end{tabular}
\caption{Qualitative comparison of single-frame estimation performance of UniStateDLO (thrid column) on real-world DLOs against two baselines: DLOFTBs (first column) and Sun et al. (second column), where the blue dots refer to the reprojection of the estimated 3-D nodes and the darker regions in the images denote the masked areas for simulating occlusions. The last column visualizes the DLO point clouds together with the estimated DLO nodes in 3-D space.}
  \label{fig:realworld_detection}
\end{figure*}

\begin{figure*} [t!]
% \vspace{-0.05cm}
    \centering
    \includegraphics[width=0.96\linewidth]{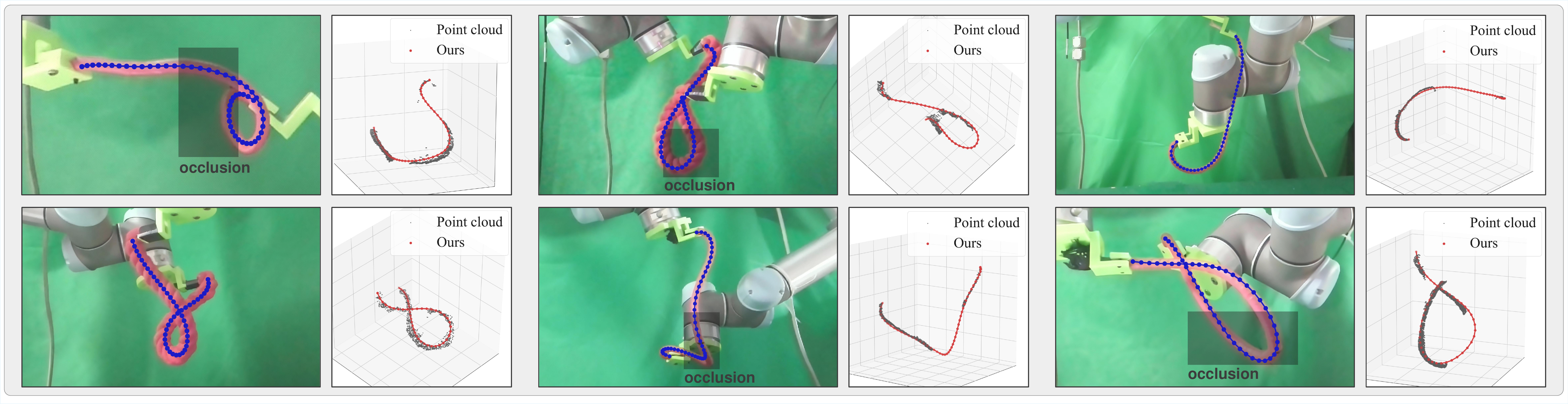} 
\caption{Additional visualized examples of UniStateDLO for single-frame state estimation on diverse real-world DLOs with occlusions. Each column depicts one case, with reprojected nodes overlaid on the image and the point cloud visualization.}
  \label{fig:realworld_detection_more_case}
  % \vspace{-0.1cm}
\end{figure*}

The real-world generalization performance of the proposed UniStateDLO is evaluated on four DLOs with distinct materials and physical properties, and their detailed parameters are shown in Fig.~\ref{fig:dlos}. Due to their varying flexibility, these DLOs exhibit different degrees of elastic and plastic deformation under external forces, presenting diverse challenges for reliable perception. Both single-frame state estimation and cross-frame tracking models are trained entirely on synthetic data and are directly applied to real-world data without any fine-tuning. 
During experiments, each DLO is rigidly grasped at both ends by dual UR5 robots, while the front-view RGB-D images are captured by an Azure Kinect camera. The DLO region is first segmented from the image via color thresholding, and the mask is then projected into 3-D space using the depth map to generate the point cloud input for our model. All inference is performed in real time on a single NVIDIA RTX 4090 GPU, where the single-frame estimation stage runs at on average 94.19 ms/frame and cross-frame tracking at 89.35 ms/frame.

% . may be fragmentary owing to self-occlusions or occlusions by obstacles.
% Some obstacles are also set in the environment to generate occlusions for point cloud of DLOs. 

% Results in Fig.~\ref{fig:exp_real_images} illustrates that our method can be directly applied to estimate the real-world DLO state with small sim-to-real gaps. 
% Even in some cases with self-intersection or occlusion by obstacles, our state estimations are still smooth and precise enough.
% Experiments on estimating the state of a moving DLO from each frame in a dynamic sequence (see Fig.~\ref{fig:exp_real_sequence}) also suggest the robustness of our method amidst heavy occlusions.

\begin{figure*} [tb]
\vspace{-0.3cm}
\centering
\setlength{\tabcolsep}{0pt} % 调整列间距
\begin{tabular}{cc}
% \specialrule{0em}{0pt}{0pt}
\vspace{-0.4cm}
\subfloat{ 
    {\includegraphics[width=8.8cm]{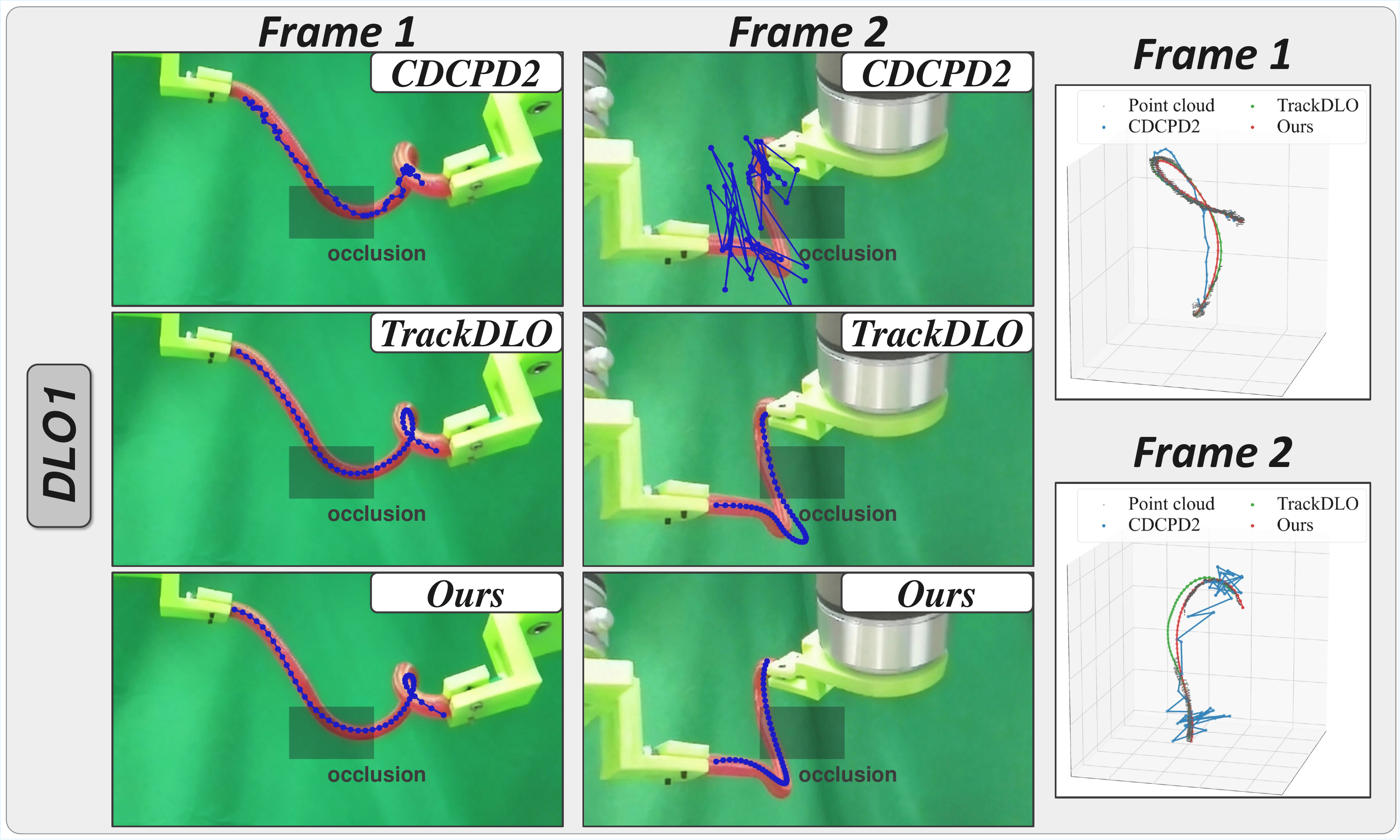}}
  } &
  \subfloat{ 
    \includegraphics[width=8.8cm]{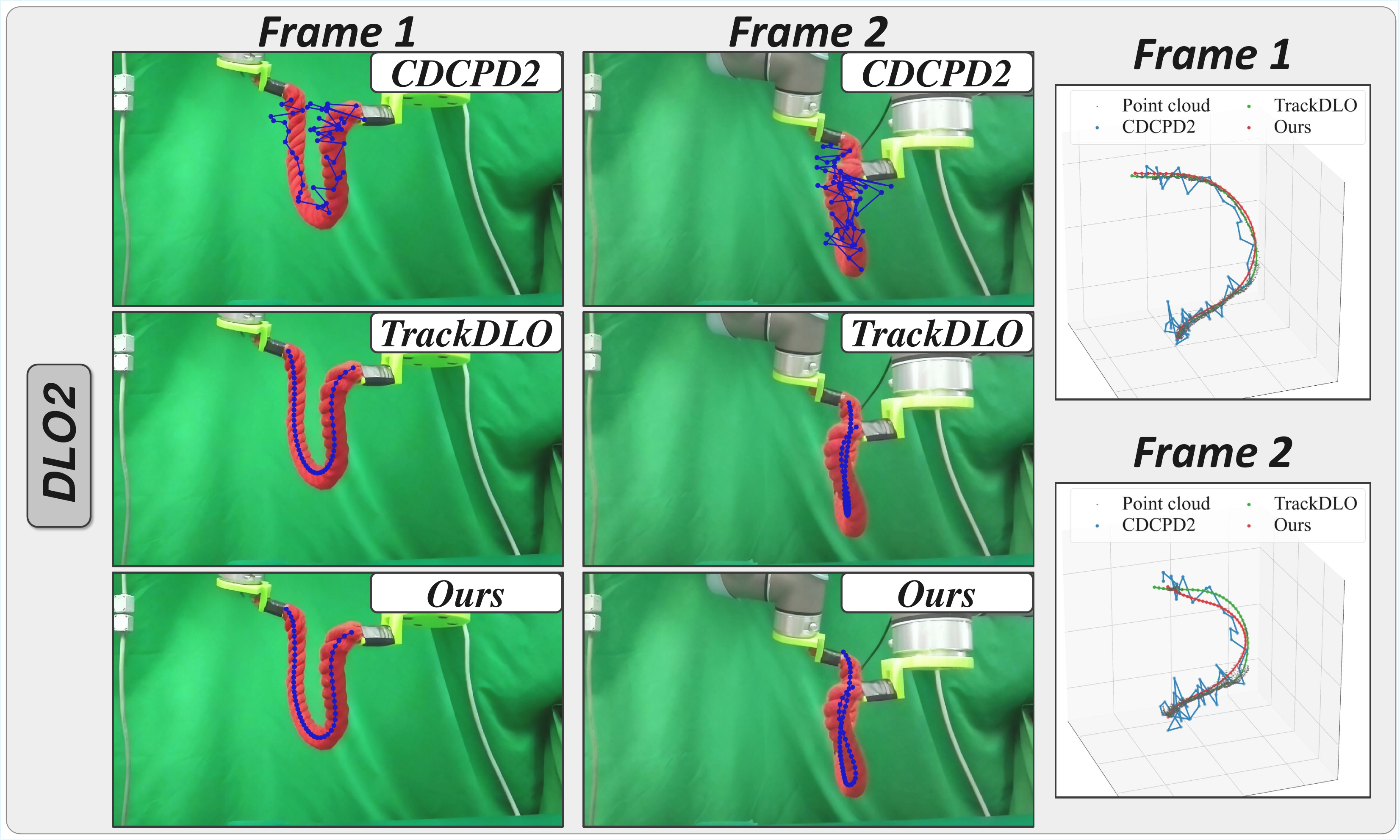}
  }
  \\
  \vspace{-0.05cm}
  \subfloat{ 
    {\includegraphics[width=8.8cm]{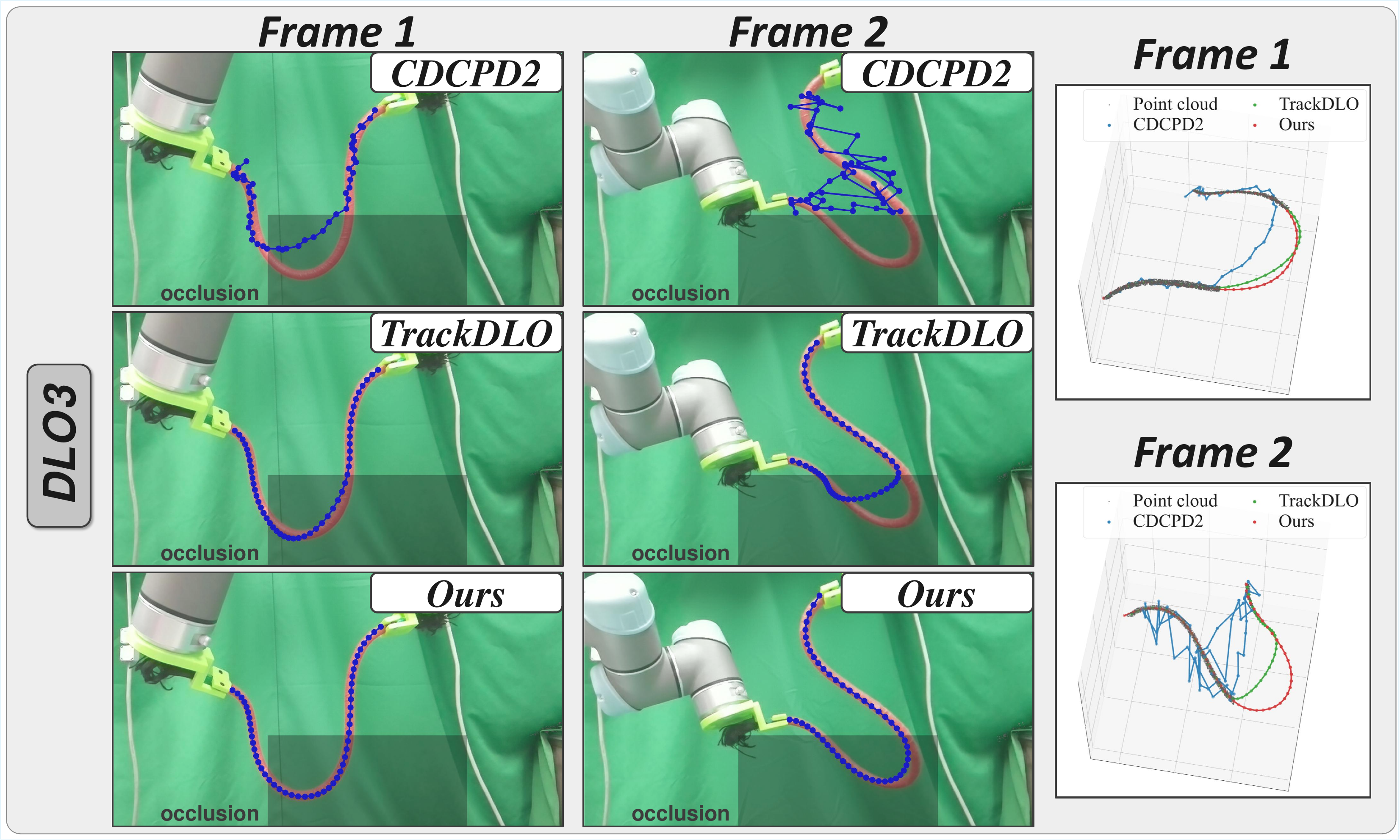}}
  } &
  \subfloat{ 
    \includegraphics[width=8.8cm]{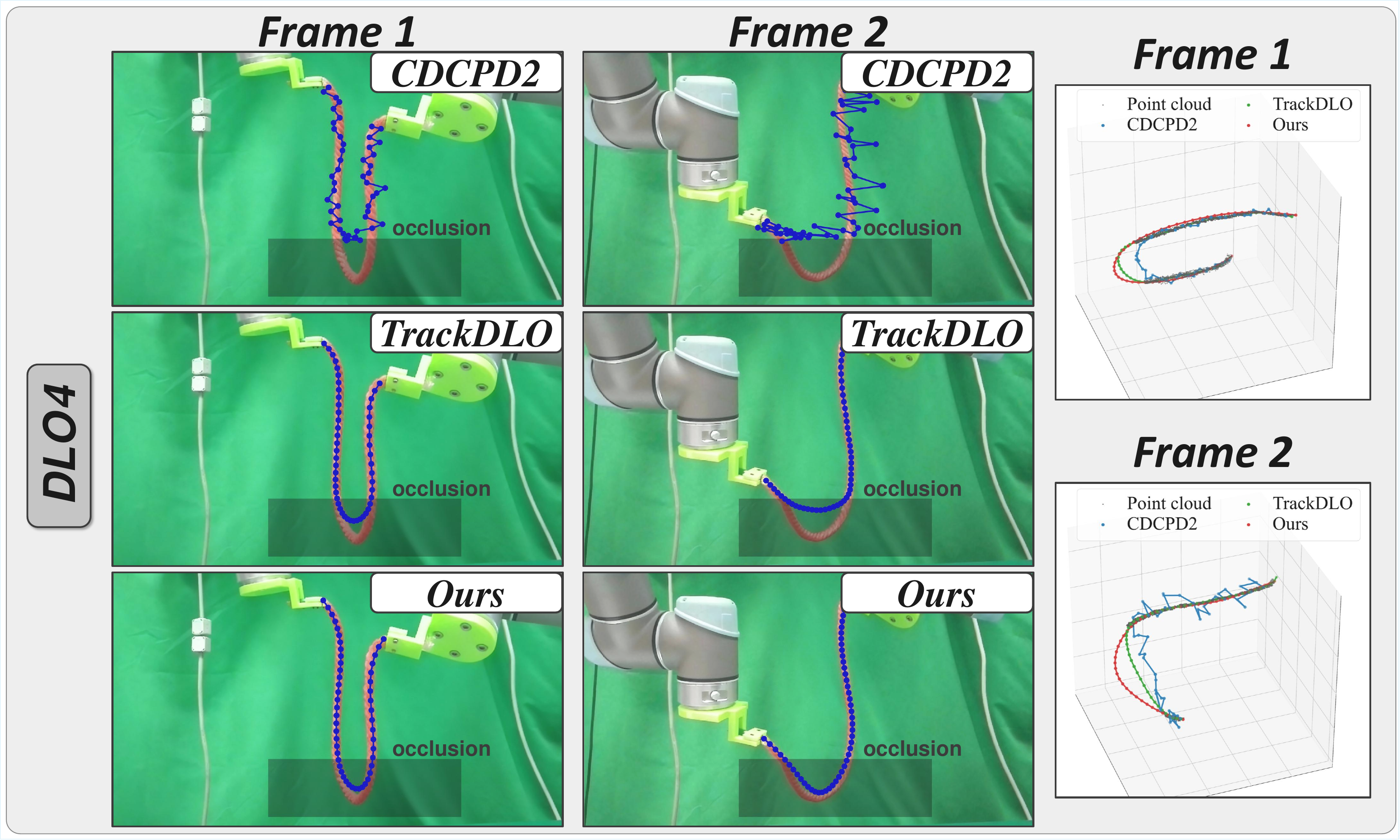}
  }
\end{tabular}
\caption{Qualitative comparison of cross-frame tracking performance of UniStateDLO (third row) on real-world DLO motion sequences against two baselines: CDCPD2 (first row) and TrackDLO (second row). For each DLO, two frames from the sequence are shown, with point clouds visualized alongside. (See the supplementary video for full sequences.)}
  \label{fig:realworld_tracking}
\end{figure*}

\begin{figure*} [t!]
    \centering
    \includegraphics[width=18cm]{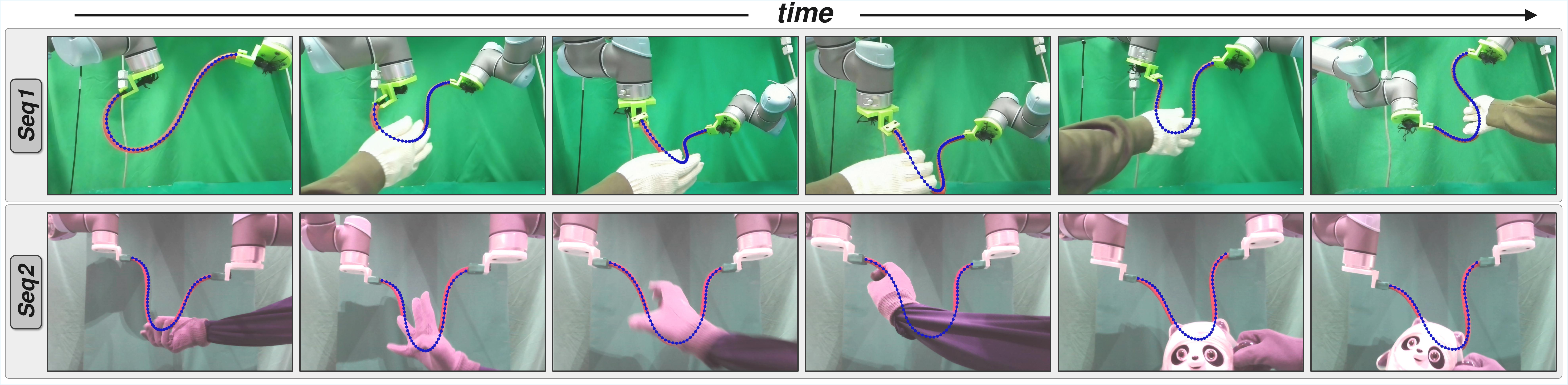} 
\caption{Tracking performance of UniStateDLO on two long-term DLO motion sequences involving large-scale deformations and dynamic, severe occlusions. Six representative frames from each sequence are shown here, arranged from left to right.}
  \label{fig:realworld_tracking_sequence}
\end{figure*}

% mannually change the number
\setcounter{figure}{\thefigure}
\addtocounter{figure}{1}
\begin{figure*} [t]
\vspace{-0.2cm}
  \centering 
  \includegraphics[width=18cm]{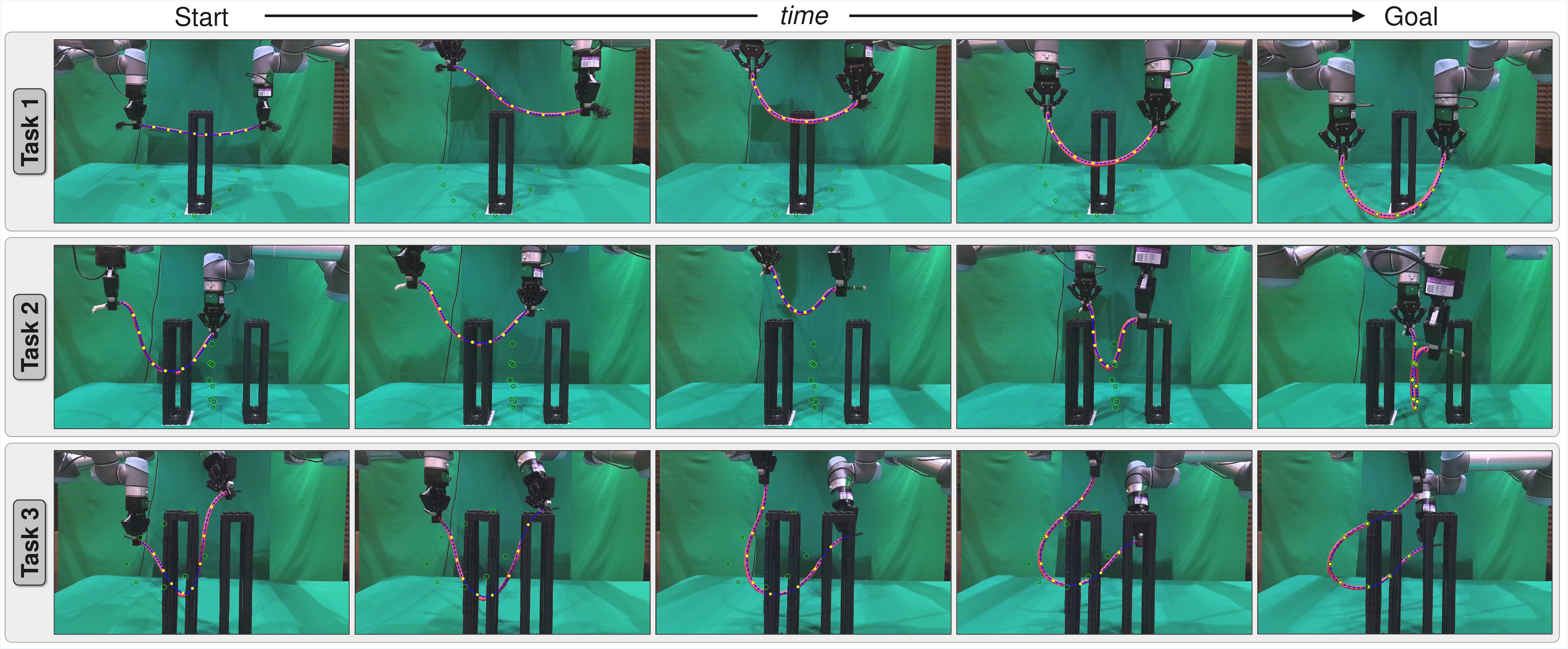} 
\caption{Autonomous robotic manipulation of DLOs using the proposed UniStateDLO as the front-end perception pipeline. For each shape control task, the yellow points denote the selected control targets, which are uniformly sampled from the predicted nodes, and the green+black circles indicate the desired DLO configurations. From left to right, we sequentially visualize the initial configuration, intermediate manipulation snapshots over time, and the final manipulated state.}
  \label{fig:manipulation}
  \vspace{-0.1cm}
\end{figure*}

\addtocounter{figure}{-2} % 减小计数器，使下一个图编号更小
\begin{figure} [t]
    \centering
    \includegraphics[width=\linewidth]{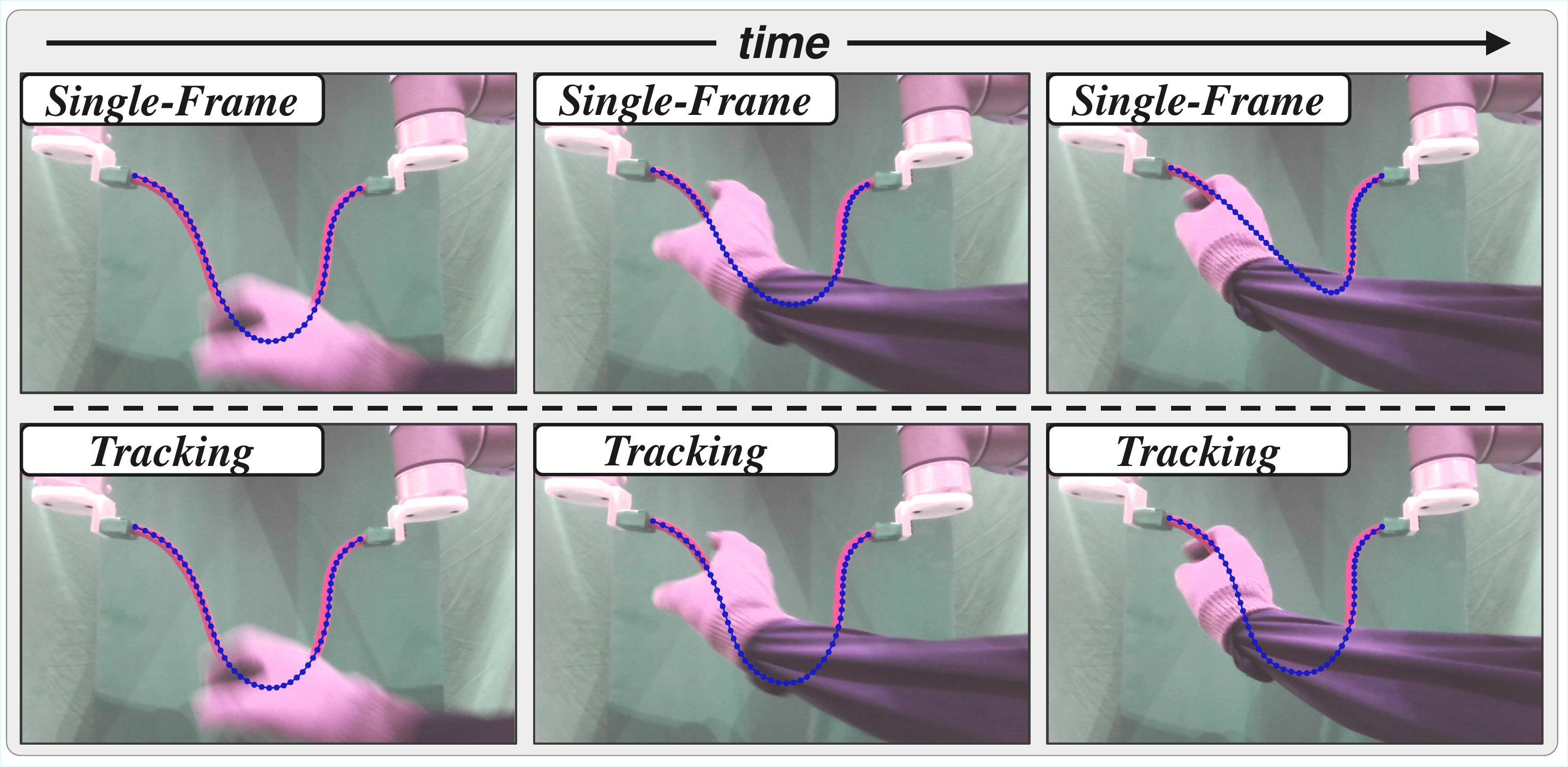} 
\caption{Comparison of single-frame estimation (top row) and cross-frame tracking (bottom row) across consecutive frames. Single-frame method fails to maintain temporal consistency and smoothness, whereas cross-frame tracking preserves both.}
  \label{fig:realworld_tracking_detection}
\end{figure}
\addtocounter{figure}{1}

\subsection{Single-Frame State Estimation}
Qualitative comparisons of real-world single-frame state estimation results of UniStateDLO against two baselines are presented in Fig.~\ref{fig:realworld_detection}, together with the corresponding point cloud visualizations. Real occlusions are simulated by randomly masking regions from the RGB images, shown as darker areas, to better illustrate the ground-truth DLO configurations. Note that for better illustration, the black robot support column in the images is removed via AI-based image editing.
The results indicate that DLOFTBs often fails to reconstruct occluded regions or incorrectly merges distinct segments, largely due to its reliance on 2-D skeleton-based ordering and manually-designed merging strategies. The approach of Sun et al. performs slightly better under simple shapes or minor occlusions but still struggles to maintain topological correctness and frequently produces inaccurate predictions when visibility is limited. In contrast, UniStateDLO consistently delivers accurate and robust state estimations across all scenarios, even under severe occlusions and complex geometries such as self-intersections, highlighting our strong generalization capability to real-world DLOs with diverse physical properties. Additional qualitative examples under varying deformation and occlusion conditions are provided in Fig.~\ref{fig:realworld_detection_more_case}.

\subsection{Cross-Frame State Tracking}
We further evaluate the real-world temporal tracking performance of UniStateDLO on continuous motion sequences for each DLO, as shown in Fig.~\ref{fig:realworld_tracking}. From each sequence, we present two representative frames by visualizing the reprojected node estimations in the image and the corresponding point clouds in 3-D. Unlike single-frame estimation, which reconstructs DLO states independently from isolated frames, this experiment assesses the model's ability to maintain geometric coherence and temporal consistency as the DLO undergoes large motions and complex deformations.
As illustrated in visualizations, our method produces smooth node trajectories that remain closely aligned with the ground truth, without noticeable accumulated drift, even when substantial portions of the DLO remain occluded over long durations. Conversely, CDCPD2 quickly loses structural integrity and becomes unstable as sequence progresses for a long time, while TrackDLO tolerates moderate occlusions but struggles to maintain consistent geometry under long-term or severe occlusion.
These results confirm that our data-driven generative modeling scheme effectively infers cross-frame per-node motion, enabling robust and stable DLO tracking and achieving strong generalization performance in real-world scenarios. Full tracking sequences are provided in the supplementary video.

Two long-term motion sequences with dynamic occlusions are visualized in Fig.~\ref{fig:realworld_tracking_sequence}, where the DLO undergoes large deformations and frequent visibility changes caused by the moving robot arms and human interactions. Throughout the sequences, our method reliably reconstructs the invisible parts while maintaining temporal smoothness and consistency. Even under rapid motions and challenging occlusion patterns, the tracked configurations remain stable and accurate. 
We further highlight the advantage of cross-frame tracking over single-frame estimation in Fig.~\ref{fig:realworld_tracking_detection}. When occlusions vary across consecutive frames, single-frame estimation can still produce plausible shapes based on the heavily partial point clouds in each individual frame, but the predicted node positions fluctuate significantly between frames. By effectively leveraging information from previous frames, the cross-frame tracking model preserves temporal coherence well, producing state estimations that more faithfully follow the true deformation and avoid abrupt shrinking or distortion within occluded regions.

\subsection{Integration in Constrained DLO Manipulation}
We further validate UniStateDLO in a closed-loop DLO shape control task, where the dual-arm robot rigidly grasps the two ends of a DLO and manipulates it toward a desired 3D configuration. The experimental setup features a highly constrained environment with multiple rigid obstacles, creating an especially challenging scenario in which continuous collision avoidance among the obstacles, the robot arms, and the DLO is required. 
Successful manipulation fundamentally relies on accurate and robust perception module to provide real-time feedback, which is extremely challenging due to the heavy occlusions, high-dimensional deformations, and dynamic interactions inherent to this constrained setting.
The obstacles frequently introduce severe occlusions and cause large portions of the DLO, sometimes even its endpoints, invisible for long periods, thereby demanding the perception module capable of reliably reconstructing the occluded parts. In addition to occlusion, the DLO undergoes substantial global motion and continuous local deformation throughout the manipulation process, requiring the perception module to handle diverse configurations while preserving temporal smoothness.

To accomplish the overall task, we adopt the complementary framework proposed in \cite{yu2024generalizable}, which combines whole-body global planning and precise closed-loop control. The global planner searches for a feasible, collision-free trajectory under complex geometric constraints without accurate models, while the closed-loop controller compensates for modeling errors during execution by leveraging the real-time state feedback provided by UniStateDLO. The local controller is implemented as a model predictive controller (MPC) with hard constraints, including local obstacle avoidance and overstretch prevention. The DLO motion model used in MPC follows the Jacobian formulation in \cite{yu2022global}, which maps the linear velocities of the robot arms to the motion of DLO nodes via a configuration-dependent Jacobian matrix.
As reported in \cite{yu2024generalizable}, most manipulation failures in previous works stem from perception issues, either the perception algorithm breaks down when large portions of the DLO become occluded, or the controller becomes unstable when the estimated states exhibit abrupt jumps across frames. Consequently, prior researches often require carefully designed tasks with restricted motion ranges and meticulously selected camera viewpoints to avoid large occlusions.
In contrast, UniStateDLO serves as a plug-and-play front-end perception module that operates robustly from a simple front-view RGB-D setup without any special task design and viewpoint selection.

Snapshots of three shape-control tasks executed in constrained environments are shown in Fig.~\ref{fig:manipulation}. A uniformly spaced subset of estimated nodes (8 yellow points) is selected as control targets, which are manipulated toward the desired goal positions (green+black circles) to form the target DLO configuration.
Despite severe occlusions occurring in both the initial and intermediate stages of manipulation, the perception module consistently provides accurate and temporally stable state estimates, enabling the controller to progressively deform the DLO toward the desired shape while avoiding collisions. Notably, in the final sequence, one endpoint of the DLO becomes completely invisible for an extended period, yet the performance of our tracking model remains unaffected.
Overall, these results demonstrate that UniStateDLO delivers high accuracy, strong robustness, and real-time performance, effectively supporting stable feedback control of deformable objects in complex and highly constrained environments.

% \vspace{-0.3cm}
\section{Conclusion and Discussion}
\subsection{Conclusion}
Overall, this paper presents UniStateDLO, a unified pipeline for accurate and robust DLO perception that addresses the fundamental challenge of frequent occlusions in constrained manipulation scenarios. By leveraging a diffusion-based generative formulation to capture the complex high-dimensional distribution of DLO states, our framework unifies both  single-frame state estimation and cross-frame tracking, enabling reliable reconstruction of complete DLO configurations from highly partial point cloud observations. 
Because DLO point clouds lack distinctive visual features, making global representations insufficient for fine-grained estimation, we introduce a two-branch architecture that captures both global structure and local geometric context, followed by a diffusion-based module to fuse two branches for the precise and robust reconstruction. 
After obtaining the initial state via single-frame estimation, cross-frame tracking is then enabled by conditioning another diffusion model on features aggregated around the previously estimated nodes, allowing the system to infer accurate and temporally consistent inter-frame node motions.
In addition, effective point cloud normalization and post-processing strategies further enhance robustness and overall performance.

Extensive simulation and real-world experiments demonstrate that UniStateDLO achieves precise and stable state estimation and tracking even under heavy occlusions and large deformations, significantly outperforming existing state-of-the-art methods. Trained exclusively on a large-scale synthetic dataset without any real-world supervision, our model generalizes effectively to a wide variety of real DLOs with different materials and physical properties. Moreover, integration into a closed-loop DLO shape control system with multiple obstacles, where our approach consistently delivers high accuracy, strong robustness, and real-time performance, further validates its effectiveness to support stable feedback control of deformable linear objects in complex, highly constrained environments as the front-end perception module.

By releasing the full synthetic dataset, code implementations, and trained models to support reproducible research, 
we hope that UniStateDLO will provide a solid and reliable perception foundation and get broad adoption in deformable linear object manipulation. We envision that this framework will ultimately enable robots to perform more sophisticated, accurate, and robust DLO manipulation in complex and challenging 3-D real-world environments.

\subsection{Limitations}
Several limitations of our approach remain and can be further improved in future work:
\begin{enumerate}
\item This article focuses on state estimation and tracking given segmented point clouds, where DLO segmentation is simplified via color thresholding. In practical scenarios with cluttered backgrounds or multiple DLOs, general segmentation approaches \cite{kirillov2023segment,zhang2023faster} or DLO-specific segmentation approaches \cite{dinkel2022wire, caporali2022fastdlo, caporali2023rtdlo, zanella2021auto} will be necessary.

\item In tasks such as knotting or shoe lacing, DLOs with very soft materials exhibit complex deformations involving multiple knots. As such behaviors require strong physical constraints that are hard to model explicitly, existing methods \cite{tang2018track,luo2025tsl} typically rely on integrating physical simulation. Since our synthetic dataset primarily includes elastic DLOs, handling such highly soft and knotted DLOs needs future research.
\item Although our method can implicitly address endpoint occlusion by normalizing the point cloud with last-frame node estimations and maintaining temporal smoothness, it does not explicitly reason about endpoint visibility. This may limit robustness in extreme cases with prolonged endpoint occlusion.
\item The proposed pipeline incorporates several carefully designed components, such as the two-branch network and multiple diffusion-based modules, which increases system complexity and may complicate deployment. Future work could focus on a more compact and streamlined design that maintains performance.
\end{enumerate}

Moreover, the front-end perception module and back-end controller are currently implemented as separate components. In future work, we plan to incorporate uncertainty modeling into the perception pipeline and integrate it more tightly with downstream planning and control to further enhance robustness in challenging manipulation settings. The framework could also benefit from more advanced generative models that provide better performance and faster inference.

% \section*{Acknowledgments}

\bibliographystyle{IEEEtran}
\bibliography{IEEEabrv, ref}
% \bibliographystyle{IEEEtran}
% \bibliography{ref.bib}

\vfill

\end{document}